\documentclass{article}
\usepackage[final, nonatbib]{neurips_2022}
\usepackage[numbers]{natbib}
\usepackage[utf8]{inputenc} 
\usepackage[T1]{fontenc}    
\usepackage{hyperref}       
\usepackage{url}            
\usepackage{booktabs}       
\usepackage{amsfonts}       
\usepackage{nicefrac}       
\usepackage{microtype}      
\usepackage{graphicx}
\usepackage{amsmath}
\usepackage{subcaption}
\usepackage{xcolor}
\usepackage{enumitem}

\newcommand{\vect}[1]{\boldsymbol{#1}}
\usepackage{soul}
\usepackage{comment}
\usepackage{amssymb}
\usepackage{multirow}
\DeclareMathAlphabet\mathbfcal{OMS}{cmsy}{b}{n}

\title{A Mechanistic Analysis of Transformers \\ for Dynamical Systems}

\author{
  Gregory Duthé\\
  ETH Zürich\\
  \texttt{duthe@ibk.baug.ethz.ch} \\
   \And
  Nikolaos Evangelou\\
  Johns Hopkins University\\
  \\
  \And
  Wei Liu\\
  Singapore-ETH Centre\\
  \And
  Ioannis G. Kevrekidis\\
  Johns Hopkins University\\
  \And
  Eleni Chatzi\\
  ETH Zürich\\
}

\begin{document}

\maketitle

\begin{abstract}
Transformers are increasingly adopted for modeling and forecasting time-series, yet their internal mechanisms remain poorly understood from a dynamical systems perspective. In contrast to classical autoregressive and state-space models, which benefit from well-established theoretical foundations, Transformer architectures are typically treated as black boxes. This gap becomes particularly relevant as attention-based models are considered for general-purpose or zero-shot forecasting across diverse dynamical regimes.
In this work, we do not propose a new forecasting model, but instead investigate the representational capabilities and limitations of single-layer Transformers when applied to dynamical data. Building on a dynamical systems perspective, we interpret causal self-attention as a linear, history-dependent recurrence and analyze how it processes temporal information. Through a series of linear and nonlinear case studies, we identify distinct operational regimes. For linear systems, we show that in the single-head attention-only setting, the convexity constraint imposed by softmax attention restricts the class of autoregressive operators that can be represented, leading to oversmoothing when the target dynamics require mixed-sign lag coefficients. For nonlinear systems under partial observability, attention instead acts as an adaptive delay-embedding mechanism, enabling effective state reconstruction when sufficient temporal context and latent dimensionality are available.
These results help bridge empirical observations with classical dynamical systems theory, providing insight into when and why Transformers succeed or fail as models of dynamical systems.
\end{abstract}

\section{Introduction}

Understanding and modeling dynamical systems using data, in the form of observations, is a central problem in nonlinear science, with applications ranging from fluid mechanics and structural dynamics \cite{amoudruz2025,LIU2022109276,RAISSI2019686} to neuroscience, chemical kinetics, weather and power systems and beyond \cite{10.1063/5.0297336,10.1063/5.0291493,chiavazzo2014reduced, wu2023interpretable}. 
Classical approaches rely either on use of explicit governing equations or on well-established data-driven identification frameworks, such as autoregressive and state–space models, for which stability, observability, and identifiability properties are well understood \cite{box1976analysis,Kantz_Schreiber_2003}. 
These frameworks provide a principled connection between data, latent state representations, and the underlying geometry of dynamical systems, including attractors and invariant manifolds.

More recently, machine-learning architectures originally developed for sequence modeling have been increasingly applied to dynamical systems modeling, particularly for purely data-driven inference, raising fundamental questions about their expressive power and their relationship to classical dynamical systems theory. 
Among these architectures, the Transformer model~\cite{vaswani2017attention} has emerged as a dominant paradigm. 
Originally introduced for natural language processing, Transformers are now widely used in computer vision, speech processing, and scientific machine learning~\cite{pmlr-v267-holzschuh25a}, including the modeling and forecasting of dynamical systems~\cite{geneva2022transformers,sitapure2023introducing,gao2025learning}. 
Their defining feature is the attention mechanism, which enables flexible aggregation of information across a temporal context \textit{through parallel rather than sequential computation}. This property has led to strong empirical performance in step-ahead prediction tasks, including for nonlinear and weakly chaotic systems~\cite{valle2025forecasting, choi2025defining}.

A growing body of work has explored the use of Transformers for dynamical and physical systems. 
Early studies demonstrated that attention-based models can learn surrogate evolution maps when provided with suitable spatiotemporal tokenizations. 
Geneva and Zabaras~\cite{geneva2022transformers}, for example, modeled diverse dynamical systems using a ``vanilla'' Transformer architecture, relying on Koopman-based embeddings to project high-dimensional states into lower-dimensional token representations. 
Subsequent work investigated the direct application of Transformers to chaotic time-series forecasting, showing that autoregressive prediction is feasible when the Lyapunov exponent is sufficiently low~\cite{valle2025forecasting}. 
More recent efforts have extended these ideas towards large pretrained scientific foundation models. 
Aurora, for instance, is proposed as a foundation model for the Earth system, trained on heterogeneous atmospheric and oceanic datasets and equipped with an encoder–processor–decoder architecture to evolve a latent three-dimensional spatial representation forward in time~\cite{bodnar2025aurora}. 
These studies indicate that Transformers, or Transformer-like operator processors, can act as general temporal integrators across complex physical systems, often at substantially reduced computational cost compared to traditional numerical pipelines.

In parallel, operator-style Transformer architectures have been developed specifically for scientific computing \cite{shih2025transformers}. 
Poseidon introduces a multiscale operator Transformer pretrained on diverse fluid-dynamics PDE datasets and leverages time-conditioned layers together with semigroup training to enable continuous-in-time evaluation~\cite{poseidon2024}. 
This places Transformers within the broader operator-learning lineage that includes Fourier- and Graph Neural Operators. 
Related theoretical work has clarified connections between attention mechanisms and classical numerical integration or projection schemes. 
Li et al.~\cite{li2020fourier, kovachki2023neuraloperators} introduced the Fourier Neural Operator framework, which learns mappings between function spaces using spectral convolution kernels and can be interpreted as performing data-driven Galerkin projections. 
Building on this perspective, Cao et al.~\cite{cao2021choose} showed that self-attention can be interpreted as a learnable integral operator, capable of recovering Fourier- or Galerkin-type behavior depending on positional encoding and kernelization. 
These results position attention mechanisms and neural operators within a shared theoretical space as flexible, possibly nonlocal (in space and even possibly in time) integrators/solvers.

A second, rapidly growing stream concerns time-series foundation models. 
Chronos treats time series as tokenized sequences via scaling and quantization and reuses T5-style Transformers to obtain zero-shot probabilistic forecasts across domains~\cite{chronos2024}. 
Subsequent models, including Chronos-Bolt, improved speed and accuracy, reinforcing the view that a single pretrained Transformer can generalize across dynamical regimes provided that the data are cast into a language-like format and extended the foundation models to multivariate systems \cite{ansari2025chronos2univariateuniversalforecasting}. 
This paradigm aligns closely with recent zero-shot and universal forecasting studies for chaotic systems~\cite{zhang2024zero, lai2025panda, hemmer2025true}, as well as with position papers calling for clearer definitions of what constitutes a foundation model for computational science~\cite{choi2025defining}.

Despite this growing body of work, the role of Transformers in modeling dynamical systems remains poorly understood from a theoretical standpoint, and to our knowledge, connections with first principles of autoregressive modeling and dynamical systems theory have not yet been firmly established.
Transformers do not maintain an internal representation that is explicitly interpreted as (i.e. matched with, mapped to) the physical state of the system. Instead, their internal embeddings are optimized for predictive performance and are not directly constrained to represent dynamical invariants such as attractors, invariant manifolds, or conserved quantities associated with Hamiltonian or symplectic structure. This contrasts with structure-preserving architectures, which enforce such physical constraints by design~\cite{Bertalan2019,lutter2019,HERNANDEZ2021109950,Bacsa2023}.
As a result, it remains unclear what classes of dynamical behavior Transformers can faithfully represent, under what conditions they succeed or fail, and how their internal computations relate to established concepts in nonlinear dynamics.

Recent studies have begun to address these gaps by probing not only forecasting accuracy but also the nature of the representations learned by Transformers. 
Kantamneni, Liu, and Tegmark analyzed how a Transformer models the simple harmonic oscillator, showing that attention induces a convex, data-driven autoregressive operator with characteristic spectral limitations~\cite{kantamneni2024transformers}. 
Related concerns arise in zero-shot dynamical studies, where long-term statistics may be preserved while the internal representation remains opaque~\cite{hemmer2025true}. 
At the same time, theoretical developments have revealed close connections between causal attention mechanisms, recurrence, and state–space structure. 
Katharopoulos et al.~\cite{katharopoulos2020transformers} showed that kernelized causal self-attention admits an exact recurrent formulation with constant memory, demonstrating that autoregressive Transformers with linear attention are, in a precise sense, recurrent neural networks. 
Dao et al.~\cite{dao2024ssd} generalized this insight through their structured state-space duality framework, proving that broad classes of causal attention mechanisms are equivalent to structured state–space models and that any kernelized attention admitting efficient recurrence must correspond to a state-space realization. 
Complementarily, Sieber et al.~\cite{sieber2024dsf} introduced what they called the "Dynamical Systems Framework", which reformulates attention mechanisms, state–space models, and recurrent neural networks within a unified recursive representation, allowing principled comparisons in terms of stability, expressivity, and state expansion.

These developments motivate interpreting Transformers used for dynamical systems modeling not merely as generic sequence-to-sequence regressors, but \textit{as data-adaptive state–space models}. Within this view, attention constructs and updates an implicit state from past observations, while subsequent feed-forward components approximate the local flow map governing state evolution. This architecture also invites a conceptual analogy with Backward Error Analysis: the "one-token-ahead" prediction of a Transformer parallels the "one-timestep-ahead" update of a numerical initial value solver with a fixed step~\cite{geneva2022transformers}, suggesting that the Transformer could, in principle, approximate what this framework terms the Inverse Modified Differential Equation (IMDE), a perturbed governing law whose discrete solution exactly matches the sampled data~\cite{Zhu2023ImplementationA}. In this work, we treat this correspondence as motivation rather than as a claim to be validated; our analyses instead focus on mechanisms that can be probed directly, namely attention as a constrained autoregressive operator and as an adaptive delay-embedding mechanism, and we return to the IMDE connection as a future direction in Section~\ref{sec:future_directions}.

To explicitly decode these learned algorithmic structures, we draw on mechanistic interpretability—an emerging field originally developed for large language models (LLMs)—which aims to reverse-engineer neural networks into discrete algorithms and resolve challenges such as superposition~\cite{elhage2022toy, elhage2021mathematical}. 
From this perspective, Transformers are viewed not as black boxes, but as collections of mechanistic circuits whose internal computations can be analyzed and related to classical modeling principles. 
In contrast with typical mechanistic interpretability studies of LLMs, we apply this lens to dynamical systems: we ask what a \emph{single-layer} Transformer captures about linear versus nonlinear dynamics, and whether its ability to ``unfold'' an attractor (as shown in Section~\ref{sec:nonlin_systems}) relies on attention behaving as a specific delay-embedding or state-aggregation mechanism. This bottom-up analysis complements recent top-down investigations of physics foundation models, where large models trained on physical simulations have been shown to encode physical features such as vorticity and diffusion as single
directions in activation space; adding or subtracting such direction during
inference induces or suppresses the corresponding physical behaviour in the
predicted dynamics, a property the authors term \emph{linear
steerability}~\cite{fear2025physics}.. We believe that understanding the mechanistic role of attention in minimal architectures could serve as a step toward bridging these concepts.

\section{Theory and Methods}

\subsection{Problem formulation}

We consider autonomous continuous-time dynamical systems of the form
\begin{equation}
    \frac{d\mathbf{s}}{dt}=f(\mathbf{s}(t)),
    \qquad
    \mathbf{s}(t)\in\mathbb{R}^{d_s},
\end{equation}
where $f:\mathbb{R}^{d_s}\to\mathbb{R}^{d_s}$ defines the system dynamics and $d_s$ is the full state dimension. Throughout this work, we focus on canonical dynamical systems exhibiting a range of properties, including limit cycles, parameter-dependent behavior, and partial-observation regimes. Our representative examples include the Van der Pol oscillator, reaction--diffusion PDEs, and the Navier--Stokes equations. We reserve problem-specific symbols for the physical variables of each example: for instance, $x(t)$ denotes displacement in the oscillator examples, $u(x,t)$ denotes the scalar Chafee-Infante field, and $\mathbf{v}(x,y,t)$ denotes the Navier--Stokes velocity field.

Sampling at a uniform interval $\Delta t$ gives a discrete trajectory
\begin{equation}
    \mathcal{D}=\{\mathbf{s}_1,\mathbf{s}_2,\ldots,\mathbf{s}_N\},
    \qquad
    \mathbf{s}_t=\mathbf{s}(t\Delta t),
\end{equation}
where $N$ is the number of sampled time points. The sampled dynamics induce the discrete-time flow map
\begin{equation}
    \mathbf{s}_{t+1}=F_{\Delta t}(\mathbf{s}_t),
\end{equation}
which is the object implicitly approximated by step-ahead prediction models, including numerical IVP solvers and Transformers.

In many practical settings, the full state $\mathbf{s}_t$ is not directly accessible. Instead, the model receives observations
\begin{equation}
    \mathbf{y}_t=g(\mathbf{s}_t)=\mathcal{H}\mathbf{s}_t,
    \qquad
    \mathbf{y}_t\in\mathbb{R}^{d_y},
\end{equation}
where $g:\mathbb{R}^{d_s}\to\mathbb{R}^{d_y}$ is the observation map, $d_y<d_s$ in the partial-observation setting, and $\mathcal{H}\in\mathbb{R}^{d_y\times d_s}$ denotes a linear observation operator when the measurements are linear. For example, in the Van der Pol oscillator one may observe only the position variable while the velocity remains unmeasured.

This partial-observability setting is central to nonlinear time-series analysis and motivates classical state-reconstruction approaches based on delay-coordinate embeddings. Takens' embedding theorem guarantees that, under suitable conditions, the attractor of the underlying dynamical system can be reconstructed from time-delayed measurements of partial observations~\cite{takens1981detecting}. A central question in this work is whether attention-based models implicitly perform an analogous reconstruction when trained on sequences of partial observations.

\paragraph{Research questions}
Our mechanistic study investigates connections between Transformer operations and classical dynamical systems formulations such as autoregressive modeling frameworks and delay-coordinate embeddings (Takens' theorem) for linear and nonlinear dynamical systems.

By training single-layer Transformers to predict the evolution of dynamical systems, we investigate:

\emph{A) For linear systems} (Section~\ref{sec:lin_systems}), where the attention mechanism approximates the dynamics directly, we ask: (A1) what does attention learn and how does this connect to classical linear system theory; (A2) what classes of dynamics (e.g., monotonic, resonant, or underdamped) can attention represent, and for which does it fail and why; and (A3) when does attention meaningfully capture multi-modal interactions, and how does this connect to delay-coordinate embeddings?

\emph{B) For nonlinear systems} (Section~\ref{sec:nonlin_systems}), where attention no longer approximates the dynamics itself but instead serves as a state reconstruction operator upon which a nonlinear map is learned, we ask: (B1) under what observation regimes---full-state versus partial---does a Transformer provide computational benefits; (B2) can attention identify representations analogous to delay-coordinate embeddings, and how does this relate to classical results such as Takens' embedding theorem; and (B3) when such delay-based representations are formed, under what conditions do they suffice to (a) capture the nonlinear attractors while preserving their effective dimensionality; (b) capture the dominant modes of the underlying dynamics; and (c) provide meaningful organization of trajectories across different system parameters?

By restricting our analysis to single-layer architectures, and addressing those questions we aim to develop interpretable insights into the fundamental mechanisms through which Transformers process temporal dynamics. The findings provide a foundation for understanding deeper architectures and we hope they can serve as a guide for designing attention-based models tailored to dynamical systems.

\subsection{Single-layer Transformer as a Discrete-Time Operator}

To analyze how attention-based models process dynamical information, we restrict our experiments to a canonical single-head, single-layer self-attention (decoder-only) Transformer architecture. Given an input sequence of $n$ tokens,
\begin{equation}
    Y_t=[\mathbf{y}_{t-n+1},\ldots,\mathbf{y}_{t}]^\top
    \in \mathbb{R}^{n\times d_y},
\end{equation}
representing system observations over a finite time window, we first form the embedded model input
\begin{equation}
    E_t=Y_tW_{\mathrm{emb}}+P,
    \qquad
    E_t\in\mathbb{R}^{n\times d_{\mathrm{lat}}},
    \label{eq:embedding_pe}
\end{equation}
where $E_t$ denotes the token matrix passed to the attention layer, $W_{\mathrm{emb}}\in\mathbb{R}^{d_y\times d_{\mathrm{lat}}}$ is a learned input embedding, and $P\in\mathbb{R}^{n\times d_{\mathrm{lat}}}$ is an optional learned positional encoding (PE). When no separate embedding is used, $W_{\mathrm{emb}}$ may be identified with the identity map; when PE is absent, $P=0$.

The model then applies a self-attention operation. We first form the attention matrix
\begin{equation}
    A_t = \mathrm{softmax}\!\left(\frac{Q_tK_t^\top}{\sqrt{d_k}}\right) \in \mathbb{R}^{n\times n},
    \label{eq:attention_matrix}
\end{equation}
where
\begin{equation}
    Q_t=E_tW_Q,\qquad K_t=E_tW_K,\qquad V_t=E_tW_V
\end{equation}
are learned linear projections, referred to as the \emph{query}, \emph{key}, and \emph{value}, respectively. The scalar $d_k$ is the dimension of the query/key projections and acts as a normalization factor that stabilizes the softmax gradients. Owing to the row-wise softmax, each row of $A_t$ is non-negative and sums to unity. The attention output is then
\begin{equation}
    Z_t = \mathrm{Attention}(Q_t,K_t,V_t) = A_t V_t.
    \label{eq:attention_def}
\end{equation}

The resulting representation $Z_t\in\mathbb{R}^{n\times d_v}$ collects, row by row, the attention outputs for each input token; we denote by
\begin{equation}
    \mathbf{z}_t=(Z_t)_{n,:}
\end{equation}
the row associated with the final, most recent token. In the Transformer block used here, the quantity fed to the output layers is the residual combination of the final token representation and the final-token attention output. We denote this effective latent coordinate by
\begin{equation}
    \mathbf{h}_t=\mathbf{e}_t+\mathbf{z}_t,
    \qquad
    \mathbf{e}_t=(E_t)_{n,:}.
    \label{eq:effective_latent}
\end{equation}
In purely attention-only ablations designed to isolate the attention contribution, the residual contribution is disabled; equivalently, $\mathbf{e}_t=0$ and $\mathbf{h}_t=\mathbf{z}_t$. In the single-head setting considered here, we take the query, key, and value projection dimensions to coincide with the latent dimension, $d_k=d_v=d_{\mathrm{lat}}$, so that $W_Q,W_K,W_V\in\mathbb{R}^{d_{\mathrm{lat}}\times d_{\mathrm{lat}}}$ and the attention output $\mathbf{z}_t$ shares the latent space of $\mathbf{e}_t$, making the residual sum in Eq.~\eqref{eq:effective_latent} well-defined.

\begin{figure}[ht]
    \centering
    \includegraphics[width=\linewidth]{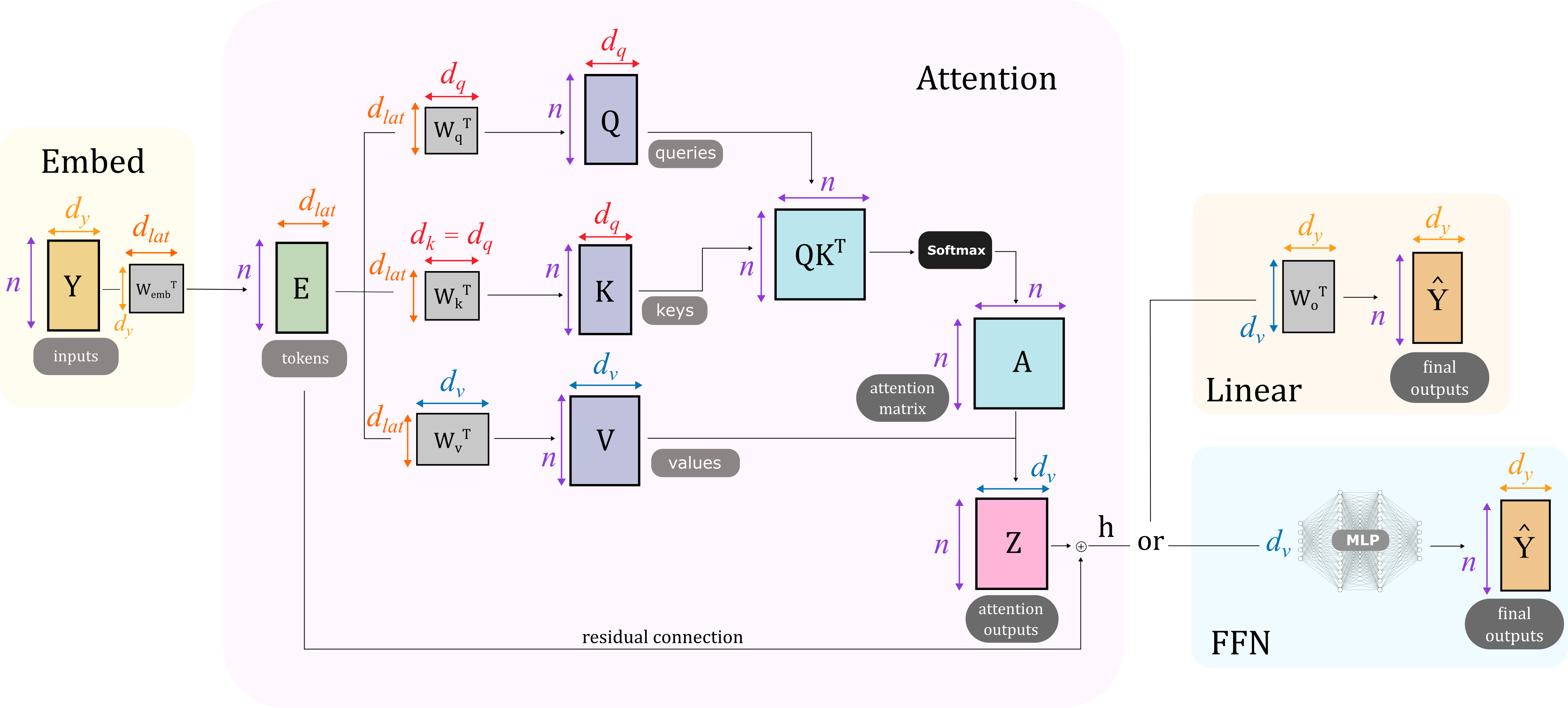}
    \caption{Schematic of the single-layer single-head self-attention Transformer architecture with optional linear output studied in this work. The optional positional encoding $P$ is omitted from the schematic. Adapted from \cite{raschka2023understanding}.}
    \label{fig:transformer_arch}
\end{figure}

The representation $\mathbf{h}_t$ is then processed by an output layer, which takes one of two forms depending on the analysis. For nonlinear dynamics, we employ the position-wise feed-forward Multi-Layer Perceptron (MLP) typical for Transformers, applied to the effective latent coordinate,
\begin{equation}
    \hat{\mathbf{y}}_{t+1}
    =\mathrm{MLP}(\mathbf{h}_t)
    =\sigma(\mathbf{h}_tW_1+b_1)W_2+b_2,
\end{equation}
where $\sigma(\cdot)$ denotes the activation function, $W_1$ and $W_2$ are learnable weight matrices, and $b_1$ and $b_2$ are learnable bias terms. For linear dynamics, we instead use a linear output projection,
\begin{equation}
    \hat{\mathbf{y}}_{t+1}=\mathbf{h}_tW_O.
\end{equation}
The MLP serves as a universal approximator for nonlinear state-transition functions, while the linear projection isolates the representational capacity of attention together with the residual readout coordinate. A schematic of the full process is shown in Figure~\ref{fig:transformer_arch}.

The presence or absence of positional encoding is a recurring experimental feature throughout this work. We use the terms ``with PE'' and ``without PE'' consistently to distinguish whether $P$ in Eq.~\eqref{eq:embedding_pe} is learned or set to zero. Positional encoding changes whether the attention operator can distinguish the order of the delays, whereas the context length $n$ controls how many input tokens, equivalently delayed observations, are available for state reconstruction. By convention, row $1$ of $Y_t$ is the oldest observation in the window and row $n$ is the most recent observation; the prediction target for an input window ending at time $t$ is $\mathbf{y}_{t+1}$.

By analyzing this minimal Transformer layer, we can isolate the specific dynamical roles of these components: (a) the attention mechanism functions as a temporal aggregator that navigates the trajectory history, and (b) the MLP serves as a universal approximator for the local state-transition function. In the step-ahead prediction setting considered here, the Transformer defines an operator that maps an implicitly reconstructed state, formed through delayed observations, rather than using the instantaneous state itself. From a dynamical systems perspective, this can be interpreted as a learned discrete-time evolution operator acting on an implicitly constructed state representation.

\subsection{Classical dynamical system formulations and recursive representations} \label{sec:dsf}

Classical data-driven modeling of dynamical systems is grounded in recursive representations, most prominently autoregressive (AR) and state--space formulations. In linear autoregressive models, the current observation is expressed as a finite-memory recursion over past outputs,
\begin{equation}
    \mathbf{y}_{t}=\sum_{\ell=1}^{p_{\mathrm{AR}}}M_{\ell}^{\mathrm{AR}}\mathbf{y}_{t-\ell}+\boldsymbol{\varepsilon}_t,
    \label{eq:ar_model}
\end{equation}
where $p_{\mathrm{AR}}$ is the AR order, $M_{\ell}^{\mathrm{AR}}$ are lag operators, and $\boldsymbol{\varepsilon}_t$ denotes process noise. Such models admit well-characterized notions of stability, identifiability, and spectral structure, and have been widely used in system identification and structural dynamics~\cite{box1976analysis,Kantz_Schreiber_2003}.

A more expressive and principled formulation is obtained through state--space models, which introduce an explicit latent state $\boldsymbol{\zeta}_t$ evolving recursively as
\begin{align}
    \boldsymbol{\zeta}_{t+1} &= A_{\mathrm{ss}}\boldsymbol{\zeta}_t+B_{\mathrm{ss}}\mathbf{y}_t+\mathbf{w}_t,\\
    \boldsymbol{\chi}_{t} &= C_{\mathrm{ss}}\boldsymbol{\zeta}_t+\boldsymbol{\eta}_t,
\end{align}
with $\boldsymbol{\chi}_{t}$ the output at time $t$, and $\mathbf{w}_t$ and $\boldsymbol{\eta}_t$ denoting process and observation noise, respectively. Here, the latent state is typically endowed with physical meaning (e.g.\ displacements, velocities, modal coordinates, or internal variables), and recursive filtering schemes, such as the Kalman filter, provide optimal state estimates under linear--Gaussian assumptions. This paradigm embodies the classical principle of \emph{latent-state reconstruction}, where memory and dynamics are encoded through a compact, physically interpretable state.

From a unifying viewpoint, autoregressive and state--space models are both instances of recursive dynamical systems. Classical autoregressive models admit equivalent state--space realizations by defining the latent state as a stack of delayed outputs and parameters (inputs, when available), a construction standard in system identification and control~\cite{ljung1999systemid}.

Recently, Sieber et al.~\cite{sieber2024dsf} introduced the Dynamical Systems Framework (DSF), which represents attention mechanisms, state--space models, and recurrent neural networks within a common linear time-varying recurrence. Within this framework, masked self-attention admits an exact recursive realization where the effective transition operators are determined by the query--key interactions and normalization terms.

\paragraph{Attention as a data-adaptive autoregressive model}\label{sec:AR}
For our purposes, the essential insight is that a single attention head induces a finite-memory linear recursion interpretable as a data-driven AR model. Consider the final row of the attention matrix, indexed by lag,
$\boldsymbol{\alpha}_t=[\alpha_{t,1},\ldots,\alpha_{t,n}]^\top$, with $\alpha_{t,\ell}>0$ and $\sum_{\ell=1}^{n}\alpha_{t,\ell}=1$. The final-token attention output can be written as
\begin{equation}
    \mathbf{z}_t
    =\sum_{\ell=1}^{n}\alpha_{t,\ell}\,\mathbf{e}_{t,\ell}W_V,
\end{equation}
where $\mathbf{e}_{t,\ell}$ denotes the row of $E_t$ associated with lag $\ell$. Incorporating the output and value projection matrices, and recalling that the output layer receives the effective latent coordinate $\mathbf{h}_t=\mathbf{e}_t+\mathbf{z}_t$, a linear readout gives
\begin{equation}
    \hat{\mathbf{y}}_{t+1}
    =\mathbf{h}_tW_O
    =\mathbf{e}_tW_O+
    \sum_{\ell=1}^{n}\alpha_{t,\ell}\,\mathbf{e}_{t,\ell}M,
    \qquad
    M=W_VW_O.
    \label{eq:attention_ar}
\end{equation}
This mirrors an AR model whose coefficients are determined by the attention weights, augmented by the direct residual contribution from the final token. In the residual-free attention-only ablation, $\mathbf{e}_t=0$, and the effective lag operator is $M_{t,\ell}^{\mathrm{eff}}=\alpha_{t,\ell}M$. However, due to the softmax normalization in attention computation, all $\alpha_{t,\ell}$ are strictly non-negative and sum to unity, implying that the attention-induced lag operators are non-negative scalar multiples of the same matrix $M$.

This non-negativity constraint introduces a fundamental representational limitation. Unlike conventional AR models, which can employ both positive and negative coefficients to represent oscillatory or phase-inverted dynamics, a single softmax attention head cannot directly encode subtractive interactions. Consequently, systems requiring mixed-sign autoregressive dependencies, such as lightly damped oscillators, may not be faithfully represented by a single attention-only head. This simple theoretical insight provides the basis for our empirical analyses in Section~\ref{sec:lin_systems}, where we demonstrate both successful and failed cases depending on the sign structure of the underlying linear dynamics.


\paragraph{Nonlinear systems and delay-coordinate embeddings}
For nonlinear systems, state reconstruction from partial observations is classically addressed through delay-coordinate embeddings. Takens' embedding theorem~\cite{takens1981detecting} establishes that for a smooth, deterministic dynamical system evolving on a compact attractor of dimension $d_A$, and for a generic scalar observation function, the delay map
\begin{equation}
    \Psi_{n,\tau}(\mathbf{s}(t))
    =\big[g(\mathbf{s}(t)),\,g(\mathbf{s}(t-\tau)),\,\ldots,\,g(\mathbf{s}(t-(n-1)\tau))\big]
    \label{eq:delay_map}
\end{equation}
constitutes an embedding of the attractor provided that $n\ge 2d_A+1$. This result formalizes the conditions under which the latent state of a nonlinear system can be reconstructed from time-delayed measurements alone, even under partial observability.

From this perspective, attention mechanisms can be interpreted as constructing adaptive, data-driven delay embeddings by aggregating information from a finite history of past observations. Crucially, while the attention operation preserves linearity, the Transformer's feed-forward network introduces the capacity to approximate nonlinear maps on the reconstructed coordinates. This decoupling of \textit{linear history aggregation} from \textit{nonlinear state evolution} provides the necessary bridge to the nonlinear analyses presented in Section~\ref{sec:nonlin_systems}.

\subsection{Scope, analysis and organization}

The formulations above establish two complementary viewpoints on data-driven dynamical modeling. 
Classical autoregressive, state–space, and delay-embedding approaches rely on either explicit latent states or geometrically justified state reconstructions, with well-defined notions of memory, minimality, and observability. 
By contrast, attention-based Transformers trained for step-ahead prediction induce a learnt recursive representation in which the effective state is constructed deterministically from a finite history of observations through data-adaptive aggregation.

The analysis that follows examines how this architectural distinction manifests in practice. 
We study the internal computations of a single-layer Transformer trained on representative dynamical systems and investigate how attention weights relate to phase-space geometry, how implicit state representations emerge under partial observability, and how the learned dynamics compare to classical autoregressive and delay-based models. 
Particular emphasis is placed on identifying structural constraints imposed by finite context length, token-based state construction, and attention normalization, and on understanding how these constraints affect the model’s ability to represent periodic, quasi-periodic, and chaotic behavior.

The subsequent sections are organized as follows. 
We first analyze linear and weakly nonlinear systems to characterize the effective autoregressive structure induced by attention and its spectral properties. 
We then consider nonlinear systems with partial observations, examining whether attention mechanisms recover embeddings consistent with classical delay-coordinate constructions. 
Finally, we investigate regimes in which the single-layer architecture fails to capture essential dynamical features, thereby delineating fundamental limitations that persist independently of training data or optimization.

This organization allows us to connect mechanistic observations of Transformer behavior directly to established concepts in dynamical systems theory, and to assess the extent to which attention-based models can be interpreted as \textit{data-adaptive }realizations of classical recursive dynamical representations.

\section{Linear Dynamical Systems}
\label{sec:lin_systems}
We begin our analysis with linear dynamical systems, the simplest and most foundational class where analytical insights are most tractable. By isolating the attention mechanism and excluding the feed-forward network, the Transformer reduces to a linear, time-varying recursive operator that admits a direct interpretation as introduced in Section~\ref{sec:dsf}. This allows us to establish a baseline understanding and derive closed-form expressions for the learned representations. Linear systems serve as an ideal starting point because their mathematical structure is well understood and admits explicit classical representations, thus allowing us to precisely characterize what the attention mechanism computes and how it relates to classical linear system theory.

\subsection{Single-DOF Structural System}

\paragraph{Dynamical System}

To connect the abstract discussion of attention as autoregression in Section~\ref{sec:AR} with a physically interpretable example, we now consider a  single-degree-of-freedom (SDOF) linear oscillator, whose discrete-time dynamics admit an exact low-order autoregressive representation. 
This setting allows us to explicitly compare the coefficients of the physical AR model with the effective coefficients induced by a single attention head, and to interpret the results directly. 
By focusing on an attention-only architecture, we isolate the linear operator induced by self-attention and assess when it can, and when it cannot, reproduce the signed recursive structure of the underlying dynamics.
The governing equation of motion of the considered second-order SDOF system is given by:
\begin{equation}
m\ddot{x}(t) + c_d\dot{x}(t) + kx(t) = F_\mathrm{ext}(t),
\end{equation}
where \(m\), \(c_d\), and \(k\) denote the mass, damping coefficient, and stiffness, respectively. Only the displacement response \(x\) is assumed to be available. In the present study, we fix \(m = 1~\mathrm{kg}\) and \(c_d = 0.5~\mathrm{Ns/m}\), corresponding to a lightly damped (underdamped) regime. For an initial stiffness of \(k = 2000~\mathrm{N/m}\), which gives a natural frequency of:
\begin{equation}
f_n = \frac{1}{2\pi}\sqrt{\frac{k}{m}} \approx 7.12~\text{Hz}.
\end{equation}
A sampling frequency of \(25~\text{Hz}\) (\(\Delta t = 0.04~\text{s}\)) ensures adequate temporal resolution satisfying the Nyquist criterion.

A free vibration case is considered in this study, $F_\mathrm{ext}(t) =0 $, where the system is initialized with a displacement of \(x_0 = 10~\mathrm{mm}\) and zero velocity, allowing the response to evolve solely under its internal dynamics. The governing second-order differential equation is numerically integrated using a high-accuracy ODE solver. The displacement response \(x(t)\) is recorded and discretized to form sequential time series data used for model training and evaluation. The predictive task is \textit{one-step-ahead forecasting}—that is, predicting the displacement \(x_{t+1}\) from a short history \(\{x_t, x_{t-1}, \ldots, x_{t-n+1}\}\). This formulation directly parallels the autoregressive structure discussed previously and allows direct comparison between the physical AR(2) system (i.e., an autoregressive model with two-step history, $p_{\mathrm{AR}}=2$ in Eq.~\eqref{eq:ar_model}) and its attention-based approximation.

\paragraph{Transformer Setup}

A minimal attention-only Transformer is employed to assess whether self-attention can recover the inherent oscillatory structure of the SDOF system. To isolate the attention-induced autoregressive coefficients, this ablation disables the residual contribution, so the readout input is $\mathbf{h}=\mathbf{z}$. The model operates on scalar input sequences \(x_i\) with PEs \(p_i\), given that no embedding layers are used.  The attention transformation can thus be expressed analytically for a two-step input history as:
\[
\alpha_{ij} = \mathrm{softmax}_j\!\left(w_q w_k\,(\tilde{x}_i \cdot \tilde{x}_j)\right), \quad \text{where } \tilde{x}_i = x_i + p_i,
\]
After softmax normalization, the predicted output takes the form:
\[
\hat{x} = w_o\left[\alpha_{2,1}(x_1 + p_1) + \alpha_{2,2}(x_2 + p_2)\right],
\]
which can be rearranged as:
\[
\hat{x} = \beta_1 x_1 + \beta_2 x_2 + b_{\mathrm{PE}},
\]
with \(\beta_i=\alpha_{2,i}w_o\) and \(b_{\mathrm{PE}}=\beta_1p_1+\beta_2p_2\).  
This formulation mirrors an AR(2) process \(x_{t+1}=a_1x_t+a_2x_{t-1}\), but \textit{with the key restriction that attention weights \(\alpha_{2,i}\) are positive and normalized, enforcing convexity in the combination of past inputs}.

\paragraph{Results and Observations}

\emph{Case 1: \(k = 2000~\mathrm{N/m}\)}  

For the first configuration, the discrete-time AR(2) coefficients derived from the physical SDOF model are \(a_1=-0.4352\) and \(a_2=-0.9802\), both negative. Since the coefficients share the same sign, the attention-based model can emulate the dynamics by adopting positive attention weights and a negative scalar output projection weight (\(w_o < 0\)). Under these conditions, the Transformer successfully captures the oscillatory characteristics of the system.

The spectrum is computed by interpreting the attention-induced linear recurrence as an AR model and evaluating its frequency response; in the figures we report the magnitude $|H(\omega)|$ of this discrete-time transfer function, i.e.\ the gain that the learned recurrence applies to a unit-amplitude harmonic input at angular frequency $\omega$. For a linear oscillatory system, an optimally fitted AR model exhibits spectral peaks of $|H(\omega)|$ at the system’s natural frequencies.

\begin{figure}[t]
\centering

\begin{subfigure}[b]{\linewidth}
    \centering
    \includegraphics[width=\linewidth]{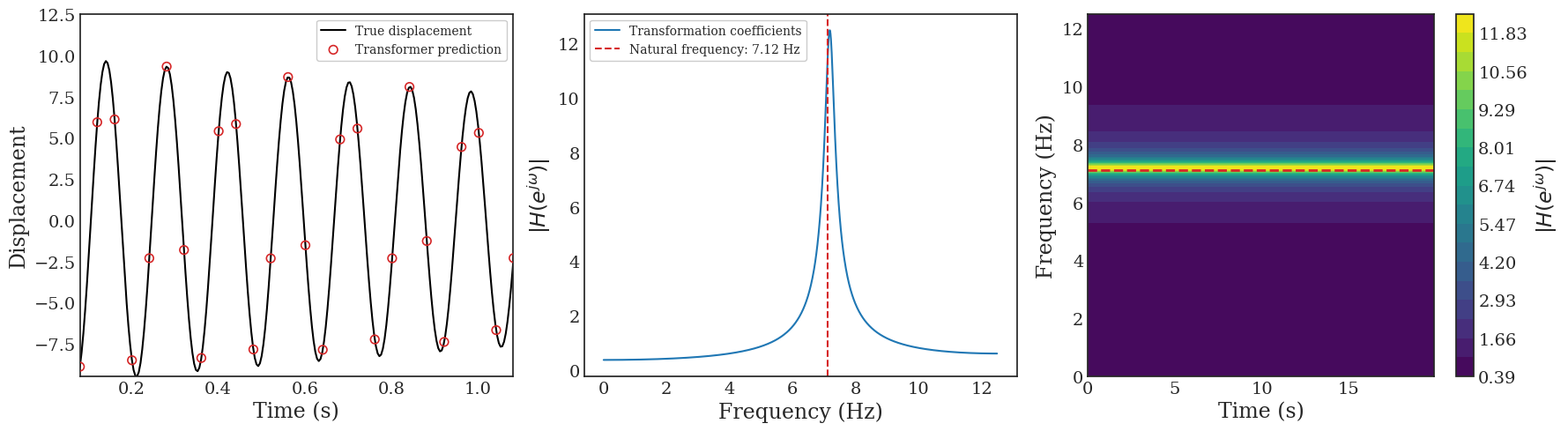}
    \caption{}
    \label{fig:case1}
\end{subfigure}

\vspace{0.8em}

\begin{subfigure}[b]{\linewidth}
    \centering
    \includegraphics[width=\linewidth]{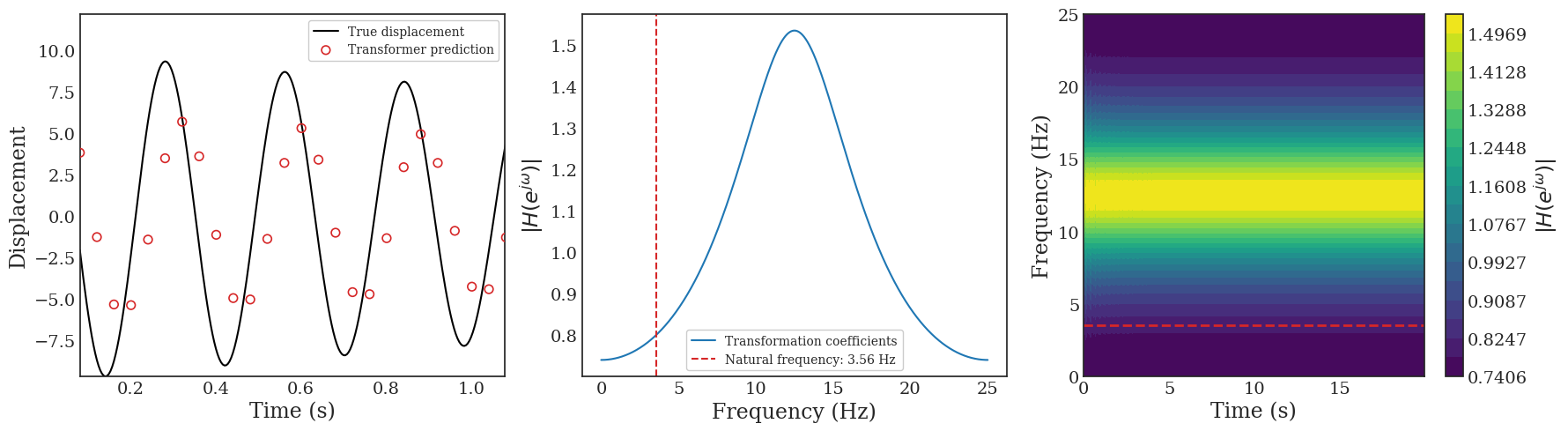}
    \caption{}
    \label{fig:case2}
\end{subfigure}

\caption{Transformer performance on linear SDOF systems. 
\textbf{(a)} Successful reproduction of oscillatory dynamics for \(k=2000~\mathrm{N/m}\) with same-sign AR coefficients: the Transformer accurately captures amplitude and phase, and preserves both the dominant modal peak in the frequency domain and the coherent modal decay ridge in the time--frequency representation. 
\textbf{(b)} Failure case for \(k=500~\mathrm{N/m}\) with mixed-sign AR coefficients: the attention-only model produces over-smoothed responses and fails to recover the resonance peak and coherent modal ridge.}
\label{fig:linear}
\end{figure}

As shown in Figure~\ref{fig:case1}, the predicted displacement sequence accurately reproduces the oscillatory motion in the time domain. The corresponding frequency-domain representation exhibits a clear peak near the analytical natural frequency (\(7.12~\text{Hz}\)), indicating that the learned dynamics captures the dominant vibration mode. The time--frequency spectrogram on the right further confirms this by showing a coherent concentration on the natural frequency. Together, these results validate that the learned dynamics capture both the correct oscillatory behavior and the damping characteristics of the system. The correspondence confirms that, when the target dynamics can be expressed as a \textit{convex weighted combination} of past states, the attention operator can emulate the underlying physical process with high fidelity.

\emph{Case 2: \(k = 500~\mathrm{N/m}\)}  

Reducing the stiffness to \(k = 500~\mathrm{N/m}\) lowers the natural frequency to approximately \(3.56~\text{Hz}\) and alters the discrete-time AR(2) coefficients to mixed signs, requiring \(a_1>0\) and \(a_2<0\). To reproduce this structure, the attention model would need to satisfy:
\[
\alpha_{2,1}w_o=a_1>0, \quad \alpha_{2,2}w_o=a_2<0.
\]
However, since \(\alpha_{2,i}\) are strictly positive due to softmax normalization, no real-valued \(w_o\) can simultaneously satisfy these conditions. The model is thus incapable of representing the required \textit{subtractive relationship} between consecutive time steps that gives rise to oscillatory or resonant behavior.

As shown in Figure~\ref{fig:case2}, the predicted trajectory remains visually oscillatory but does not reproduce the correct dynamical signature. The corresponding frequency-domain representation lacks a distinct resonance peak at the physical natural frequency, and the time--frequency spectrogram does not recover the coherent modal decay ridge.

This experiment highlights a fundamental limitation of the attention mechanism as a linear dynamical operator: due to the \textit{non-negativity constraint} imposed by the softmax function, attention cannot reproduce the signed coefficient patterns essential for modeling oscillatory systems with phase-alternating behavior. 

This observation is specific to the single-head attention-only setting used here to expose the underlying mechanism, and should not be interpreted as a general limitation of practical Transformer architectures. As shown in \ref{sec:two_head_sdof_appendix}, a two-head attention-only model is capable of recovering the Case 2 response by combining the two head outputs with output projection coefficients of opposite signs, thereby representing the mixed-sign autoregressive behavior required by the system.

\subsection{Extension to 2-DOF Systems}

\paragraph{Dynamical System}
To examine the scalability of attention-based architectures beyond single-mode dynamics, we next consider a two-degree-of-freedom (2-DOF) linear structural system with coupled masses. This system introduces modal interaction and a higher-dimensional state-space, offering a more stringent test for the Transformer's ability to capture multi-modal dependencies.
The 2-DOF system provides a minimal setting in which the effective state dimension must increase to encode multiple interacting modes. Unlike the single-DOF case, the latent dynamics now require the representation of at least a second-order linear recurrence per mode, together with cross-coupling terms. This makes the role of temporal context and observation richness explicit: the attention-induced state must either grow in dimension or compensate through longer memory in order to remain expressive.

The governing equation in this 2-DOF system, comprising a coupled system of two second-order ODEs, is expressed as:
\begin{equation}
\mathbfcal{M}\ddot{\mathbf{x}}(t) + \mathbf{C}_{\mathrm{damp}}\dot{\mathbf{x}}(t) + \mathbf{K}\mathbf{x}(t) = \mathbf{0},
\end{equation}
where $\mathbf{x}(t) = [x_1(t), x_2(t)]^\top$ denotes the displacement vector, and the system matrices are defined as:
\[
\mathbfcal{M} = 
\begin{bmatrix}
m_1 & 0 \\ 
0 & m_2
\end{bmatrix}, \quad
\mathbf{C}_{\mathrm{damp}} = 
\begin{bmatrix}
c_{d,1} + c_{d,2} & -c_{d,2} \\ 
-c_{d,2} & c_{d,2}
\end{bmatrix}, \quad
\mathbf{K} = 
\begin{bmatrix}
k_1 + k_2 & -k_2 \\ 
-k_2 & k_2
\end{bmatrix}.
\]
Although the system is second-order in time, numerical integration requires recasting it as a first-order system with displacements and velocities as state variables, yielding a four-dimensional state space. In our setting, only the displacement responses are observed. We adopt the parameters $m_1 = m_2 = 1.0$~kg, $c_{d,1}=c_{d,2}=0.5$~Ns/m, $k_1 = 1000$~N/m, and $k_2 = 1500$~N/m. Solving the undamped eigenvalue problem $\mathbf{K}\boldsymbol{\phi} = \omega^2 \mathbfcal{M}\boldsymbol{\phi}$ yields two natural frequencies:
\[
f_1 \approx 4.1~\text{Hz}, \quad f_2 \approx 9.5~\text{Hz}.
\]
Free vibration is simulated by initializing the first mass with a displacement perturbation while keeping all other states at rest:
\[
\mathbf{x}(0) = \begin{bmatrix} 10.0 \\ 0.0 \end{bmatrix}, \quad 
\dot{\mathbf{x}}(0) = \begin{bmatrix} 0.0 \\ 0.0 \end{bmatrix}.
\]
This excitation activates both modes and allows observation of their natural decay through damping. The "ground truth" system response is obtained by numerical integration using a Runge–Kutta solver (RK45) with relative and absolute tolerances set to $1e-10$ and $1e-12$, respectively, and displacement histories are recorded for subsequent model training and evaluation.

\paragraph{Transformer Setup}
The Transformer model receives vector-valued inputs $\mathbf{x}_t \in \mathbb{R}^2$ across consecutive time steps, serving as historical observations for one-step-ahead prediction. Both full and partial observation scenarios are examined:
\begin{itemize}
    \item \emph{Full displacement observation:} both displacement channels $(x_1, x_2)$ are available.
    \item \emph{Partial displacement observation:} only a single displacement, $x_1$, is provided.
\end{itemize}
In each setting, the number of input tokens $n$ is varied to assess the model’s ability to reconstruct modal information from limited temporal and spatial cues. 
Increasing $n$ effectively enlarges the span of the induced recursive state, allowing the attention mechanism to approximate higher-dimensional linear dynamics through delayed aggregation. In this sense, the context window plays a role analogous to state augmentation in classical state--space realizations of autoregressive models.

\paragraph{Results and Observations}

\begin{figure}[t]
\centering
\begin{subfigure}[b]{0.75\linewidth}
    \centering
    \includegraphics[width=\linewidth]{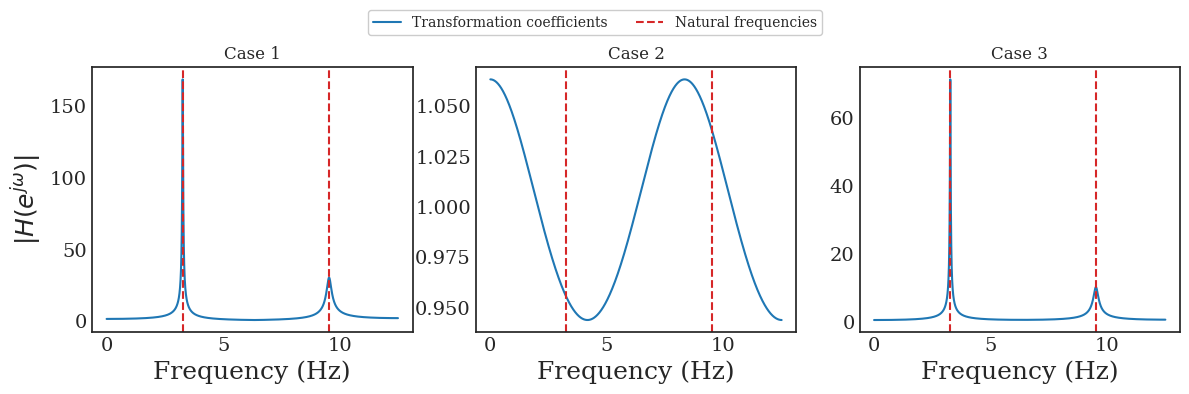}
    \caption{}
    \label{fig:2D_spectrum}
\end{subfigure}
\hfill
\begin{subfigure}[b]{0.24\linewidth}
    \centering
    \includegraphics[width=\linewidth]{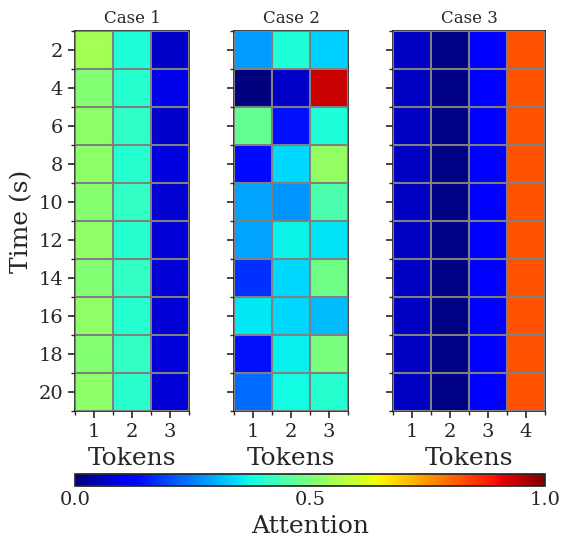}
    \caption{}
    \label{fig:2D_attention}
\end{subfigure}

\caption{(a) Spectral analysis of Transformer-predicted responses for the 2-DOF linear system under different observability and temporal context settings.
(b) Attention heatmaps of the Transformer models.}
\label{fig:2D_modal_spectra}
\end{figure}

\begin{figure}[t]
\centering
\begin{subfigure}[t]{0.49\linewidth}
    \centering
    \includegraphics[width=\linewidth]{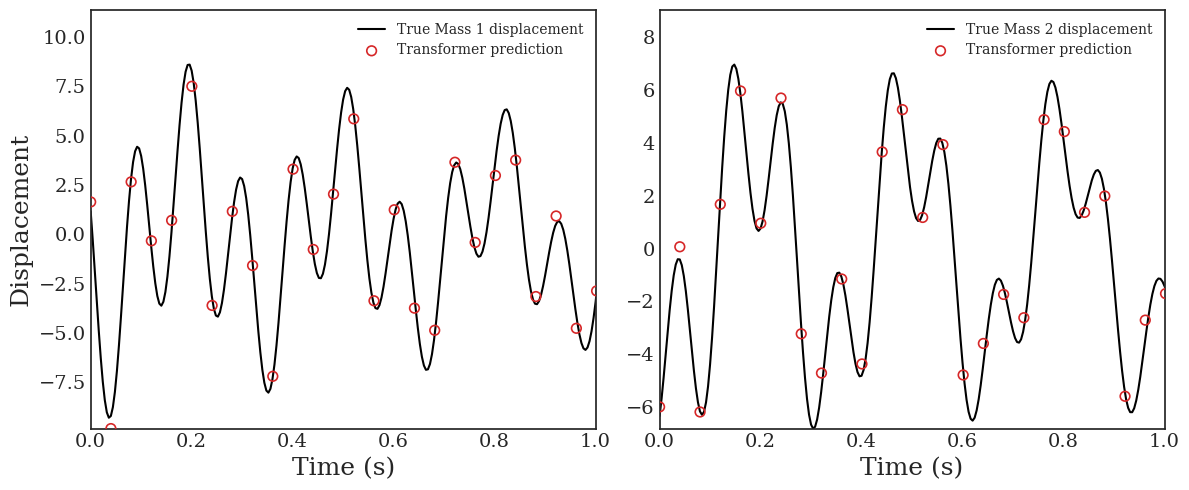}
    \caption{Case 1}
    \label{fig:2D_case1}
\end{subfigure}
\hfill
\begin{subfigure}[t]{0.245\linewidth}
    \centering
    \includegraphics[width=\linewidth]{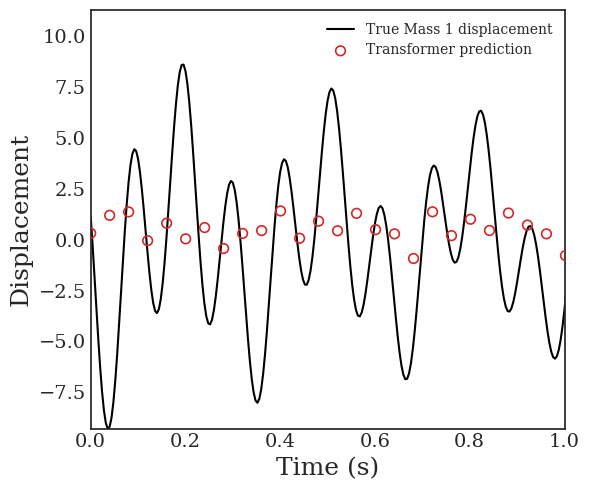}
    \caption{Case 2}
    \label{fig:2D_case2}
\end{subfigure}
\hfill
\begin{subfigure}[t]{0.245\linewidth}
    \centering
    \includegraphics[width=\linewidth]{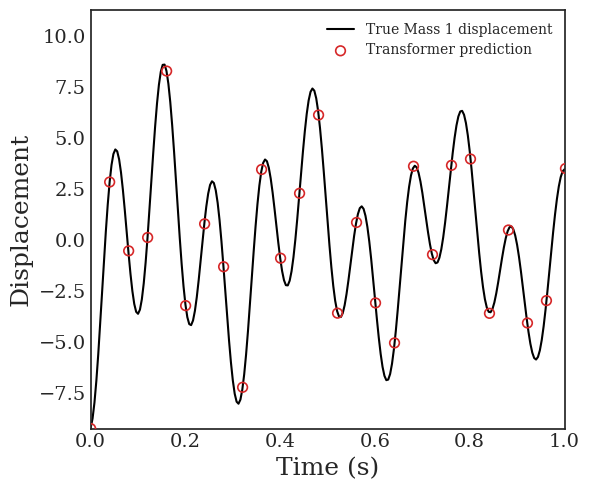}
    \caption{Case 3}
    \label{fig:2D_case3}
\end{subfigure}
\caption{Time-domain prediction results for the 2-DOF linear system.}
\label{fig:2D_trajectory}
\end{figure}

\emph{Case 1: Full Displacement Observation ($n=3$ input tokens).}
With both displacement channels available, the Transformer successfully identifies the two inherent modal frequencies of the system. In the first panel of Figure~\ref{fig:2D_spectrum} we can observe that the spectral analysis of the predicted sequences exhibits distinct peaks at 4.1~Hz and 9.5~Hz, closely matching the analytical modal frequencies. The corresponding time-domain trajectories in Figure~\ref{fig:2D_case1} further show that the model accurately predicts the responses of both DOFs. This confirms that the attention mechanism can approximate the underlying coupled dynamics when complete state information is available. The recovery of the correct modal frequencies indicates that the attention-induced linear recurrence implicitly approximates the underlying state evolution operator. For linear systems, modal frequencies are directly linked to the eigenvalues of the discrete-time state transition matrix.

\emph{Case 2: Partial Displacement Observation ($n=3$ input tokens).}
When only $x_1(t)$ is observed, the Transformer fails to reconstruct the coupled modal structure. As shown in the second panel of Figure~\ref{fig:2D_spectrum}, the resulting spectral density is dominated by a low-frequency component and lacks distinct peaks at the true modal frequencies. The time-domain comparison in Figure~\ref{fig:2D_case2} shows the same failure mode: the predicted trajectory deviates from the true response, indicating that the model does not adequately capture the underlying dynamics. This degradation arises because the model no longer receives spatial coupling information through $x_2(t)$, while the short temporal context prevents it from inferring intermodal interactions from temporal correlations alone. In essence, partial observability \textit{combined with limited temporal context} yields insufficient information for the attention mechanism to reconstruct the effective state needed to represent intermodal coupling.

\emph{Case 3: Partial Displacement Observation ($n=4$ input tokens).}
Extending the temporal window to four delayed observations restores the model's ability to identify both modal frequencies, even under partial observation. As seen in the third panel of Figure~\ref{fig:2D_spectrum}, the spectrum again exhibits clear peaks at the correct modal frequencies. Correspondingly, Figure~\ref{fig:2D_case3} shows that the model accurately predicts the response from the delayed observations of the first DOF. The longer temporal context allows the Transformer to implicitly capture delayed intermodal correlations that would otherwise require direct spatial measurements. This highlights a fundamental trade-off between spatial observability and temporal context in attention-based modeling of dynamical systems, directly analogous to the classical delay-coordinate embedding principle where temporal history compensates for missing state variables.

The attention maps in Figure \ref{fig:2D_attention} provide additional insight into the role of temporal context. In the full-observation case with $n=3$ input tokens, the attention pattern is relatively stable over time, suggesting that the model learns a consistent use of the available displacement history. The partial-observation case with $n=3$ input tokens behaves differently. Its attention map exhibits rapid switching of attention among the available lags, therefore suggesting that the model has not learned a stable delay-coordinate representation of the coupled state. Instead, with only a short scalar history, attention appears to select among delayed samples in a phase-dependent manner. When the partial-observation window is increased to $n=4$ input tokens, the attention becomes more structured across a broader set of delays. This longer scalar history enables the model to form a richer implicit delay representation.

In summary, these results demonstrate that Transformer-based representations can recover multi-modal linear dynamics when provided with sufficient information--either through direct access to spatial degrees of freedom or through extended temporal context that enables implicit state reconstruction.

\section{Nonlinear Systems and Sparse Observations}
\label{sec:nonlin_systems}
The analytical approach used for linear systems in the previous section, comparing discrete-time AR coefficients with the representations learned by attention, does not directly extend to nonlinear dynamics. Since attention acts as a \textit{linear mixer} of historical states, it cannot alone represent nonlinear vector fields; the feed-forward network (MLP) becomes structurally essential for approximating the nonlinear flow map. 
Through three case studies (the Van der Pol oscillator, the Chafee-Infante reaction–diffusion PDE, and the Navier–Stokes equations), we demonstrate that Transformer architectures (i) provide computational benefits primarily in partial observation regimes, (ii) operate as delay-embedding mechanisms that preserve essential physical state information, and (iii) discover latent spaces that maintain the effective dimensionality of the dynamics while capturing dominant modes and relevant system parameters.
However, discrete-time modeling has inherent limitations. Although discrete-time forward predictors can approximate the forward flow map arbitrarily well, the identified maps are frequently \textit{non-invertible}~\cite{cui2023certified}. Additionally, the topological structure of infinite-time attractors is often captured incorrectly (e.g., yielding invariant circles instead of limit cycles)~\cite{rico1992discrete}. These limitations must be taken into account when designing and interpreting Transformer-based dynamical models, particularly in settings where invertibility, periodic behavior, and parametric dependence play a central role.

\subsection{Van der Pol Oscillator}

\label{sec:Van_der_Pol}
\paragraph{Dynamical system} 
The Van der Pol oscillator is a classic nonlinear dynamical system described by the second-order differential equation:
\begin{equation}
    \ddot{x} - \mu(1 - x^2)\dot{x} + x = 0,
    \label{eq:vdp_second_order}
\end{equation}
where \( \mu \in \mathbb{R} \) is a scalar parameter controlling the damping strength. Solving equation~\ref{eq:vdp_second_order} requires the simultaneous integration of the state $x$ and its time derivative $\dot{x}$.

For our experiments, we fix \( \mu = 0.5 \), a regime that yields a non-stiff oscillator with a smooth, stable limit cycle emerging around the unstable origin. 

We generate training and test data by sampling multiple short trajectories from randomly initialized states across the system's phase space. Initial conditions \( \mathbf{x}_0 = [x_0, \dot{x}_0]^\top \) are drawn from a uniform distribution \( x, \dot{x} \sim \mathcal{U}(-3, 3) \), ensuring coverage of diverse dynamical behaviors, including transients and convergence to the limit cycle. A total of 1500 initial conditions are used.

Each trajectory is integrated forward in time over the interval \( [0, 6.5] \) using the \texttt{BDF} solver from \texttt{SciPy}'s \texttt{solve\_ivp}, with relative and absolute tolerances of \( 10^{-6} \) and \( 10^{-9} \), respectively. The integration output is sampled at a fixed time step \( \Delta t = 0.1 \). We treat these accurate simulations as ground truth.

To evaluate generalization near the system’s asymptotic behavior, we additionally simulate a long trajectory from initial state \( \mathbf{x}_0 = [2, 0]^\top \), chosen to lie very close to the limit cycle. This trajectory is integrated over 65 time units (ten times longer than the short trajectories) using the same solver and discretization settings.

\begin{figure}[ht!]
    \centering
    \begin{subfigure}[b]{0.3\linewidth}
\includegraphics[height=4.1cm, trim=0 0 0 0, clip]{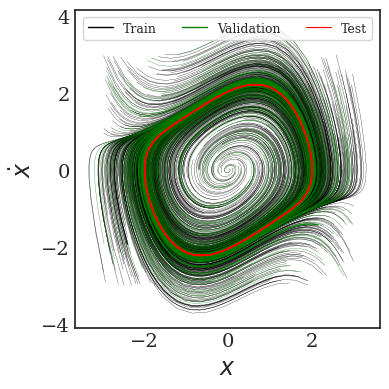}
        \caption{}
    \label{fig:vdp_phase_portrait}
    \end{subfigure}
    \hfill
    \begin{subfigure}[b]{0.3\linewidth}
        \includegraphics[height=4.2cm]{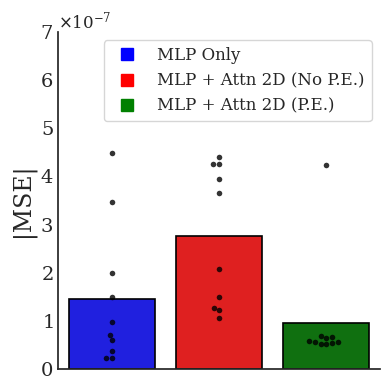}
        \caption{}
    \label{fig:boxplot_full_obs}
    \end{subfigure}
    \hfill
    \begin{subfigure}[b]{0.3\linewidth}
    \includegraphics[height=4.2cm]{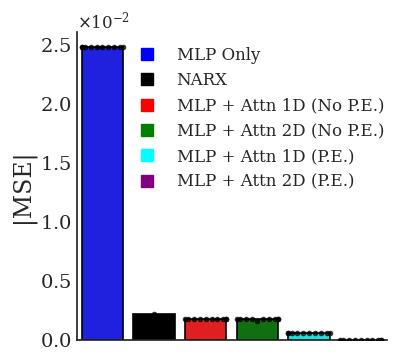}
        \caption{}
    \label{fig:boxplot_partial_obs}
    \end{subfigure}
    \hfill
    \caption{(a) Phase portrait of the Van der Pol oscillator with $( \mu = 0.5 )$. The training (black), validation (green), and test (red) trajectories are shown. (b) Prediction error (MSE) under full observation across different models. The Transformer performs comparably to the MLP, indicating that attention mechanisms do not confer a significant advantage when all state variables are observed; the full-observation NARX baseline (MSE $\approx 7.1 \times 10^{-4}$, see Table~\ref{tab:narx_vdp}) is omitted here as it lies several orders of magnitude above the plotted range, (c) 
    Prediction error (MSE) under partial observation (only \( x \) observable). The Transformer significantly outperforms the MLP, demonstrating its ability to learn latent dynamics from delayed inputs.}
\label{fig:van_der_pol_transformer_mlp}
\end{figure}

The dataset is divided into training and validation subsets in an 80:10 ratio. The long trajectory on the limit cycle is kept separate and used exclusively for testing. 
Figure~\ref{fig:vdp_phase_portrait} shows the resulting phase portrait, with different colors for the training (black), validation (green), and test (red) trajectories, clearly illustrating the long-term attractor behavior.

We conduct two experiments to illustrate the effectiveness of Transformer architectures under partial observability and to examine the resulting latent representations. These experiments investigate how the linear, attention-induced recurrence interacts with nonlinear dynamics when combined with a feed-forward mapping, and how the availability of state information versus delayed observations affects the implicit state reconstruction.

\paragraph{Transformer setup}
In our first experiment, we compare the predictive performance of a typical Transformer model, with a single attention head followed by a feedforward MLP layer, against traditional feedforward MLPs when provided full state observations. For this first experiment we also test Transformers with and without PE. The number of neurons used for the MLPs is the same across the Transformer and MLP architectures. For the Transformer model, the prediction task involves forecasting future states from a context window of $n=5$ past observations, consistent with the Takens-style requirement $n\ge 2d_A+1$ for an attractor of dimension $d_A=2$. In contrast, the MLP model is tasked with learning the \textit{time-one map}: given the current state, it predicts the next state after time $\Delta t$ without incorporating any additional historical observations. In this fully observed setting, where all state variables are available at each time step, one might expect both approaches to perform comparably--since the current state alone already encapsulates the full information needed to predict the future evolution of the system.

In our second experiment, we consider a partial observability setting where only $x$ is provided to the model. This reflects a situation where some components of the system are unmeasured. In this setting, successful prediction requires reconstruction of the unobserved state from time-delayed measurements, aligning with the delay-coordinate embedding perspective formalized by Takens’ theorem \cite{takens1981detecting}.

\paragraph{Results and observations}
For the full observation case, we train each model ten times with different random seeds, keeping all other hyperparameters constant, to assess robustness. The resulting performance across seeds is summarized in Figure~\ref{fig:boxplot_full_obs}, which shows the mean squared error (MSE) for all runs on the test limit cycle. The single MLP only, and the single-layer single-head Transformers exhibit similar performance, both having $\mathrm{MSE}\approx 10^{-7}$. This aligns with our expectations: when the full state is observable, a simple nonlinear function approximator can effectively learn the time-one map without requiring access to historical information.

For our second experiment, we consider the partial observability setting where only $x$ is provided to the model. Figure~\ref{fig:boxplot_partial_obs} shows a performance gap: the Transformer (MLP + attention) outperforms the MLP-only model, as expected. This difference arises because the MLP only has access to a single scalar observation $x$,  at a single point in time, and must predict future states without access to the full state of the system or a history of past measurements. As a result, it cannot distinguish between points in a trajectory that share the same $x$ value but differ in their phase.
In contrast, the Transformer can leverage a window of $n=5$ past observations, enabling it to use past information to distinguish between trajectories that share the same $x$ value but differ in their phase. These results further support the interpretation that attention mechanisms enable autoregressive modeling capabilities. Further, under partial observability, the use of delayed observations provides the conditions under which such aggregations can support effective state reconstruction, in line with delay-coordinate embedding theory.

As an additional point of comparison, we evaluate a representative polynomial NARX baseline on the same full- and partial-observation tasks. As shown in Figure \ref{fig:boxplot_partial_obs} the NARX model performs better than the MLP Only model, comparably to the Transformer models with no P.E. but worse than the Transformer models with P.E.  The numerical values of those results are reported in \ref{app:vdp_narx}. The NARX model comparison is not intended as an exhaustive comparison against classical system-identification methods but rather it provides a delay-regression baseline that helps highlight the benefit of temporal history.

\begin{figure}[ht!]
    \centering
    
    \begin{subfigure}[t]{0.26\textwidth}
        \centering
        \includegraphics[height=3.4cm]{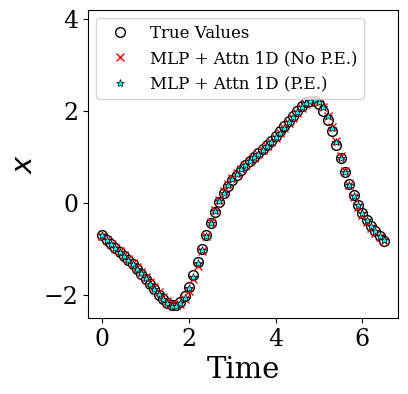}
        \caption{}
        \label{fig:Van_der_pol_1D_comparison}
    \end{subfigure}%
    \begin{subfigure}[t]{0.26\textwidth}
        \centering
        \includegraphics[height=3.4cm]{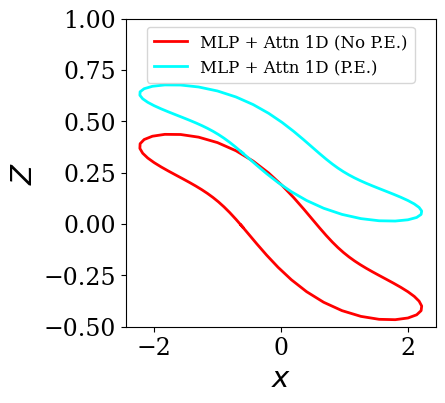}
        \caption{}
        \label{fig:Van_Der_Pol_Z_Comparison}
    \end{subfigure}%
    \begin{subfigure}[t]{0.26\textwidth}
        \centering
        \includegraphics[height=3.4cm]{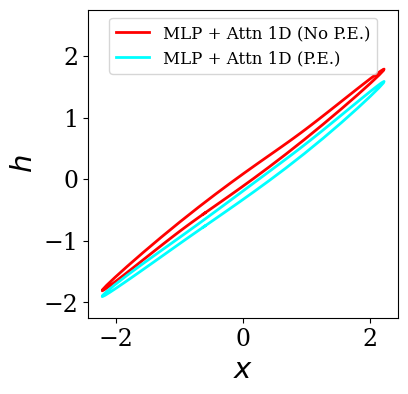}
        \caption{}
        \label{fig:Van_Der_Pol_Z_y_Comparison}
    \end{subfigure}%
    \begin{subfigure}[t]{0.26\textwidth}
        \centering
        \includegraphics[height=3.4cm]{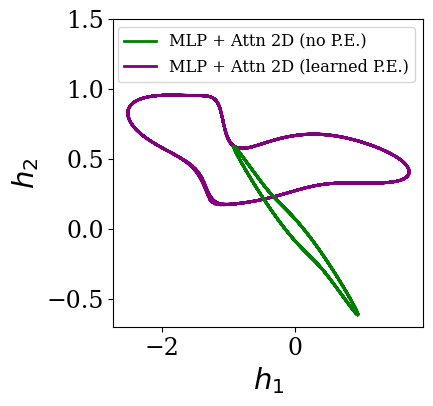}
        \caption{}
        \label{fig:Van_Der_Pol_2D_Latent_Space}
    \end{subfigure}
        
    \begin{subfigure}[t]{0.34\textwidth}
        \centering
        \includegraphics[height=3.4cm]{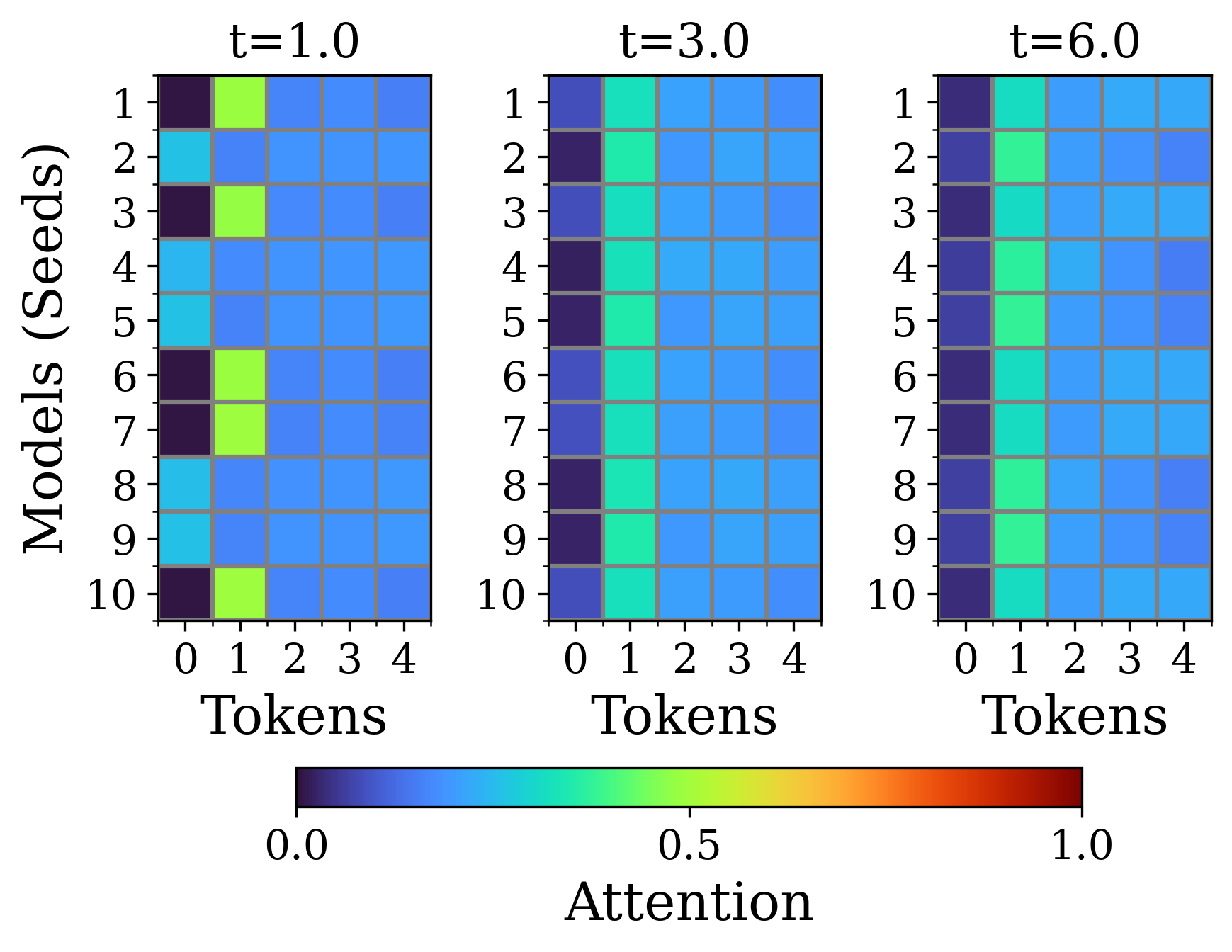}
        \caption{}
        \label{fig:attn_1D_learned}
    \end{subfigure}%
    \begin{subfigure}[t]{0.34\textwidth}
        \centering
        \includegraphics[height=3.4cm]{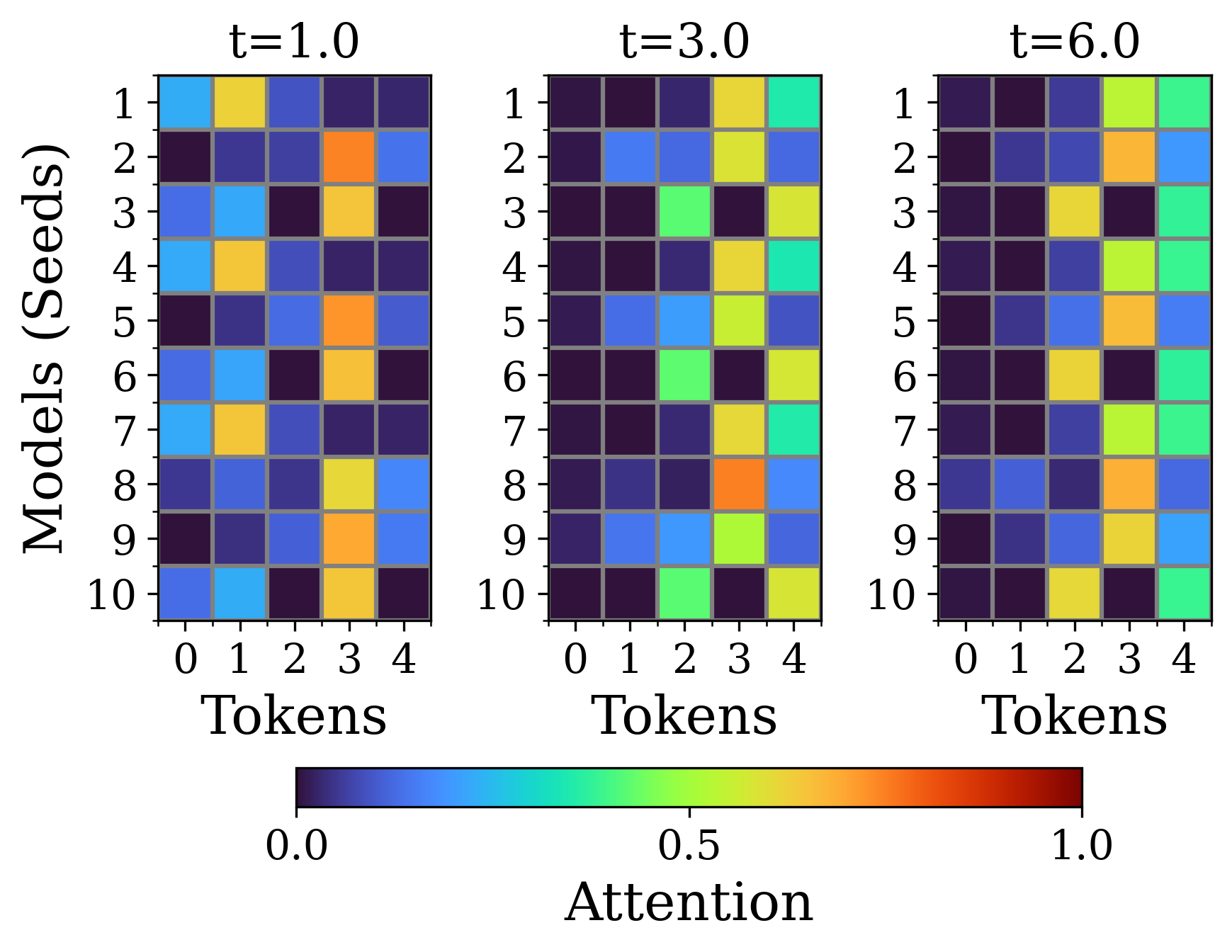}
        \caption{}
        \label{fig:attn_2D_learned}
    \end{subfigure}%
    \begin{subfigure}[t]{0.34\textwidth}
        \centering
        \includegraphics[height=3.4cm]{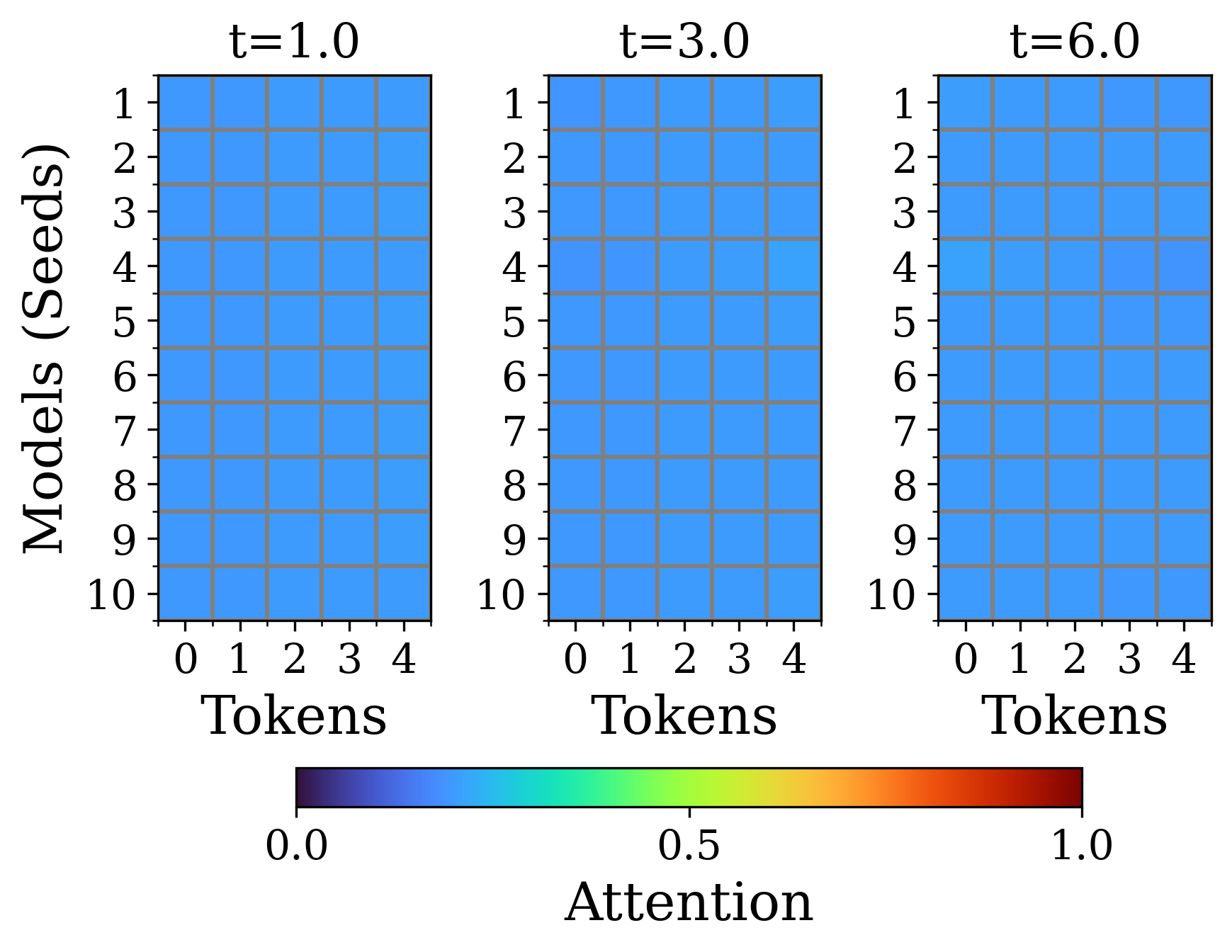}
        \caption{}
        \label{fig:attn_2D_none}
    \end{subfigure}
    
    \caption{
        (a) Predicted time series on the limit cycle. Ground truth values (black circles) are compared against predictions from two Transformer models with identical architectures: one without positional encoding (No P.E., red crosses) and one with learned positional encoding (P.E., cyan stars). 
        (b) Latent correction term \( \mathbf{z} \) plotted against \( x(t) \), showing phase-dependent separation.
        (c) Visualization of the effective latent coordinate \(\mathbf{h}_t=\mathbf{e}_t+\mathbf{z}_t\) against \(x_t\).
        (d) Latent space trajectories of 2D Transformer models with and without positional encoding.
Attention pattern visualizations across 10 trained models: 
(e) 1D MLP + Attention with learned P.E., 
(f) 2D MLP + Attention with learned P.E., and 
(g) 2D MLP + Attention without P.E.
    }
    \label{fig:combined_latent_viz}
\end{figure}

To further investigate the mechanisms underlying these observations, we analyze the Transformer's latent space for two of the trained models. As shown in Figure~\ref{fig:Van_der_pol_1D_comparison}, the predicted trajectories closely match the ground truth, confirming that the models can accurately capture the system's dynamics. 
As shown in Figure~\ref{fig:transformer_arch}, the attention output $\mathbf{z}_t$  can be interpreted as a correction term that incorporates the temporal history of the system before being added to the final embedded token $\mathbf{e}_t$. The output layers receive the effective latent coordinate $\mathbf{h}_t=\mathbf{e}_t+\mathbf{z}_t$. This suggests that for the same value of $x_t$, which can appear at different phases of the trajectory, the readout-relevant coordinate $\mathbf{h}_t$ carries the information required to distinguish between phases. This is shown first in Figure~\ref{fig:Van_Der_Pol_Z_Comparison}, where we plot $x_t$ against $\mathbf{z}_t$ to isolate the attention contribution. For the same value of $x_t$, we see that the Transformer learns two different corrections (values of $\mathbf{z}_t$), which matches our expectations.

Although this analysis is not strictly necessary in the 1D ($d_{\mathrm{lat}}=1$) setting, we include it here to build intuition for cases where it becomes more informative. An equivalent visualization plots the effective latent coordinate $\mathbf{h}_t=\mathbf{e}_t+\mathbf{z}_t$, as shown in Figure~\ref{fig:Van_Der_Pol_Z_y_Comparison}. This figure again reveals that the same value of $x_t$ corresponds to two distinct values of $\mathbf{h}_t$. In the next examples, we use $\mathbf{h}_t$ consistently to visualize the Transformer's latent space.

We next consider a 2D latent space ($d_{\mathrm{lat}}=2$) for the Van der Pol system. Recall that the Transformer input consists of five one-dimensional delays of $x(t)$. The scalar observation is embedded by a learned linear map
\[
\mathbf{e}_t=x_t W_{\mathrm{emb}}\in\mathbb{R}^{2},
\qquad
W_{\mathrm{emb}}\in\mathbb{R}^{1\times 2}.
\]
This embedding precedes the attention mechanism and is applied independently to each delayed input.

In this setting, we visualize the learned latent trajectories by plotting the two components of $\mathbf{h}_t=\mathbf{e}_t+\mathbf{z}_t$, where $\mathbf{z}_t=(z_{t,1},z_{t,2})$ is the final-token attention output, as shown in Figure~\ref{fig:Van_Der_Pol_2D_Latent_Space}. For both Transformer models---with and without PE---the resulting latent trajectories recover a limit-cycle structure. This indicates that the Transformer learns a representation aligned with the intrinsic dynamics of the system. To assess robustness, in \ref{sec:van_der_pol_appendix} we present the same visualization as in Figures~\ref{fig:Van_Der_Pol_Z_Comparison}, \ref{fig:Van_Der_Pol_Z_y_Comparison}, and \ref{fig:Van_Der_Pol_2D_Latent_Space} for 10 independently trained models (different random seeds).

We then proceed to investigate the entries in the attention matrix, focusing on the last row where the model uses five history tokens to make the next prediction in time. We report these results across all 10 seeds for: (a) 1D MLP + Attention with learned P.E., (b) 2D MLP + Attention with learned P.E., and (c) 2D MLP + Attention without P.E., as shown in Figures \ref{fig:attn_1D_learned}, \ref{fig:attn_2D_learned}, and \ref{fig:attn_2D_none}, respectively.
A consistent observation across all models is that attention is distributed over multiple past tokens rather than concentrating on a single time step. 
This behavior is expected from a dynamical systems perspective: under partial observability, a scalar measurement is not enough to define the system and multiple delayed observations are required to reconstruct the system state, in accordance with delay-coordinate embedding theory. The distinction between models trained with and without PE lies in the structure of the learned attention pattern. Without PE, attention weights collapse to an approximately uniform averaging over the delay window, an order-invariant aggregation that acts as a simple history-averaging estimator, essentially a sample mean of the delayed embeddings up to the learned value and output projections. With learned PE, the attention mechanism can assign lag-specific weights, and the resulting pattern is closer to a learned finite-memory filter acting on delay coordinates. This distinction also explains why PE can reshape the attention structure without systematically improving prediction accuracy: the unencoded delay window may already carry sufficient information for one-step prediction, and PE changes how that information is organized rather than what is available. Whether the structured regime corresponds exactly to classical constructs, such as an optimal autoregressive filter or a Padé-type rational approximation of the transfer function, remains open and is discussed as a future direction in Section~\ref{sec:future_directions}.

For both cases (1D and 2D inner dimensions), we provide additional visualizations of the query (Q), key (K), and value (V) for representative models in \ref{sec:van_der_pol_appendix}. In our view, these plots do not provide significant additional insight into the Transformer's latent space, but we include them for completeness.

\subsection{Chafee-Infante system}
\label{sec:Chafee_Infante}

To test whether basic Transformer architectures can uncover meaningful latent spaces for PDEs as well as ODEs, we consider the Chafee-Infante reaction diffusion equation. The equation has the form

\begin{equation}
    u_t = u - u^3 + \nu u_{xx},
    \label{eq:chafee_infate}
\end{equation}
and for our experiments we consider boundary conditions $u(0,t) = u(\pi,t) = 0$ and $\nu =0.16$. For $\nu =  0.16$ it has been shown that the long-term dynamics live in a two-dimensional \textit{inertial manifold} \cite{foias1988computation,gear2011slow,evangelou2023double,koronaki2024nonlinear}.

\begin{figure}[htbp]
    \centering
    \begin{subfigure}[b]{0.34\textwidth}
        \centering
        \includegraphics[width=\textwidth]{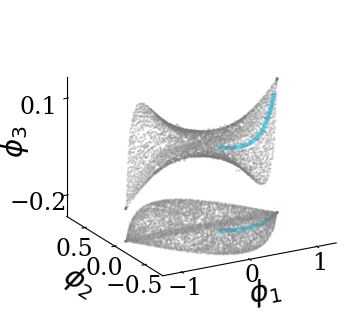}
        \caption{}
        \label{fig:fourier_modes_traj}
    \end{subfigure}
    \begin{subfigure}[b]{0.34\textwidth}
        \centering
        \includegraphics[width=\textwidth]{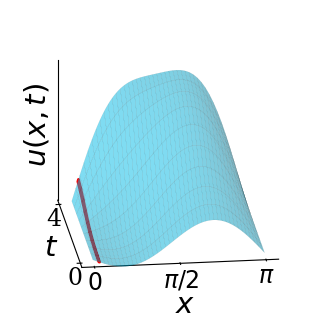}
        \caption{}
        \label{fig:uxt_traj}
    \end{subfigure}
    \begin{subfigure}[b]{0.3\textwidth}
        \centering
        \includegraphics[width=\textwidth]{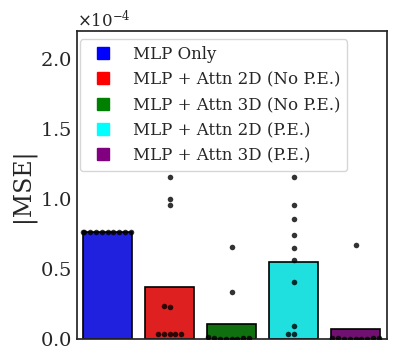}
        \caption{}
        \label{fig:CI_Errors_Box_Plot}
    \end{subfigure}
    \caption{(a) Sampled data expressed in terms of the first three Fourier modes $(\phi_1,\phi_2,\phi_3)$ for the Chafee-Infante PDE, revealing a two-dimensional manifold embedded in $\mathbb{R}^3$. 
    The projection of the manifold onto the $(\phi_1,\phi_2)$ plane is shown, and a reference trajectory is highlighted in light blue both in the ambient space and in the projection. 
    (b) The (light blue) reference trajectory reconstructed in the physical space $u(x,t)$ is shown. 
    The single pointwise observation at the 10th grid point $x_{10}$, denoted $u_{10}(t)$, is plotted in red. 
    (c) Distribution of reconstruction errors for trajectories with missing observations.}
    \label{fig:fourier_and_uxt_part1}
\end{figure}

We follow the same Galerkin projection approach and sampling scheme as in \cite{gear2011slow,evangelou2023double} to ensure that the data lie near/on this low-dimensional manifold. Specifically, we approximate the solution as
\begin{equation}
    u(x,t) \approx \sum_{k=1}^{3} \phi_k(t)\,\sin(kx),
    \label{eq:galerkin}
\end{equation}
which yields a system of three spectral ODEs:
\begin{equation}
    \frac{d\vect{\phi}}{dt} = \vect{f}(\vect{\phi}),
    \label{eq:spectral_odes}
\end{equation}
where \( \vect{\phi} = (\phi_1, \phi_2, \phi_3) \).
 The sampled data were obtained by integrating Eq.~\eqref{eq:spectral_odes} from a range of initial conditions and discarding transients, yielding samples that lie on the two-dimensional inertial manifold embedded in the three-dimensional Fourier space. The inertial manifold is fully parametrized by the first two modes $\phi_1$ and $\phi_2$, shown in Figure \ref{fig:fourier_modes_traj} as projection. The manifold is symmetric with respect to the origin as can be seen from the projection in Figure \ref{fig:fourier_modes_traj}; as we discuss below this has implications for the embedding structure that the Transformer can capture or reveal.

For our purposes, we use these samples as initial conditions and generate short trajectories by integrating the spectral system over the interval \([0, 4]\). Each trajectory is sampled at 10 uniformly spaced time points.
We then map each trajectory from the Fourier space back to the physical solution $u(x,t)$ using Eq.~\eqref{eq:galerkin}. For the reconstruction a uniform spatial grid consisting of 256 points was used.
As observations for the Transformer model, we extract the signal at the 10th grid point $x_{10} \approx 0.11$ of the uniform 256-point spatial discretization over $[0, \pi]$ denoted $u(x_{10}, t) = u_{10}(t)$. We assume this is the only available measurement.

For all experiments reported below, we split the sampled short trajectories into training, validation, and test sets using a 70\% / 15\% / 15\% ratio, respectively.
Each sampled trajectory (consisting of 10 snapshots) is decomposed into overlapping input windows of $n=5$ delayed tokens for training.
The choice $n=5$ is consistent with the Takens-style requirement $n\ge 2d_A+1$ for reconstructing a manifold of dimension $d_A=2$. 

 \paragraph{Transformer setup}

 \begin{figure}[htbp]
    \centering
    \begin{subfigure}[b]{0.95\textwidth}
        \centering        \includegraphics[width=\textwidth]{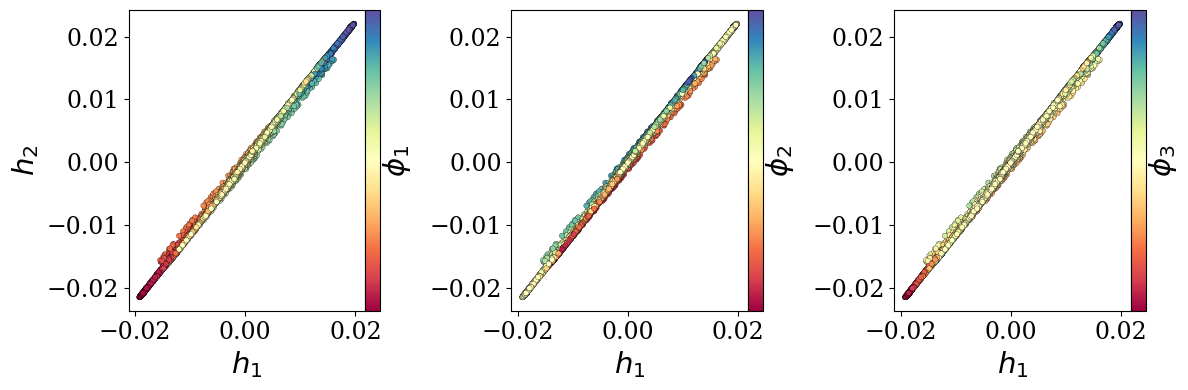}
        \caption{}
    \label{fig:CI_Projection_2D}
    \end{subfigure}

    \vspace{0.6em}

    \begin{subfigure}[b]{0.48\textwidth}
        \centering
        \includegraphics[width=\textwidth]{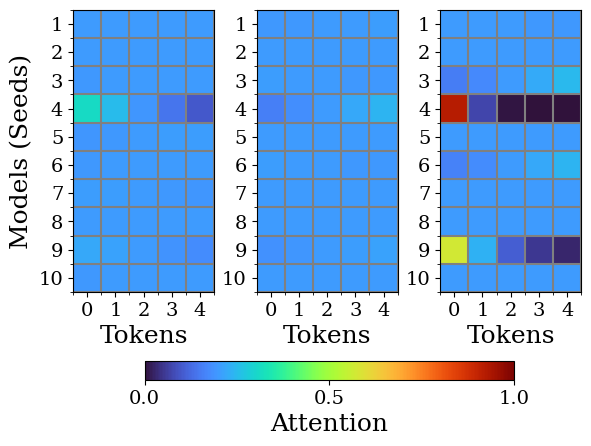}
        \caption{}
        \label{fig:ci2d_attn_pe}
    \end{subfigure}
    \hfill
    \begin{subfigure}[b]{0.48\textwidth}
        \centering
        \includegraphics[width=\textwidth]{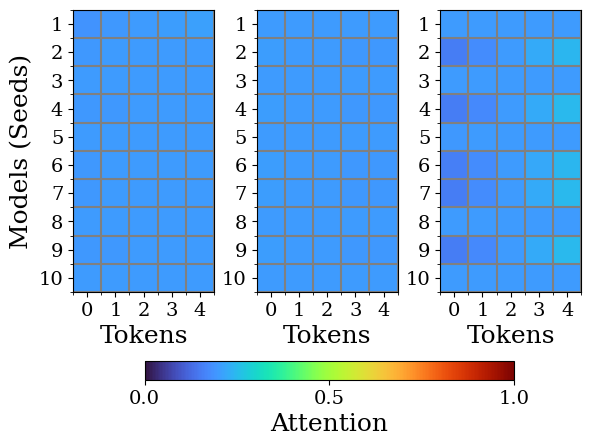}
        \caption{}
        \label{fig:ci2d_attn_nope}
    \end{subfigure}
    \caption{Representative model MLP + Attn. 2D. (a)  Effective latent coordinate $\mathbf{h}_t=\mathbf{e}_t+\mathbf{z}_t$ plotted in the $(h_{t,1},h_{t,2})$ plane colored by $\phi_1$, $\phi_2$, and $\phi_3$.
    Attention matrices for three representative data points across ten models trained with different random seeds: (b) models with PE and (c) models without PE.}
\label{fig:2D_CI_Latent_Space_Transformer}
\end{figure}

 \begin{figure}[htbp]
    \centering
    \begin{subfigure}[b]{0.95\textwidth}
        \centering
        \includegraphics[width=\textwidth]{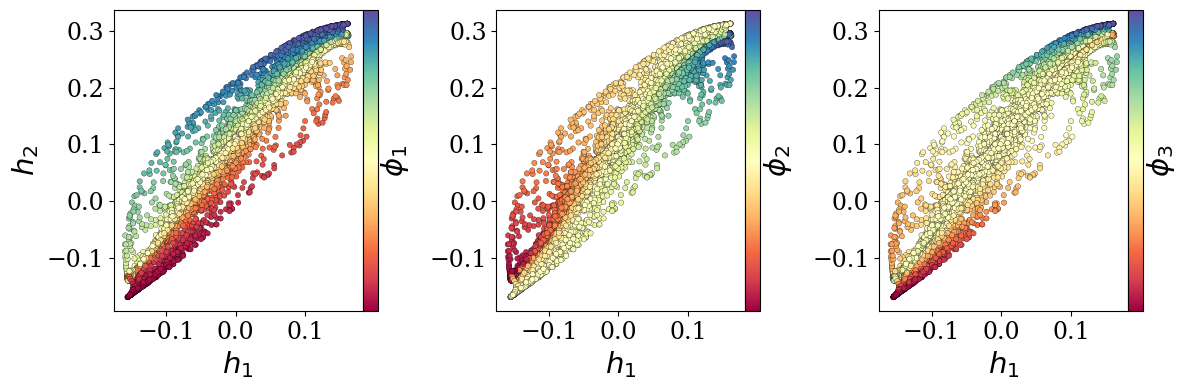}
        \caption{}
        \label{fig:CI_Projection_3D}
    \end{subfigure}
        \begin{subfigure}[b]{0.95\linewidth}
        \centering
        \includegraphics[width=\linewidth]{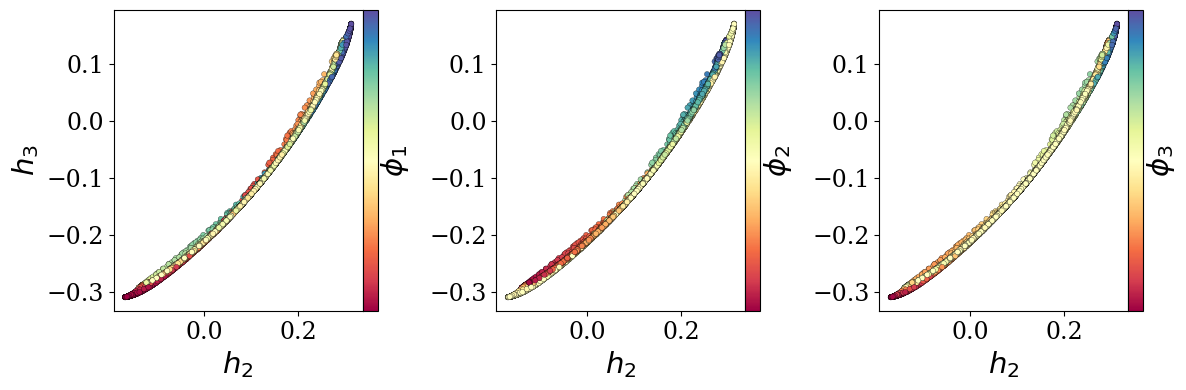} 
        \caption{}
        \label{fig:CI_Projection_3D_B}
    \end{subfigure}
    \begin{subfigure}[b]{0.48\textwidth}
        \centering\includegraphics[width=\textwidth]{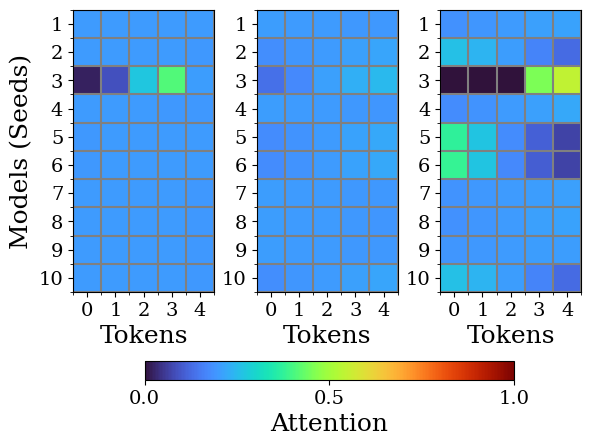}
        \caption{}
        \label{fig:attn_pe_3D}
    \end{subfigure}
    \hfill
    \begin{subfigure}[b]{0.48\textwidth}
        \centering
        \includegraphics[width=\textwidth]{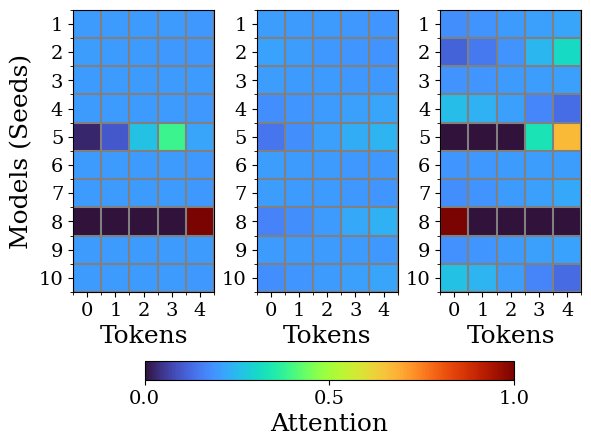}
        \caption{}
        \label{fig:attn_nope}
    \end{subfigure}
    \caption{Representative model MLP + Attn. 3D. The effective latent coordinate $\mathbf{h}_t=\mathbf{e}_t+\mathbf{z}_t$ plotted in the (a) $(h_{t,1},h_{t,2})$ plane and (b) $(h_{t,2},h_{t,3})$ plane. In both subfigures the three panels show the same points
    colored by $\phi_1$, $\phi_2$, and $\phi_3$, respectively. Attention matrices for three representative data points across ten models trained with different random seeds: (c) models with PE and (d) models without PE.}\label{fig:3D_CI_Latent_Space_Transformer}
\end{figure}

 To examine the benefit of temporal attention in settings with limited spatial information, we begin by comparing a simple MLP-only baseline against Transformer-based models. Both are trained to predict the next value of the signal $u_{10}(t)=u(x_{10},t)$. The MLP uses only $u_{10}(t)$ and attempts to predict $u_{10}(t+\Delta t)$. The Transformer receives the delayed input window
\[
    Y_t=[u_{10}(t-4\Delta t),\ldots,u_{10}(t)]^{\top}
\]
as input tokens, processes this window via a single attention layer, and predicts $u_{10}(t+\Delta t)$.

We consider Transformer variants with $d_{\mathrm{lat}}=2$ and $d_{\mathrm{lat}}=3$, trained with and without PE for 10 different random seeds.  All models share the same MLP architecture and training protocol (e.g., number of epochs, learning rate). This setup mirrors our earlier experiments on the Van der Pol system and allows for a direct assessment of the Transformer's ability to recover latent structure when only partial observations are available. The choices $d_{\mathrm{lat}}=2$ and $d_{\mathrm{lat}}=3$ were made using prior knowledge of the true dimensionality of this system. Recall that the data were sampled from a two-dimensional (symmetric) nonlinear manifold embedded in a three-dimensional space. 

From these experiments (Figure \ref{fig:CI_Errors_Box_Plot}), we observe that the MLP-only baseline exhibits the highest MSE among all architectures. In contrast, the Transformer-based models generally achieve lower errors, though the variants with $d_{\mathrm{lat}}=2$ display substantial variance across training runs, with some instances performing worse than the MLP. This suggests that using $d_{\mathrm{lat}}=2$ can lead to unstable training dynamics and inconsistent predictive performance. The models with $d_{\mathrm{lat}}=3$ demonstrate the best overall performance and lowest median errors. Notably, the inclusion of positional encoding (PE) does not appear to provide any systematic improvement in either the 2D or 3D configurations.

\paragraph{Results and observations}
Despite the fact that the underlying inertial manifold of the Chafee-Infante system is two-dimensional, the Transformer with $d_{\mathrm{lat}}=2$ does not recover a clean, unfolded representation. As illustrated in Figure~\ref{fig:CI_Projection_2D}, the effective latent coordinate $\mathbf{h}_t=\mathbf{e}_t+\mathbf{z}_t$ (cf.\ Eq.~\eqref{eq:effective_latent}) does not form a smooth two-dimensional surface but instead collapses into a narrow, thickened curve. This indicates that the model is not able to discover the full two-dimensional parametrization of the attractor.

To understand why this occurs, we recall that from a delay-embedding perspective Takens' theorem suggests that $n=5$ delayed observations are sufficient to reconstruct an attractor of dimension $d_A=2$ from a generic scalar observable under generic conditions. In our setting, we do provide five delayed measurements of $u_{10}(t)$, so the temporal information supplied to the model is, in principle, sufficient. The difficulty therefore does not stem from a lack of sufficiently long temporal history.

Takens’ theorem guarantees that such delay coordinates contain sufficient information to reconstruct the underlying attractor, but it does not guarantee that this information can be faithfully represented after parametric compression into a prescribed latent dimension.
In the Transformer, this compression is enforced by the choice of the internal latent dimensionality, which constrains how the attention-induced state can represent the unfolded delay embedding.
Thus, we argue that the limitation arises from the dimensionality of the Transformer's latent space. By restricting the latent dimension to $d_{\mathrm{lat}}=2$, we require the model to (i) integrate information across the five delayed tokens via attention, (ii) infer the nonlinear relationships among these delays, (iii) construct an appropriate attention contribution $\mathbf{z}_t$, and simultaneously (iv) represent the resulting unfolded geometry in only two latent coordinates. This places a strong constraint on the model, as it must find the right delay embedding and compress it directly into a two-dimensional representation. In practice, this appears to overconstrain the learning problem and prevents the Transformer from fully unfolding the underlying manifold. To make this a bit more clear to the reader, an embedding ``unfolds'' the manifold when it avoids self-intersections: different states of the underlying dynamics must map to distinct points in latent space, rather than being folded on top of one another.
The key issue is that setting $d_{\mathrm{lat}}=2$ forces premature compression of the delay embedding. In contrast, $d_{\mathrm{lat}}=3$ provides sufficient room for the attention mechanism to first unfold the delay coordinates into a richer intermediate representation, after which the nonlinear MLP can project onto the intrinsic two-dimensional inertial manifold. This separation of concerns (unfolding followed by projection) appears essential for successful state reconstruction.

This difficulty might be exacerbated by the symmetry of the Chafee-Infante inertial manifold. Because the attractor is symmetric with respect to the origin, a single spatial observable such as $u_{10}(t)$ induces a delay embedding that is not injective, yielding a "folded" representation. We provide a deeper discussion and additional visualizations of this folding effect in \ref{sec:chafee_infante_folding}. The geometry presented to the model is therefore more intricate than a simple two-dimensional sheet, making the unfolding problem even more challenging under strict 2D latent constraints.

For the remainder of this section we focus on the latent space for the models trained with $d_{\mathrm{lat}}=3$. In Figure~\ref{fig:CI_Projection_3D} and Figure~\ref{fig:CI_Projection_3D_B}, we show projections of the Transformer's latent space colored by the Fourier modes $\phi_1,\phi_2,\phi_3$. The latent space is effectively two-dimensional since $h_{t,3}$ is a function of $h_{t,2}$, while $h_{t,1}$ and $h_{t,2}$ vary independently (where $\mathbf{h}_t=(h_{t,1},h_{t,2},h_{t,3})$). In addition, Figure \ref{fig:CI_Projection_3D} shows that the Fourier modes are functions on the Transformer's latent space. This suggests that the representation the Transformer learns, despite observing only the single spatial coordinate $u_{10}$, suffices to recover the true underlying dynamics.

The observed attention patterns in Figure~\ref{fig:attn_pe_3D} indicate that the model distributes weight across multiple delayed tokens rather than collapsing onto a single dominant lag. This is consistent with the need to integrate information across several delays in order to reconstruct the effective state from a single spatial observation. The absence of a sharply peaked attention profile suggests that no single delay is sufficient, and that the attention mechanism functions as a distributed linear aggregator over the reconstructed delay coordinates. Interestingly, in this case the learned distribution of attention weights between the models with PE and without PE appear similar.

Furthermore, to assess the robustness of the learned representations we include in \ref{sec:chafee_one_variable_extended} additional visualizations across the 10  trained models for different random seeds. We also provide representative visualizations of the query, key, value and attention-output quantities ($Q_t$, $K_t$, $V_t$, and $Z_t$).

\subsection{Navier-Stokes: Flow past a cylinder}
\label{sec:cylinder_flow}
\paragraph{Dynamical system}
The final system we consider is the 2D flow past a cylinder, governed by the incompressible Navier--Stokes equations:
\begin{equation}
    \frac{\partial \mathbf{v}}{\partial t} + (\mathbf{v} \cdot \nabla)\mathbf{v} = -\frac{1}{\rho} \nabla p + \nu \nabla^2 \mathbf{v}, \qquad \nabla \cdot \mathbf{v} = 0
\end{equation}
where $\mathbf{v}=(v_x,v_y)$ is the velocity field, $p$ the pressure, $\rho$ the density, and $\nu$ the kinematic viscosity. The flow is thus characterized by three scalar fields: the two velocity components and the pressure.

For our experiments, we use the dataset from \cite{geneva2022transformers} containing flow trajectories for various Reynolds numbers $Re \in [100, 750]$, where each Reynolds number corresponds to a single trajectory. After discarding initial transients, the flow exhibits periodic vortex shedding and the dynamics reside on a stable limit cycle. Consequently, the data reside on a low-dimensional manifold, justifying the use of a compact latent representation~\cite{deane1991low, menier2025interpretable}. To assess generalization across and beyond the training distribution, we partition the available trajectories into a training set spanning $Re \in [125, 725]$ and a test set composed of six in-distribution Reynolds numbers $\{133, 233, 333, 433, 533, 633\}$ together with three out-of-distribution values $\{100, 733, 750\}$ that lie just below and above the training interval.

\begin{figure}[ht!]
    \centering
    \includegraphics[width=0.92\textwidth]{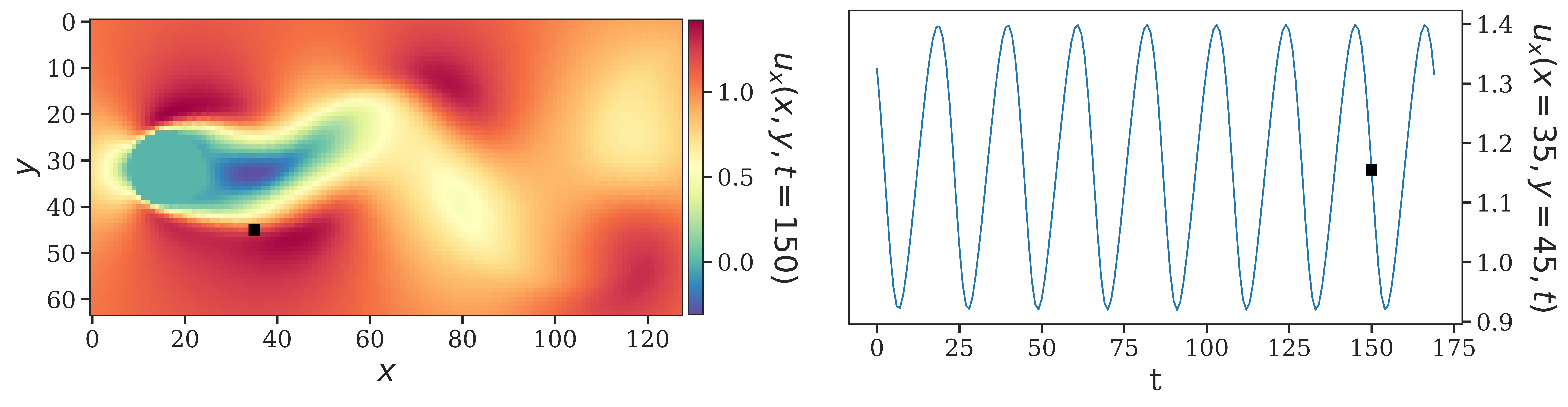}
    \caption{Selected spatial location ($x=35$, $y=45$) on the $v_x$ velocity field used for training, and corresponding time-series from that location.}
    \label{fig:selected_point}
\end{figure}

We select a single spatial location from the $v_x$ component of the velocity field at coordinates $x = 35$, $y = 45$, as shown in Figure~\ref{fig:selected_point}. This scalar observation is made across all parameter values and times, providing a one-dimensional time series for each Reynolds number. The specific choice of coordinates is not critical; any spatial location that exhibits representative dynamics (i.e., not on a boundary or in a stagnant region) would suffice.

\paragraph{Transformer setup}
As in all previous experiments, the modeling and architectural choices are guided by Takens' embedding theorem, which suggests that $n\ge 2d_A+1$ time-delayed observations are sufficient to capture the underlying dynamics on a manifold of dimension $d_A$. We assume that the intrinsic dimension of the system is at most 3: two dimensions to describe the limit cycle, and one for the Reynolds number variation.

Based on this analysis, we use a model with $d_{\mathrm{lat}}=3$, and in our experiments, we find that 5–7 input tokens provide sufficient temporal context (with the upper end of this range consistent with Takens-style delay embedding heuristics), with diminishing returns beyond this range. The choice of the internal latent dimension determines the capacity of this surrogate state to represent both the oscillatory dynamics and the variation induced by the Reynolds number. Beyond prediction, this experiment also probes whether the learned latent representation is not only minimal but also sufficient for reconstructing the full spatial field--that is, whether the Transformer discovers a compact embedding from which the complete flow state can be recovered.

\paragraph{Results and observations}

We perform two experiments to assess how explicit parameter information affects the learned representation. In the first experiment, the Transformer receives only the scalar time history of $v_x$ at the selected spatial location. In the second experiment, the normalized Reynolds number is added as an additional input. The purpose of this comparison is not simply to show that parameter conditioning improves prediction, which is expected. Rather, it allows us to examine how the absence or presence of $Re$ changes the geometry of the latent space, and whether this geometry is sufficiently organized to support reconstruction of the full flow field.

\begin{figure}[htbp]
    \centering
    \begin{subfigure}[b]{0.6\textwidth}
        \centering
        \includegraphics[width=\linewidth]{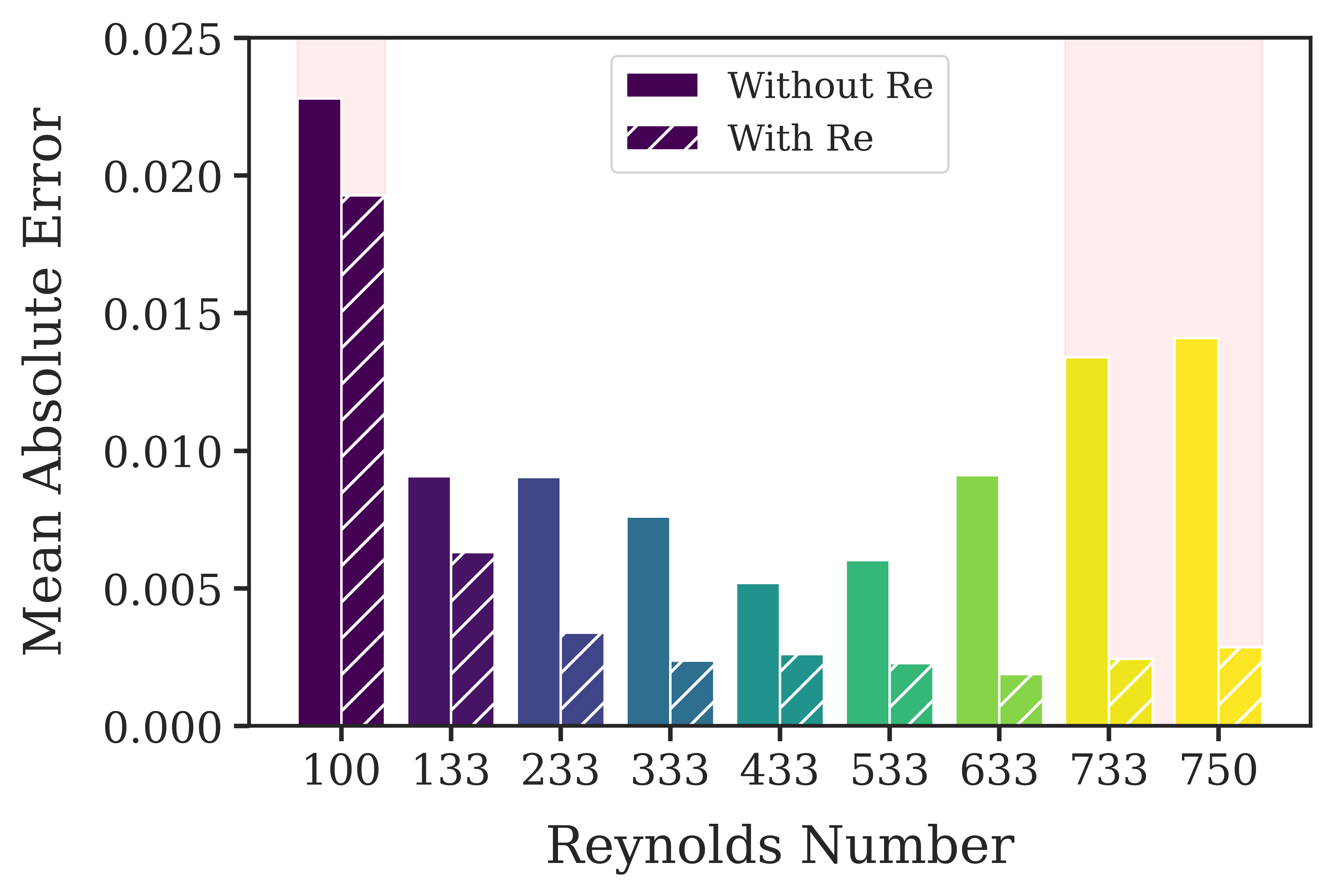}
        \caption{}
        \label{fig:combined_re_comparison}
    \end{subfigure}

    \begin{subfigure}[b]{0.49\textwidth}
        \centering
        \includegraphics[trim={0.7cm 0 0.0cm 0}, width=\linewidth]{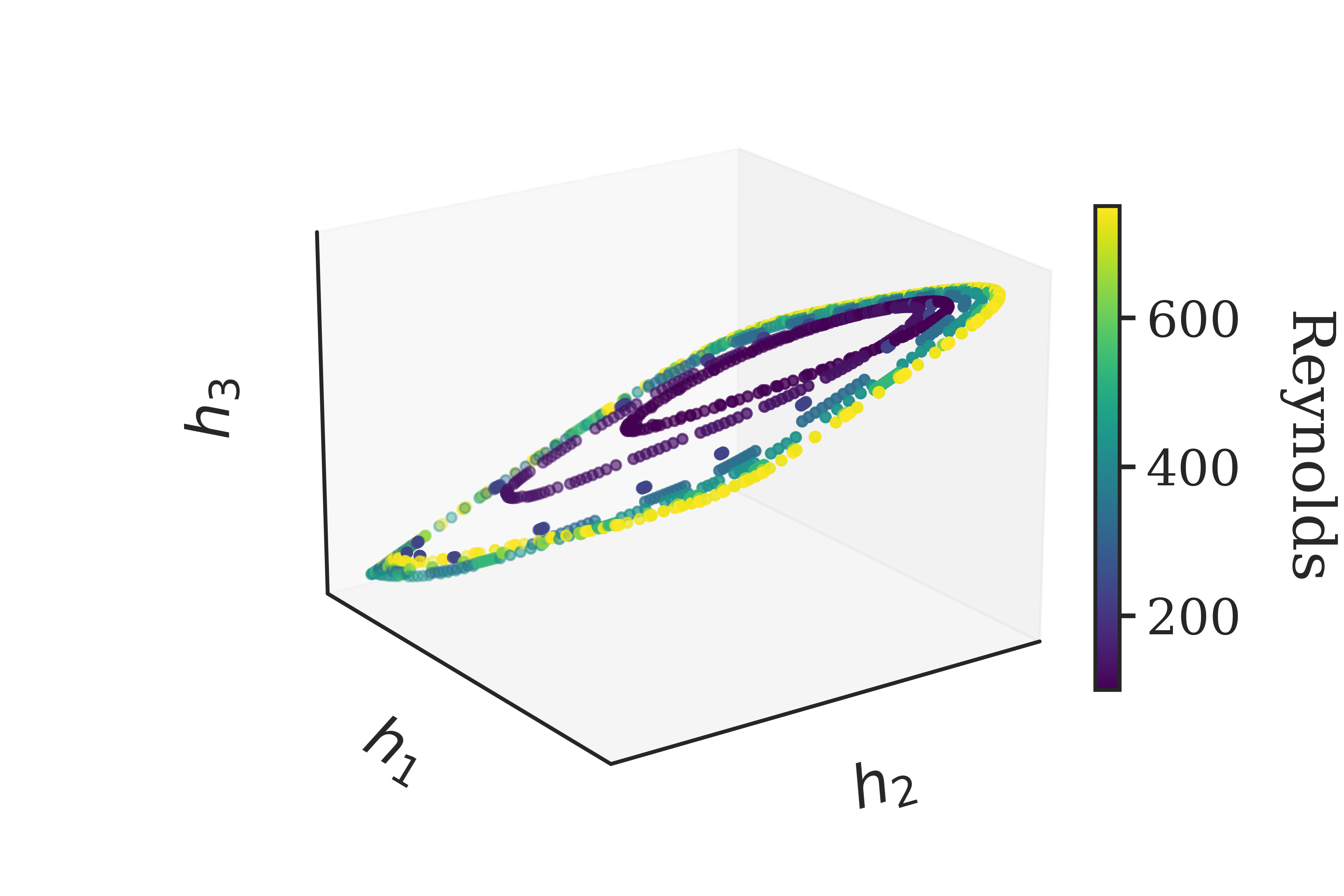}
        \caption{}
        \label{fig:cylinder_latent_no_re}
    \end{subfigure}
    \hfill
    \begin{subfigure}[b]{0.49\textwidth}
        \centering
        \includegraphics[trim={0.7cm 0 0.0cm 0},width=\linewidth]{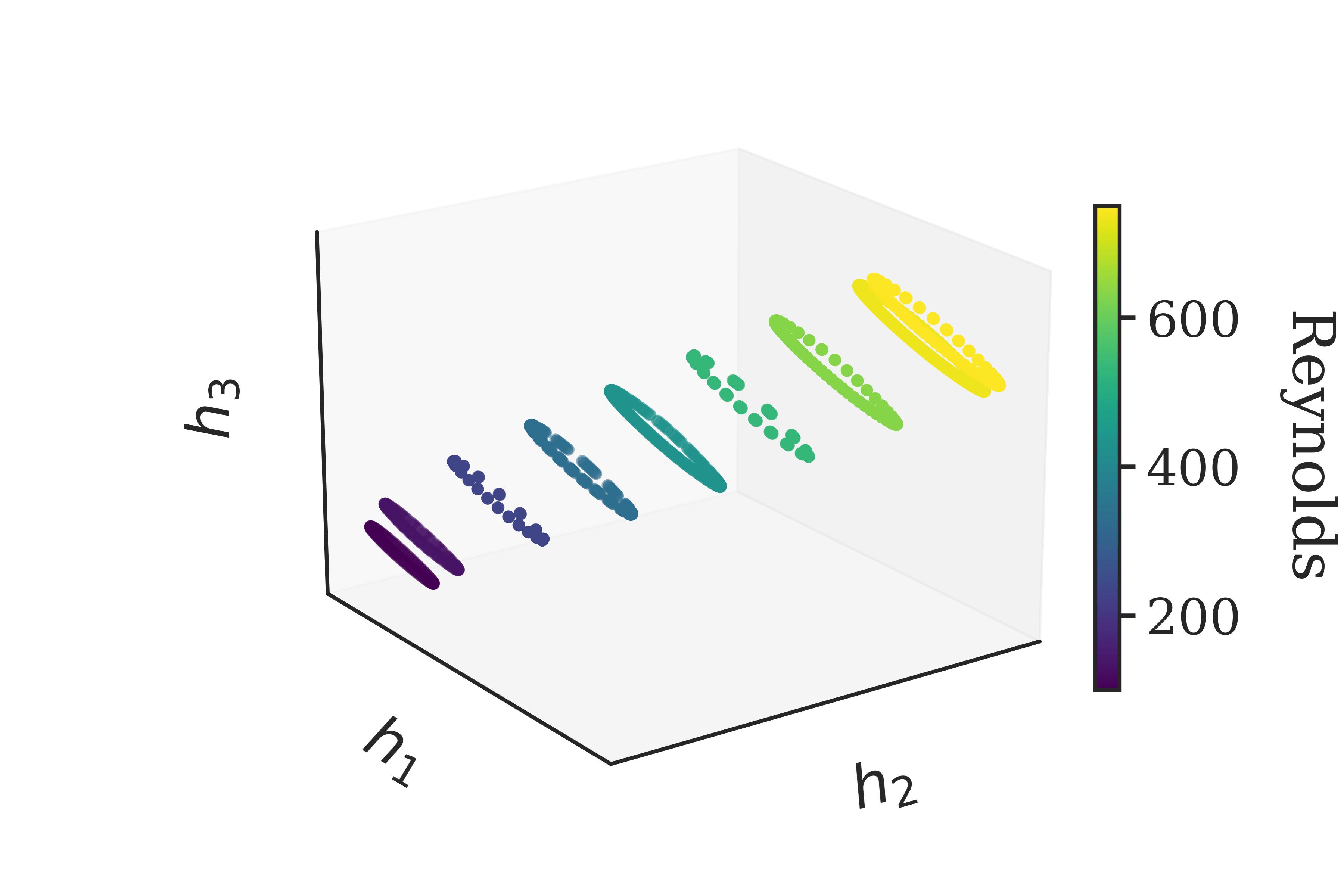} 
        \caption{}
        \label{fig:cylinder_latent_w_re}
    \end{subfigure}

    \caption{Comparison of parameter-unaware and parameter-aware Transformer training on the cylinder flow. Models are trained on $Re \in [125, 725]$ and evaluated on a held-out test set of nine Reynolds numbers, with the three out-of-distribution values $\{100, 733, 750\}$ indicated by shaded bands in (a). (a) Mean absolute error stratified by Reynolds number; the hashed bars indicate the parameter-aware model. (b,c) Three-dimensional visualization of the same latent representation, for (b) parameter-unaware and (c) parameter-aware training. The parameter-aware model achieves lower and more uniform errors across the Reynolds number range and produces a latent space with clearly separated limit cycles along a continuous manifold parameterized by $Re$. Without explicit Reynolds number input, limit cycles corresponding to different parameter values exhibit significant overlap, and the out-of-distribution cycles overlap with their in-distribution neighbors.}
    \label{fig:cylinder_comparison}
\end{figure}

Figure~\ref{fig:cylinder_comparison} summarizes the prediction results and latent representations for both models. Both models remain stable one-step predictors over the test set, with the largest mean absolute error across the two cases remaining approximately $2.3 \times 10^{-2}$. However, the distribution of error across Reynolds numbers differs substantially. Without explicit Reynolds number input, the error is lowest for intermediate Reynolds numbers and increases toward the lower and upper ends of the parameter range. This effect is especially visible for the held-out Reynolds numbers. The parameter-unaware model exhibits a clear loss of accuracy outside, or near the edge of, the training interval. In contrast, the parameter-aware model produces lower errors overall and a more uniform error profile across Reynolds numbers. For the upper held-out Reynolds numbers, the errors remain close to those observed for neighboring in-distribution cases, indicating that the model has learned a locally consistent parameter-conditioned representation.

The effective latent coordinates $\mathbf{h}_t=\mathbf{e}_t+\mathbf{z}_t$ (cf.\ Eq.~\eqref{eq:effective_latent}), visualized in Figures~\ref{fig:cylinder_latent_no_re} and~\ref{fig:cylinder_latent_w_re}, explain this difference. When the model is trained without $Re$, the latent trajectories associated with different Reynolds numbers overlap strongly. In this representation, nearby latent coordinates may correspond to flow states from different parameter regimes. Such overlap is not necessarily harmful for one-step prediction at intermediate Reynolds numbers, where neighboring regimes have similar local dynamics. It becomes problematic near the parameter extremes and for held-out cases, where a small displacement in latent space may correspond to a physically different shedding regime. Thus, the parameter-unaware model learns a representation that is predictive, but not uniquely organized with respect to Reynolds number.

When $Re$ is supplied as an input, the latent geometry changes qualitatively. The model separates the limit cycles associated with different Reynolds numbers and arranges them along a smooth parameter-dependent structure. The out-of-distribution trajectories are placed near the cycles corresponding to adjacent Reynolds numbers, rather than being superposed with unrelated parts of the latent space. This suggests that explicit parameter conditioning constrains the Transformer to remain on the appropriate parameter-conditioned manifold. In this sense, $Re$ acts as a coarse but physically meaningful coordinate that removes an ambiguity left unresolved by the scalar delay history alone.

This observation is consistent with, but also refines, the delay-embedding interpretation used throughout this work. For a fixed Reynolds number, a sufficiently informative delay history should in principle reconstruct the state on the corresponding invariant set. Here, however, the model is trained across a family of systems indexed by $Re$. The Reynolds number is therefore a hidden parameter when it is not provided to the model. A finite scalar delay window may contain information about this parameter, but the training objective only penalizes one-step prediction error and does not explicitly enforce an injective representation with respect to $Re$. Consequently, trajectories with similar short-time behavior can be mapped to overlapping regions of latent space. Providing $Re$ resolves this ambiguity directly and encourages the latent space to separate phase along each limit cycle from variation across the parameter family.

\begin{figure}[htbp]
    \centering
    \begin{subfigure}[b]{\textwidth}
        \centering
        \includegraphics[width=\textwidth]{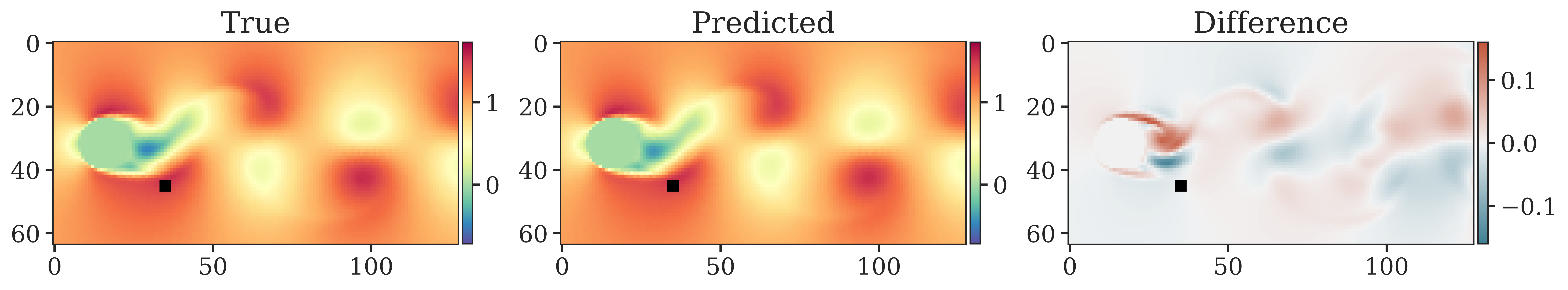}
        \caption{}
        \label{fig:cylinder_recon_no_Re}
    \end{subfigure}
    \begin{subfigure}[b]{\linewidth}
        \centering
        \includegraphics[width=\linewidth]{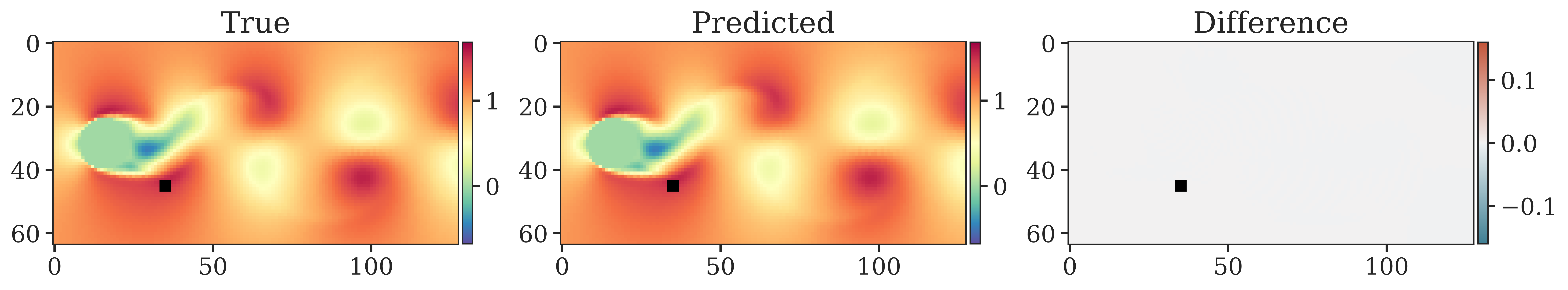} 
        \caption{}
        \label{fig:cylinder_recon_w_Re}
    \end{subfigure}
    \caption{Full-field reconstruction of the $v_x$ velocity via Geometric Harmonics from the learned latent representation. (a) Using only the effective latent coordinate $\mathbf{h}_t$ from the parameter-unaware model, reconstruction errors are concentrated near the cylinder wake where vortex shedding dynamics are most sensitive to Reynolds number. (b) With the parameter-aware latent representation, the reconstruction accurately recovers the velocity field across the domain. The black square indicates the observation location used for training.}
    \label{fig:cylinder_reconstruction}
\end{figure}

The consequences of this latent-space organization become clearer in the full-field reconstruction experiment shown in Figure~\ref{fig:cylinder_reconstruction}. We use Geometric Harmonics, a kernel-based regression method proposed by Coifman in~\cite{lafon2004diffusion}, to reconstruct the complete velocity field $v_x$ from the learned latent representation. The choice of reconstruction method is not central to our analysis; any sufficiently expressive regression scheme (e.g., Gaussian Process Regression or a neural network) could serve the same purpose. This reconstruction task tests a stronger property than one-step prediction: the latent representation must retain enough information to identify the full spatial flow state, not only the future value of the observed scalar signal. With the parameter-unaware latent space, the broad flow structure is recovered, but the difference field shows pronounced errors in the wake region. This is precisely the region where vortex shedding is most sensitive to Reynolds number. The error pattern is therefore consistent with the overlap observed in the latent space: similar latent coordinates can correspond to distinct wake configurations when the parameter is not specified.

By contrast, the parameter-aware latent representation supports an accurate reconstruction across the domain. The difference field is nearly featureless at the plotted scale, indicating that the latent coordinates retain the information needed to identify both the phase of the shedding cycle and the Reynolds-number-dependent spatial structure. The reconstruction therefore confirms that explicit parameter conditioning does more than reduce scalar prediction error. It makes the learned representation more nearly injective with respect to the physical flow fields.

To verify that this conclusion is not an artifact of a particular train-test split or of the Geometric Harmonics fit, we perform a 5-fold cross-validation study using identical reconstruction hyperparameters for both models. Across folds, the parameter-aware representation achieves a full-field test MAE of $(1.12 \pm 0.17) \times 10^{-4}$ and a wake-region MAE of $(2.48 \pm 0.40) \times 10^{-4}$. The parameter-unaware representation gives substantially larger errors, with a full-field test MAE of $(7.45 \pm 0.57) \times 10^{-3}$ and a wake-region MAE of $(1.33 \pm 0.11) \times 10^{-2}$. Thus, the parameter-unaware reconstruction error is about 70 times larger over the full field and about 50 times larger in the wake. The small fold-to-fold variability relative to this gap indicates that the difference is a property of the learned latent representation, rather than a regression artifact. More details regarding the reconstruction are reported in \ref{sec:cylinder_recon}.

These experiments demonstrate that simple, single-layer Transformers can learn meaningful latent representations of complex fluid dynamics from a single scalar observation. However, with this minimal architecture, additional input information such as system parameters may be necessary to fully disentangle complex behaviors in the latent space and enable accurate reconstruction of complete spatial fields. Deeper architectures with multiple layers or attention heads may reduce this dependence, though we leave this investigation to future work.

\section{Conclusions and Future Directions}
In this work, we have attempted to go beyond standard forecasting benchmarks to provide a mechanistic understanding of how Transformer architectures represent dynamical systems in their latent space. Rather than assessing models through aggregate performance or post hoc interpretability, we adopted a bottom-up perspective focused on the self-attention mechanism in isolation within minimal, single-layer architectures. We summarize the key findings below and outline directions for future research.

\subsection{Summary of key insights}
The key insights drawn from our numerical experiments across linear systems (Section~\ref{sec:lin_systems}) and nonlinear systems (Section~\ref{sec:nonlin_systems}) are summarized below:

\paragraph{Insights from linear systems}
For linear dynamical systems, our analysis adopted a classical autoregressive (AR) viewpoint that allowed a direct, mechanistic comparison between analytically derived AR models and the effective recursions learned by attention-only Transformer architectures.


We showed that a single-head causal attention mechanism induces a linear, time-varying autoregressive operator whose coefficients are determined by the attention weights and output projections. This enabled a direct comparison between the closed-form AR coefficients derived from the underlying physical system and the effective coefficients learned by the Transformer. In regimes where the true dynamics admit an AR representation with same-sign coefficients, the attention mechanism accurately recovers the dominant modal decay ridge, effectively learning a data-adaptive AR model consistent with classical linear system theory.

Our experiments demonstrate that attention-only Transformers can correctly capture certain oscillatory regimes where the discrete-time AR coefficients share the same sign (Case 1). In contrast, the model fails when the target dynamics require mixed-sign autoregressive coefficients. This failure arises from the softmax normalization inherent in attention, which enforces non-negativity and convexity of the attention weights. As a result, a single attention head cannot represent subtractive interactions between past states, leading to oversmoothing and an incorrect dynamic signature in Case~2: the resonance peak and coherent modal decay ridge are not recovered, even though the predicted trajectory remains visibly oscillatory. This limitation indicates that even simple linear oscillators may require additional architectural components, such as multi-head attention.

In multi-degree-of-freedom linear systems, we show that attention can recover multiple interacting modes when enough delayed information is available to form an expressive effective state. In this case, the attention mechanism aggregates past observations in a way that captures multiple spectral modes. However, this ability is still limited by the same non-negativity constraint discussed above.

\paragraph{Insights from nonlinear systems}

Attention provides little advantage when the full state vector is directly observable. In the Van der Pol oscillator under full observation, a feed-forward MLP trained to approximate the time-one map achieves predictive accuracy comparable to that of a Transformer with access to temporal context. In this setting, the system is already Markovian, and temporal aggregation does not introduce additional useful information.
In contrast, under partial observability the role of attention becomes important. When incomplete measurements are available, the Transformer leverages temporal context through attention to construct an implicit state from delayed observations. This enables the model to disambiguate states that are indistinguishable from a single snapshot but correspond to different phases or dynamical regimes. As a result, Transformers provide a clear computational benefit in experimental and data-driven settings where sensing is limited. These findings are further corroborated by recent work showing that Transformer-based models are particularly well-suited for history-dependent flows with limited data, whereas simpler architectures may suffice when dynamics depend solely on instantaneous variables~\cite{urdeitx2025can}.

Under partial observability, the behavior of attention is consistent with classical delay-coordinate reconstruction principles. Rather than introducing new dynamical information, attention learns a data-adaptive mechanism for aggregating delayed measurements into an internal representation that supports prediction. Across all nonlinear case studies, attention distributes weight across multiple past observations, reflecting the need to integrate information over time to reconstruct the effective state. From a dynamical systems perspective, this behavior aligns closely with Takens-style delay embeddings, where temporal context compensates for missing state variables. 

The quality of the learned latent representations depends critically on both temporal context and latent dimensionality. Consistent with Takens’ embedding theorem, our experiments confirm that $n\ge 2d_A+1$ delayed observations are sufficient to reconstruct an attractor of dimension $d_A$. Systems with higher intrinsic dimensionality require longer input windows to enable accurate reconstruction. When $d_A$ is unknown, a practical strategy may be to begin with a conservative delay window, estimate the intrinsic dimension of the resulting delay-coordinate dataset using methods such as Diffusion Maps or autoencoders, and then revisit the choice of $n$.
In addition to sufficient temporal history, the internal latent dimension of the Transformer must be large enough to unfold the reconstructed geometry. Insufficient latent dimensionality leads to overlapping or collapsed representations that degrade the final predictive accuracy. This effect is particularly evident in symmetric low-dimensional manifolds and parameter-varying systems. In the cylinder flow example, Transformers trained without explicit parameter input implicitly encode Reynolds number variation, but the resulting latent space does not uniquely separate different parameter values. As a result, flow states with similar short-time scalar histories but different parameter values may be mapped to overlapping regions of latent space. This non-injectivity is most visible near the parameter extremes and in the out-of-distribution test cases, where the local dynamics are less well supported by neighboring training trajectories. Providing the Reynolds number as an explicit input resolves this ambiguity by conditioning the latent representation on the relevant physical parameter. The resulting latent space separates the family of limit cycles more clearly and supports substantially more accurate full-field reconstruction.
We note that this ambiguity may also be exacerbated by the relatively close Reynolds numbers considered in our experiments, which limits the degree of parametric separation present in the data. 

A recurring theme across the nonlinear examples is that one-step prediction accuracy and latent-state identifiability are not equivalent. A representation may contain enough information to minimize local prediction error while still folding together dynamically distinct states or parameter regimes. This distinction is especially important in scientific applications, where the learned representation is often used for reconstruction, reduced-order modeling, or physical interpretation rather than prediction alone.

Together, these results indicate that attention-based models succeed in nonlinear settings when architectural choices---such as the number of input tokens $n$ and latent dimension---are aligned with the intrinsic dimensionality of the dynamics and any relevant parameter variations. Grounding these choices in delay-embedding and dynamical systems considerations can yield strong predictive performance without unnecessary model complexity, while also improving interpretability of the learned representations.

\subsection{Future directions}
\label{sec:future_directions}
These findings suggest that standard Transformer components inherited from Natural Language Processing (NLP) are not inherently optimal for physical dynamics. Future research should focus on aligning architectural inductive biases with the mathematical structure of dynamical systems:
\begin{itemize}
\item \emph{Multi-head attention and spectral expressivity}: 
The two-head experiment in \ref{sec:two_head_sdof_appendix} shows that signed recombination across heads can remove the simplest mixed-sign restriction present in the single-head scalar setting. A remaining question is how this mechanism scales in larger multi-head architectures: whether different heads specialize in distinct modal components, whether the resulting head-wise recombination preserves the correct spectral structure over longer rollouts, and whether multi-head attention can also reduce folding or collapse in nonlinear latent representations.

 \item \emph{Mechanistic Interpretability for stiff and multiscale systems:} 
Our results indicate that the primary computational benefit of training Transformers arises in partial observation regimes, where attention supports delay-based state reconstruction. Whether this conclusion extends to systems with strong stiffness or pronounced multiscale structure is an open question. In such regimes, attention may offer additional advantages by adaptively integrating information across disparate time scales, even when the full state is observable. Systematically investigating these settings would clarify when attention-based architectures genuinely outperform classical approaches. From the perspective of Backward Error Analysis, such investigations could reveal whether Transformers learn an Inverse Modified Differential Equation (IMDE) \cite{Zhu2023ImplementationA}, effectively adapting their implicit numerical scheme to the system's varying time scales and stiffness. A controlled validation would require full-state observations at a prescribed sampling interval, together with a comparison between the Transformer's learned discrete map and the IMDE computed for a reference integrator.

 \item \emph{Shared latent space geometry in foundation models}: Foundation models for dynamical systems require multiple systems or parameter regimes to share a common latent representation. Understanding how distinct dynamical behaviors are embedded, separated, or entangled within this shared space is essential for both interpretability and generalization. Our Navier–Stokes results illustrate this challenge: while Transformers can implicitly track parameter variations, explicit conditioning significantly disentangles the latent space. Developing architectures that formally separate state dynamics from parameter manifolds, and analyzing the resulting geometry, could be critical for universal forecasting models.

\item \emph{Normalization, Positional Encoding, and latent dimensionality}: Several architectural components inherited from NLP appear suboptimal for dynamical systems. Softmax normalization enforces non-negativity and convexity that restrict representable dynamics; positional encodings do not consistently benefit time-continuous physical systems; and our Chafee-Infante results show that latent dimension must exceed the intrinsic manifold dimension to allow proper unfolding before projection. These observations motivate the development of alternative normalization schemes, position-handling mechanisms, and principled guidelines for latent dimensionality--potentially drawing on numerical analysis and dynamical systems theory.

\item \emph{Role of Positional Encoding (PE) in dynamical systems:} We observed that PE yields a measurable improvement for the Van der Pol example, while in other examples, such as Chafee-Infante and the cylinder flow, PE does not systematically improve prediction accuracy or latent organization. In addition, the structure of the learned attention can vary across different initializations of the same system, suggesting that there may be multiple ways to \textit{weigh} the time-dependent behavior of these systems. We believe this merits a deeper investigation in future work, both in terms of \emph{which} structural properties of a dynamical system (e.g., stiffness of the dynamics the dynamics, spectral content) determine whether PE has an effect, and in terms of \emph{what} the resulting attention patterns represent in classical terms. In particular, it would be of interest to better understand whether a connection exists between the learned attention weights and classical estimators, such as a sample mean of delays or an optimal autoregressive filter. Establishing this connection rigorously would help explain when and why attention-based architectures gain an advantage from explicit positional information.

\end{itemize}

Ultimately, we hope this study serves as a bridge between  the empirical success of large-scale models and classical dynamical systems theory. It cautions against treating Transformers as "black-box" universal approximators, highlighting that their inductive biases, specifically regarding spectral filtering and manifold topology, must be carefully aligned with the physical systems they are intended to model.

\section*{Acknowledgments}
\noindent
GD and EC were partially supported by the French-Swiss project MISTERY funded by the French National Research Agency (ANR PRCI Grant No. 266157) and the Swiss National Science Foundation (Grant No. 200021L\_212718).
NE and IGK were partially supported by the US Department of Energy and the US National Science Foundation.

\bibliographystyle{plainnat} 
\bibliography{ref_fixed}

@article{vaswani2017attention,
  title={Attention is all you need},
  author={Vaswani, Ashish and Shazeer, Noam and Parmar, Niki and Uszkoreit, Jakob and Jones, Llion and Gomez, Aidan N and Kaiser, {\L}ukasz and Polosukhin, Illia},
  journal={Advances in neural information processing systems},
  volume={30},
  year={2017}
}

@article{geneva2022transformers,
  title={{Transformers} for modeling physical systems},
  author={Geneva, Nicholas and Zabaras, Nicholas},
  journal={Neural Networks},
  volume={146},
  pages={272--289},
  year={2022},
  publisher={Elsevier}
}

@article{zhang2024zero,
  title={Zero-shot forecasting of chaotic systems},
  author={Zhang, Yuanzhao and Gilpin, William},
  journal={arXiv preprint arXiv:2409.15771},
  year={2024}
}

@article{lai2025panda,
  title={{Panda}: A pretrained forecast model for universal representation of chaotic dynamics},
  author={Lai, Jeffrey and Bao, Anthony and Gilpin, William},
  journal={arXiv preprint arXiv:2505.13755},
  year={2025}
}

@article{kantamneni2024transformers,
  title={How do {Transformers} model physics? {Investigating} the simple harmonic oscillator},
  author={Kantamneni, Subhash and Liu, Ziming and Tegmark, Max},
  journal={Entropy},
  volume={26},
  number={11},
  pages={997},
  year={2024},
  publisher={MDPI}
}

@article{choi2025defining,
  title={Defining foundation models for computational science: A call for clarity and rigor},
  author={Choi, Youngsoo and Cheung, Siu Wun and Kim, Youngkyu and Tsai, Ping-Hsuan and Diaz, Alejandro N and Zanardi, Ivan and Chung, Seung Whan and Copeland, Dylan Matthew and Kendrick, Coleman and Anderson, William and others},
  journal={arXiv preprint arXiv:2505.22904},
  year={2025}
}

@article{hemmer2025true,
  title={True zero-shot inference of dynamical systems preserving long-term statistics},
  author={Hemmer, Christoph J{\"u}rgen and Durstewitz, Daniel},
  journal={arXiv preprint arXiv:2505.13192},
  year={2025}
}

@article{bodnar2025aurora,
  title     = {A foundation model for the {Earth} system},
  author    = {Bodnar, Cristian and Bruinsma, Wessel P. and Lucic, Ana and Stanley, Megan and Allen, Anna and Brandstetter, Johannes and Garvan, Patrick and Riechert, Maik and Weyn, Jonathan A. and Dong, Haiyu and Gupta, Jayesh K. and Thambiratnam, Kit and Archibald, Alexander T. and Wu, Chun-Chieh and Heider, Elizabeth and Welling, Max and Turner, Richard E. and Perdikaris, Paris},
  journal   = {Nature},
  volume    = {641},
  number    = {8065},
  pages     = {1180--1187},
  year      = {2025},
  doi       = {10.1038/s41586-025-09005-y},
  url       = {https://doi.org/10.1038/s41586-025-09005-y}
}

@misc{poseidon2024,
      title={{Poseidon}: Efficient foundation models for {PDEs}}, 
      author={Maximilian Herde and Bogdan Raonić and Tobias Rohner and Roger Käppeli and Roberto Molinaro and Emmanuel de Bézenac and Siddhartha Mishra},
      year={2024},
      eprint={2405.19101},
      archivePrefix={arXiv},
      primaryClass={cs.LG},
      url={https://arxiv.org/abs/2405.19101}, 
}

@misc{chronos2024,
      title={{Chronos}: Learning the language of time series}, 
      author={Abdul Fatir Ansari and Lorenzo Stella and Caner Turkmen and Xiyuan Zhang and Pedro Mercado and Huibin Shen and Oleksandr Shchur and Syama Sundar Rangapuram and Sebastian Pineda Arango and Shubham Kapoor and Jasper Zschiegner and Danielle C. Maddix and Hao Wang and Michael W. Mahoney and Kari Torkkola and Andrew Gordon Wilson and Michael Bohlke-Schneider and Yuyang Wang},
      year={2024},
      eprint={2403.07815},
      archivePrefix={arXiv},
      primaryClass={cs.LG},
      url={https://arxiv.org/abs/2403.07815}, 
}

@misc{cao2021choose,
      title={Choose a {Transformer}: {Fourier} or {Galerkin}}, 
      author={Shuhao Cao},
      year={2021},
      eprint={2105.14995},
      archivePrefix={arXiv},
      primaryClass={cs.LG},
      url={https://arxiv.org/abs/2105.14995}, 
}

@article{kovachki2023neuraloperators,
author = {Kovachki, Nikola and Li, Zongyi and Liu, Burigede and Azizzadenesheli, Kamyar and Bhattacharya, Kaushik and Stuart, Andrew and Anandkumar, Anima},
title = {Neural operator: learning maps between function spaces with applications to {PDEs}},
year = {2023},
issue_date = {January 2023},
publisher = {JMLR.org},
volume = {24},
number = {1},
issn = {1532-4435},
journal = {J. Mach. Learn. Res.},
month = jan,
articleno = {89},
numpages = {97}
}

@InProceedings{takens1981detecting,
author="Takens, Floris",
editor="Rand, David
and Young, Lai-Sang",
title="Detecting strange attractors in turbulence",
booktitle="{Dynamical Systems and Turbulence, Warwick 1980}",
year="1981",
publisher="Springer Berlin Heidelberg",
address="Berlin, Heidelberg",
pages="366--381",
isbn="978-3-540-38945-3"
}

@article{evangelou2023double,
  title={Double diffusion maps and their latent harmonics for scientific computations in latent space},
  author={Evangelou, Nikolaos and Dietrich, Felix and Chiavazzo, Eliodoro and Lehmberg, Daniel and Meila, Marina and Kevrekidis, Ioannis G},
  journal={Journal of Computational Physics},
  volume={485},
  pages={112072},
  year={2023},
  publisher={Elsevier}
}

@inproceedings{gear2011slow,
  title={Slow manifold integration on a diffusion map parameterization},
  author={Gear, Charles William and Kevrekidis, IG and Sonday, BE},
  booktitle={{Numerical Analysis and Applied Mathematics ICNAAM 2011}: International Conference on Numerical Analysis and Applied Mathematics},
  volume={1389},
  number={1},
  pages={13--16},
  year={2011}
}

@article{koronaki2024nonlinear,
  title={Nonlinear dimensionality reduction then and now: {AIMs} for dissipative {PDEs} in the {ML} era},
  author={Koronaki, Eleni D and Evangelou, Nikolaos and Martin-Linares, Cristina P and Titi, Edriss S and Kevrekidis, Ioannis G},
  journal={Journal of Computational Physics},
  volume={506},
  pages={112910},
  year={2024},
  publisher={Elsevier}
}

@article{foias1988computation,
  title={On the computation of inertial manifolds},
  author={Foias, C and Jolly, MS and Kevrekidis, IG and Sell, George R and Titi, ES},
  journal={Physics Letters A},
  volume={131},
  number={7-8},
  pages={433--436},
  year={1988},
  publisher={Elsevier}
}

@article{li2020fourier,
  title={{Fourier} neural operator for parametric partial differential equations},
  author={Li, Zongyi and Kovachki, Nikola and Azizzadenesheli, Kamyar and Liu, Burigede and Bhattacharya, Kaushik and Stuart, Andrew and Anandkumar, Anima},
  journal={arXiv preprint arXiv:2010.08895},
  year={2020}
}

@article{valle2025forecasting,
  title={Forecasting chaotic time series: Comparative performance of {LSTM}-based and {Transformer}-based neural network},
  author={Valle, Joao and Bruno, Odemir Martinez},
  journal={Chaos, Solitons \& Fractals},
  volume={192},
  pages={116034},
  year={2025},
  publisher={Elsevier}
}

@book{box1976analysis,
  title={{Time Series Analysis}: Forecasting and Control, 5th Edition},
  author={George E. P. Box and Gwilym M. Jenkins and Gregory C. Reinsel and Greta M. Ljung },
  year={2015},
  edition={5},
  publisher={Wiley},
  address={Hoboken, NJ}
}

@book{ljung1999systemid,
  title     = {{System Identification}: Theory for the User},
  author    = {Lennart Ljung},
  edition   = {2},
  year      = {1999},
  publisher = {Prentice Hall},
  address   = {Upper Saddle River, NJ}
}

@article{elhage2022toy,
  title={Toy models of superposition},
  author={Elhage, Nelson and Hume, Tristan and Olsson, Catherine and Schiefer, Nicholas and Henighan, Tom and Kravec, Shauna and Hatfield-Dodds, Zac and Lasenby, Robert and Drain, Dawn and Chen, Carol and others},
  journal={arXiv preprint arXiv:2209.10652},
  year={2022}
}

@article{elhage2021mathematical,
  title={A mathematical framework for {Transformer} circuits},
  author={Elhage, Nelson and Nanda, Neel and Olsson, Catherine and Henighan, Tom and Joseph, Nicholas and Mann, Ben and Askell, Amanda and Bai, Yuntao and Chen, Anna and Conerly, Tom and others},
  journal={Transformer Circuits Thread},
  volume={1},
  number={1},
  pages={12},
  year={2021}
}

@inproceedings{katharopoulos2020transformers,
author = {Katharopoulos, Angelos and Vyas, Apoorv and Pappas, Nikolaos and Fleuret, Fran\c{c}ois},
title = {{Transformers} are {RNNs}: fast autoregressive transformers with linear attention},
year = {2020},
publisher = {JMLR.org},
booktitle = {Proceedings of the 37th {International Conference on Machine Learning}},
articleno = {478},
numpages = {10},
series = {ICML'20}
}

@inproceedings{dao2024ssd,
author = {Dao, Tri and Gu, Albert},
title = {{Transformers} are {SSMs}: generalized models and efficient algorithms through structured state space duality},
year = {2024},
publisher = {JMLR.org},
booktitle = {Proceedings of the 41st {International Conference on Machine Learning}},
articleno = {399},
numpages = {31},
location = {Vienna, Austria},
series = {ICML'24}
}

@misc{sieber2024dsf,
      title={Understanding the differences in foundation models: attention, state space models, and recurrent neural networks}, 
      author={Jerome Sieber and Carmen Amo Alonso and Alexandre Didier and Melanie N. Zeilinger and Antonio Orvieto},
      year={2024},
      eprint={2405.15731},
      archivePrefix={arXiv},
      primaryClass={cs.LG},
      url={https://arxiv.org/abs/2405.15731}, 
}

@book{Kantz_Schreiber_2003, place={Cambridge}, edition={2}, title={Nonlinear time series analysis}, publisher={Cambridge University Press}, author={Kantz, Holger and Schreiber, Thomas}, year={2003}}

@article{10.1063/5.0297336,
    author = {Panja, Madhurima and Das, Dhiman and Chakraborty, Tanujit and Ray, Arnob and Athulya, R. and Hens, Chittaranjan and Dana, Syamal K. and Murukesh, Nuncio and Ghosh, Dibakar},
    title = {Forecasting precipitation in the {Arctic} using probabilistic machine learning informed by causal climate drivers},
    journal = {Chaos: An Interdisciplinary Journal of Nonlinear Science},
    volume = {35},
    number = {11},
    pages = {113108},
    year = {2025},
    month = {11},
    issn = {1054-1500},
    doi = {10.1063/5.0297336},
    url = {https://doi.org/10.1063/5.0297336},}

@article{10.1063/5.0291493,
    author = {Ghosh, Indranil and Fatoyinbo, Hammed O. and Muni, Sishu S.},
    title = {Time series analysis of coupled slow--fast neuron models: From {Hurst} exponent to {Granger} causality},
    journal = {Chaos: An Interdisciplinary Journal of Nonlinear Science},
    volume = {35},
    number = {10},
    pages = {103136},
    year = {2025},
    month = {10},
    issn = {1054-1500},
    doi = {10.1063/5.0291493},
    url = {https://doi.org/10.1063/5.0291493},}

@article{LIU2022109276,
title = {Physics-guided deep {Markov} models for learning nonlinear dynamical systems with uncertainty},
journal = {Mechanical Systems and Signal Processing},
volume = {178},
pages = {109276},
year = {2022},
issn = {0888-3270},
doi = {10.1016/j.ymssp.2022.109276},
url = {https://www.sciencedirect.com/science/article/pii/S0888327022004204},
author = {Wei Liu and Zhilu Lai and Kiran Bacsa and Eleni Chatzi}
}

@misc{amoudruz2025,
      title={{Bayesian} inference for {PDE}-based inverse problems using the optimization of a discrete loss}, 
      author={Lucas Amoudruz and Sergey Litvinov and Costas Papadimitriou and Petros Koumoutsakos},
      year={2025},
      eprint={2510.15664},
      archivePrefix={arXiv},
      primaryClass={stat.ME},
      url={https://arxiv.org/abs/2510.15664}, 
}

@article{RAISSI2019686,
title = {Physics-informed neural networks: A deep learning framework for solving forward and inverse problems involving nonlinear partial differential equations},
journal = {Journal of Computational Physics},
volume = {378},
pages = {686-707},
year = {2019},
issn = {0021-9991},
doi = {10.1016/j.jcp.2018.10.045},
url = {https://www.sciencedirect.com/science/article/pii/S0021999118307125},
author = {M. Raissi and P. Perdikaris and G.E. Karniadakis}
}

@misc{lutter2019,
      title={Deep {Lagrangian} networks: using physics as model prior for deep learning}, 
      author={Michael Lutter and Christian Ritter and Jan Peters},
      year={2019},
      eprint={1907.04490},
      archivePrefix={arXiv},
      primaryClass={cs.LG},
      url={https://arxiv.org/abs/1907.04490}, 
}

@article{Bertalan2019,
    author = {Bertalan, Tom and Dietrich, Felix and Mezić, Igor and Kevrekidis, Ioannis G.},
    title = {On learning {Hamiltonian} systems from data},
    journal = {Chaos: An Interdisciplinary Journal of Nonlinear Science},
    volume = {29},
    number = {12},
    pages = {121107},
    year = {2019},
    month = {12},
    issn = {1054-1500},
    doi = {10.1063/1.5128231},
    url = {https://doi.org/10.1063/1.5128231},}

@article{Bacsa2023,
  title={Symplectic encoders for physics-constrained variational dynamics inference},
  author={Bacsa, Kiran and Lai, Zhilu and Liu, Wei and Todd, Michael and Chatzi, Eleni},
  journal={Scientific Reports},
  volume={13},
  number={1},
  pages={2643},
  year={2023},
  doi={10.1038/s41598-023-29186-8},
  publisher={Nature Publishing Group}
}

@article{HERNANDEZ2021109950,
title = {Structure-preserving neural networks},
journal = {Journal of Computational Physics},
volume = {426},
pages = {109950},
year = {2021},
issn = {0021-9991},
doi = {10.1016/j.jcp.2020.109950},
url = {https://www.sciencedirect.com/science/article/pii/S0021999120307245},
author = {Quercus Hernández and Alberto Badías and David González and Francisco Chinesta and Elías Cueto}
}

@book{lafon2004diffusion,
  title={Diffusion maps and geometric harmonics},
  author={Lafon, St{\'e}phane S},
  year={2004},
  publisher={Yale University}
}

@article{deane1991low,
  title={Low-dimensional models for complex geometry flows: application to grooved channels and circular cylinders},
  author={Deane, Anil E and Kevrekidis, Ioannis G and Karniadakis, George Em and Orszag, SA0746},
  journal={Physics of Fluids A: Fluid Dynamics},
  volume={3},
  number={10},
  pages={2337--2354},
  year={1991},
  publisher={American Institute of Physics}
}

@article{fear2025physics,
  title={Physics steering: causal control of cross-domain concepts in a physics foundation model},
  author={Fear, Rio Alexa and Mukhopadhyay, Payel and McCabe, Michael and Bietti, Alberto and Cranmer, Miles},
  journal={arXiv preprint arXiv:2511.20798},
  year={2025}
}

@article{chiavazzo2014reduced,
  title={Reduced models in chemical kinetics via nonlinear data-mining},
  author={Chiavazzo, Eliodoro and Gear, Charles W and Dsilva, Carmeline J and Rabin, Neta and Kevrekidis, Ioannis G},
  journal={Processes},
  volume={2},
  number={1},
  pages={112--140},
  year={2014},
  publisher={MDPI}
}

@misc{ansari2025chronos2univariateuniversalforecasting,
      title={{Chronos-2}: From univariate to universal forecasting}, 
      author={Abdul Fatir Ansari and Oleksandr Shchur and Jaris Küken and Andreas Auer and Boran Han and Pedro Mercado and Syama Sundar Rangapuram and Huibin Shen and Lorenzo Stella and Xiyuan Zhang and Mononito Goswami and Shubham Kapoor and Danielle C. Maddix and Pablo Guerron and Tony Hu and Junming Yin and Nick Erickson and Prateek Mutalik Desai and Hao Wang and Huzefa Rangwala and George Karypis and Yuyang Wang and Michael Bohlke-Schneider},
      year={2025},
      eprint={2510.15821},
      archivePrefix={arXiv},
      primaryClass={cs.LG},
      url={https://arxiv.org/abs/2510.15821}, 
}

@inproceedings{menier2025interpretable,
  title={Interpretable learning of effective dynamics for multiscale systems},
  author={Menier, Emmanuel and Kaltenbach, Sebastian and Yagoubi, Mouadh and Schoenauer, Marc and Koumoutsakos, Petros},
  booktitle={Proceedings {A}},
  volume={481},
  number={2305},
  pages={20240167},
  year={2025},
  organization={The Royal Society}
}

@article{raschka2023understanding,
  title={Understanding and coding the self-attention mechanism of large language models from scratch},
  author={Raschka, S},
  journal={Blog of S. Raschka, https://sebastianraschka. com/blog/2023/self-attention-from-scratch. html (accessed December 2025)},
  year={2023}
}

@InProceedings{pmlr-v267-holzschuh25a,
  title = 	 {{PDE}-{Transformer}: Efficient and versatile {Transformers} for physics simulations},
  author =       {Holzschuh, Benjamin and Liu, Qiang and Kohl, Georg and Thuerey, Nils},
  booktitle = 	 {Proceedings of the 42nd {International Conference on Machine Learning}},
  pages = 	 {23562--23602},
  year = 	 {2025},
  editor = 	 {Singh, Aarti and Fazel, Maryam and Hsu, Daniel and Lacoste-Julien, Simon and Berkenkamp, Felix and Maharaj, Tegan and Wagstaff, Kiri and Zhu, Jerry},
  volume = 	 {267},
  series = 	 {Proceedings of Machine Learning Research},
  month = 	 {13--19 Jul},
  publisher =    {PMLR},
  url = 	 {https://proceedings.mlr.press/v267/holzschuh25a.html},
}

@inproceedings{
gao2025learning,
title={Learning effective dynamics across spatio-temporal scales of complex flows},
author={Han Gao and Sebastian Kaltenbach and Petros Koumoutsakos},
booktitle={The Second Conference on {Parsimony and Learning} (Proceedings Track)},
year={2025},
url={https://openreview.net/forum?id=TU9e5yChcU}
}

@article{wu2023interpretable,
  title={Interpretable weather forecasting for worldwide stations with a unified deep model},
  author={Wu, Haixu and Zhou, Hang and Long, Mingsheng and Wang, Jianmin},
  journal={Nature Machine Intelligence},
  volume={5},
  number={6},
  pages={602--611},
  year={2023},
  publisher={Nature Publishing Group UK London}
}

@article{shih2025transformers,
  title={{Transformers} as neural operators for solutions of differential equations with finite regularity},
  author={Shih, Benjamin and Peyvan, Ahmad and Zhang, Zhongqiang and Karniadakis, George Em},
  journal={Computer Methods in Applied Mechanics and Engineering},
  volume={434},
  pages={117560},
  year={2025},
  publisher={Elsevier}
}

@article{sitapure2023introducing,
  title={Introducing hybrid modeling with time-series-transformers: A comparative study of series and parallel approach in batch crystallization},
  author={Sitapure, Niranjan and Sang-Il Kwon, Joseph},
  journal={Industrial \& Engineering Chemistry Research},
  volume={62},
  number={49},
  pages={21278--21291},
  year={2023},
  publisher={ACS Publications}
}

@article{urdeitx2025can,
  title={Can {Transformers} overcome the lack of data in the simulation of history-dependent flows?},
  author={Urdeitx, P and Alfaro, I and Gonzalez, D and Chinesta, F and Cueto, E},
  journal={arXiv preprint arXiv:2512.16305},
  year={2025}
}

@article{rico1992discrete,
  title={Discrete-vs. continuous-time nonlinear signal processing of {Cu} electrodissolution data},
  author={Rico-Martinez, Ramiro and Krischer, K and Kevrekidis, IG and Kube, MC and Hudson, JL},
  journal={Chemical Engineering Communications},
  volume={118},
  number={1},
  pages={25--48},
  year={1992},
  publisher={Taylor \& Francis}
}

@inproceedings{cui2023certified,
  title={Certified invertibility in neural networks via mixed-integer programming},
  author={Cui, Tianqi and Bertalan, Thomas and Pappas, George J and Morari, Manfred and Kevrekidis, Yannis and Fazlyab, Mahyar},
  booktitle={{Learning for Dynamics and Control Conference}},
  pages={483--496},
  year={2023},
  organization={PMLR}
}

@article{Zhu2023ImplementationA,
  title={Implementation and ({Inverse Modified}) error analysis for implicitly-templated {ODE}-nets},
  author={Aiqing Zhu and Tom S. Bertalan and Beibei Zhu and Yifa Tang and Ioannis G. Kevrekidis},
  journal={SIAM J. Appl. Dyn. Syst.},
  year={2023},
  volume={23},
  pages={2643-2669},
  url={https://api.semanticscholar.org/CorpusID:257900843}
}

\newpage
\appendix

\setcounter{figure}{0}
\renewcommand{\thefigure}{S\arabic{figure}}

\section{Additional Results}

\subsection{Two-head attention for Case 2 of the 1D linear system}
\label{sec:two_head_sdof_appendix}

The single-head SDOF analysis in Section~\ref{sec:lin_systems} was used as a minimal mechanistic setting to isolate the mechanism of the attention head. It was not intended to imply that the same restriction necessarily applies to practical Transformer architectures, which typically use multiple attention heads and additional nonlinear components. To clarify this point, we include a two-head attention-only example for Case 2 of the 1D linear system in the main text. This is the mixed-sign autoregressive case, where the underlying AR representation requires one positive and one negative lag coefficient.

Following the notation in Section~\ref{sec:lin_systems}, for a single attention head with two delayed inputs the prediction can be written as
\begin{equation}
\hat{x}
=
w_o\left[\alpha_{2,1}(x_1+p_1)+\alpha_{2,2}(x_2+p_2)\right],
\end{equation}
where $\alpha_{2,1},\alpha_{2,2}>0$, $\alpha_{2,1}+\alpha_{2,2}=1$, and $w_o$ denotes the scalar output projection coefficient in the 1D setting. Rearranging gives
\begin{equation}
\hat{x}
=
\beta_1 x_1+\beta_2 x_2+C,
\qquad
\beta_i=\alpha_{2,i}w_o.
\end{equation}
Since both $\alpha_{2,1}$ and $\alpha_{2,2}$ are positive and both coefficients are scaled by the same scalar $w_o$, $\beta_1$ and $\beta_2$ must have the same sign. This is why a single head cannot directly represent the mixed-sign AR behavior required by Case 2 of the 1D linear system.

For two attention heads, each head forms its own convex combination of the delayed inputs. The two head outputs are then combined through the output projection. Denoting the attention weights of head $r$ by $\alpha_{2,i}^{(r)}$, and writing the output projection as $W_O=[w_{o,1},w_{o,2}]$, where $w_{o,r}$ is the output projection coefficient associated with head $r$, the two-head prediction can be written as
\begin{equation}
\hat{x} = \sum_{r=1}^{2} w_{o,r}
\left[
\alpha_{2,1}^{(r)}(x_1+p_1)
+
\alpha_{2,2}^{(r)}(x_2+p_2)
\right].
\end{equation}

Rearranging gives
\begin{equation}
\hat{x}
=
\beta_1 x_1+\beta_2 x_2+C,
\end{equation}
with
\begin{equation}
\beta_i
=
\sum_{r=1}^{2}
w_{o,r}\alpha_{2,i}^{(r)},
\qquad i=1,2.
\label{eq:two_head_effective_coeffs_simple}
\end{equation}
Thus, although each $\alpha_{2,i}^{(r)}$ is still positive due to the softmax operation, the final effective coefficients $\beta_i$ are no longer restricted to be positive scalar multiples of the same value. This is because different heads are combined through different output projection coefficients $w_{o,r}$.

In the trained two-head model for Case 2 of the 1D linear system, the learned output projection coefficients are $w_{o,1}=1.087$ and $w_{o,2}=-1.244$. This shows that the output layer combines the two head outputs with opposite signs, allowing the model to form effective AR coefficients with mixed signs even though the attention weights inside each head remain positive.

\begin{figure}[h]
\centering
\includegraphics[width=0.33\linewidth]{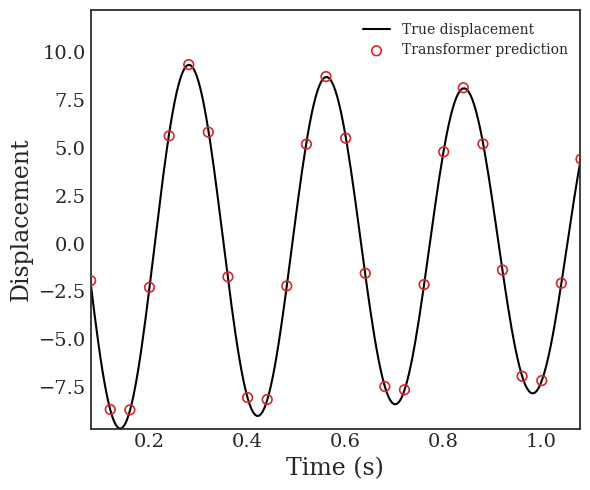}
\caption{
Two-head attention-only Transformer on Case 2 of the 1D linear system, which exhibits mixed-sign autoregressive behavior.
Unlike the single-head model, the two-head model reproduces the response.
The learned output projection is $W_O=[1.087,-1.244]$, indicating that the two head outputs are combined with opposite signs.
}
\label{fig:1D_2head}
\end{figure}

Figure~\ref{fig:1D_2head} shows the corresponding numerical result. In contrast to the single-head model for Case 2, the two-head model correctly reproduces the response. This example illustrates that the single-head result should be interpreted as a mechanistic boundary case, while multi-head attention can circumvent the mixed-sign restriction through signed head-wise recombination.

\subsection{Van der Pol}

\subsubsection{NARX baseline comparison}
\label{app:vdp_narx}

As an additional baseline to the deep learning models we report in the main text, we compare the MLP and Transformer models discussed in Section \ref{sec:Van_der_Pol} against a classical polynomial NARX model. The NARX model is fit using the FROLS (Forward Regression Orthogonal Least Squares) algorithm from the \texttt{sysidentpy} library, with a third-degree polynomial basis and the same five-delay window used throughout our Transformer experiments. FROLS is deterministic given the training data and automatically selects the most significant regressors. Consequently, unlike the neural models, no seed variability is reported for NARX.

We consider the same two observability regimes as in \ref{sec:Van_der_Pol}. Under full observation, we fit a paired NARX model using $x$ and $\dot{x}$ as inputs to one another. Under partial observation, we fit a model using only delayed values of $x$. Table~\ref{tab:narx_vdp} reports the resulting one-step-ahead MSE on the limit-cycle test trajectory.

\begin{table}[h]
\centering
\caption{One-step-ahead MSE on the Van der Pol limit-cycle test trajectory,
comparing the MLP and Transformer models against a polynomial NARX baseline
(FROLS, degree 3, 5 delays). NARX is deterministic given the training data, so
no standard deviation is reported.}
\label{tab:narx_vdp}
\begin{tabular}{lcc}
\hline
\textbf{Model} & \textbf{Mean MSE} & \textbf{Std MSE} \\
\hline
\multicolumn{3}{l}{\textit{Full Observation (2D)}} \\
MLP Only                  & $1.455\times 10^{-07}$ & $1.459\times 10^{-07}$ \\
MLP + Attn 2D (No P.E.)   & $2.765\times 10^{-07}$ & $1.452\times 10^{-07}$ \\
MLP + Attn 2D (P.E.)      & $9.504\times 10^{-08}$ & $1.157\times 10^{-07}$ \\
NARX (poly, FROLS)        & $7.107\times 10^{-04}$ & $0.000\times 10^{+00}$ \\
\hline
\multicolumn{3}{l}{\textit{Partial Observation (1D)}} \\
MLP Only                  & $2.485\times 10^{-02}$ & $1.564\times 10^{-06}$ \\
MLP + Attn 1D (No P.E.)   & $1.789\times 10^{-03}$ & $2.772\times 10^{-06}$ \\
MLP + Attn 2D (No P.E.)   & $1.770\times 10^{-03}$ & $5.505\times 10^{-05}$ \\
MLP + Attn 1D (P.E.)      & $5.848\times 10^{-04}$ & $2.259\times 10^{-05}$ \\
MLP + Attn 2D (P.E.)      & $2.065\times 10^{-05}$ & $9.495\times 10^{-06}$ \\
NARX (poly, FROLS)        & $2.192\times 10^{-03}$ & $0.000\times 10^{+00}$ \\
\hline
\end{tabular}
\end{table}

Consistent with the discussion in Section~\ref{sec:Van_der_Pol}, the NARX baseline performs
markedly worse than all deep learning models under full observation, where the dominant difficulty lies in the nonlinearity of the time-one map rather than in any historical dependence a delay-embedding scheme could exploit.

Under partial observation, NARX performs much better than the MLP-only models and is comparable to the Transformer variants without PE, but is outperformed by the best-performing
Transformer ($d_{\mathrm{lat}}=2$ with PE) by roughly
two orders of magnitude, indicating that the data-adaptive nature of
attention could provide a more expressive reconstruction of the missing state than
a fixed-degree polynomial expansion of the delay vector.

\subsubsection{Transformer Modeling Results}
\label{sec:van_der_pol_appendix}
In this section, we provide additional complementary results for the Van der Pol oscillator described in Section~\ref{sec:Van_der_Pol}.

Specifically for the models we focused in the main text we provide additional visualizations. 

\begin{figure}[h]
    \centering
    \includegraphics[width=0.95\linewidth]{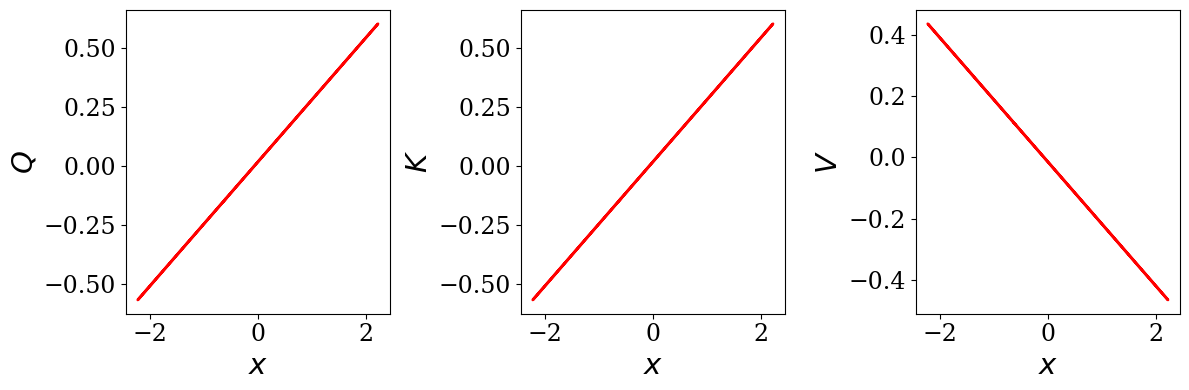}
    \caption{
    Visualization of attention-related latent variables for one of the Transformers we trained with $d_{\mathrm{lat}}=1$ and without PE. }    \label{fig:van_der_pol_latents_no_pe}
\end{figure}

\begin{figure}[h]
    \centering
    \includegraphics[width=0.95\linewidth]{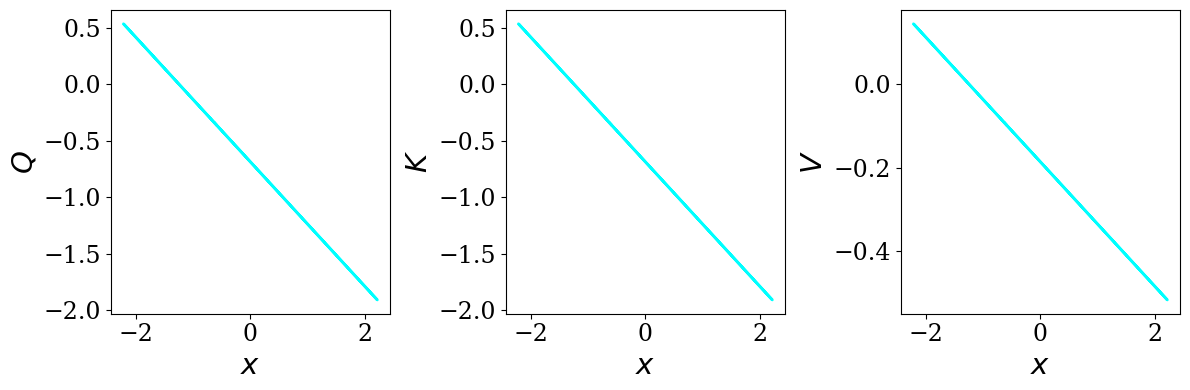}
    \caption{
    Visualization of attention-related latent variables for the Transformer with $d_{\mathrm{lat}}=1$ we trained with PE.
    }
    \label{fig:van_der_pol_latents_pe}
\end{figure}

For the model trained with latent dimension $d_{\mathrm{lat}}=1$, in Figures \ref{fig:van_der_pol_latents_no_pe} and \ref{fig:van_der_pol_latents_pe}
we plot \( x(t) \) against the query, key, and value vectors for the models trained  with PE and without PE. As discussed in the main text, we find that visual inspection of these components does not offer additional interpretability for our use cases.

\begin{figure}[h]
    \centering
    \includegraphics[width=\linewidth]{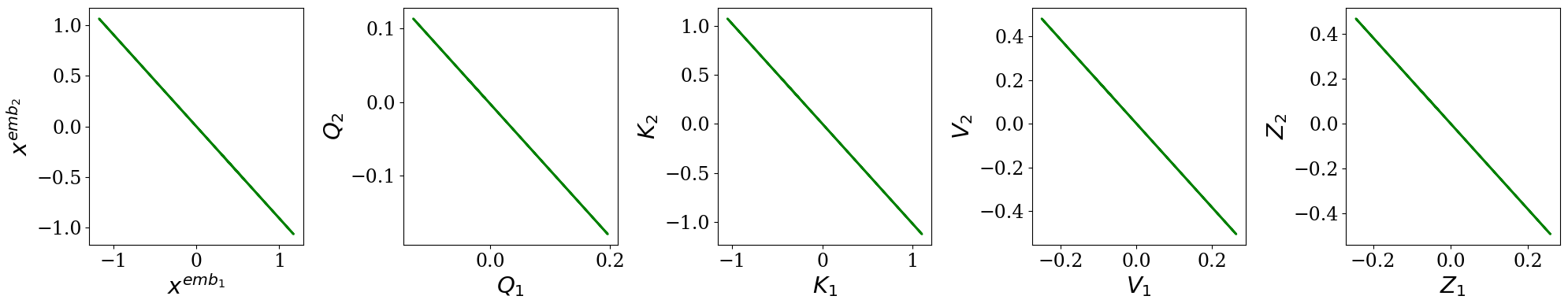}
    \caption{
        Visualization of the learned latent space for the Transformer model with $d_{\mathrm{lat}}=2$ without PE.
    }
    \label{fig:van_der_pol_2d_latent_no_pe}
\end{figure}

\begin{figure}[h]
    \centering
    \includegraphics[width=\linewidth]{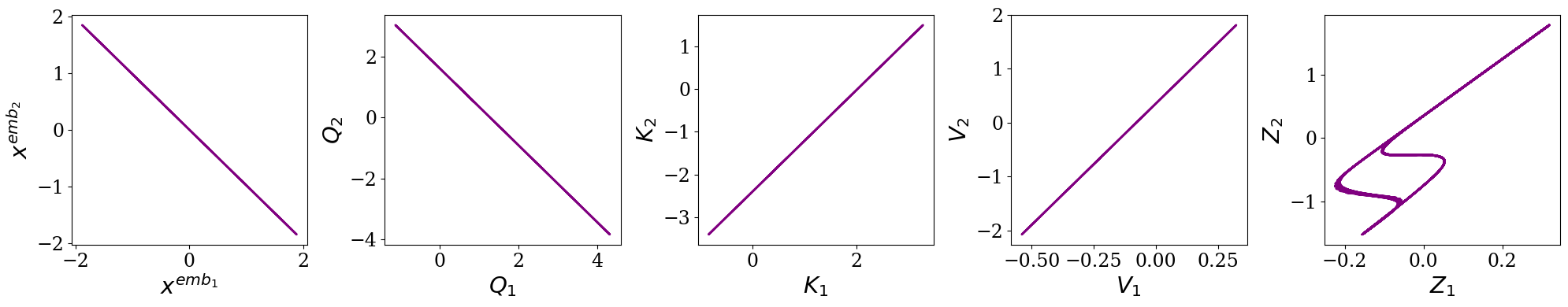}
    \caption{
        Visualization of the learned latent space for the Transformer model with $d_{\mathrm{lat}}=2$ with PE.
    }
    \label{fig:van_der_pol_2d_latent_pe}
\end{figure}

We also include visualizations for models trained with a latent dimension $d_{\mathrm{lat}}=2$, Figures \ref{fig:van_der_pol_2d_latent_no_pe} and \ref{fig:van_der_pol_2d_latent_pe}, in which the embedded inputs, queries, values, and correction terms are all plotted in the latent space with $d_{\mathrm{lat}}=2$.

\vspace{0.5cm}

\begin{figure}[h]
    \centering
    \begin{subfigure}{0.48\linewidth}
        \centering
        \includegraphics[width=\linewidth]{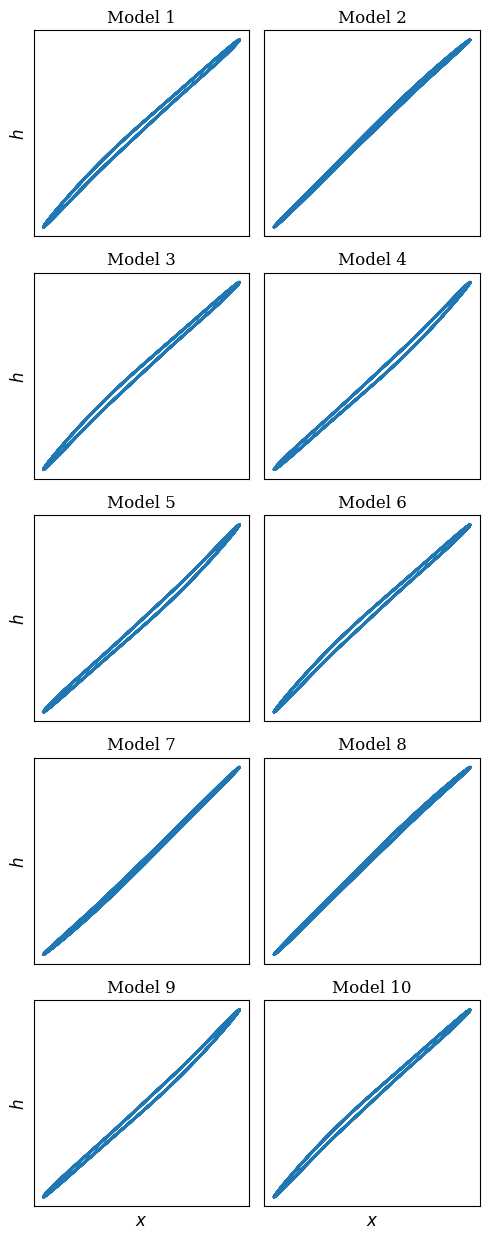}
        \caption{}
    \end{subfigure}
    \hfill
    \begin{subfigure}{0.48\linewidth}
        \centering
        \includegraphics[width=\linewidth]{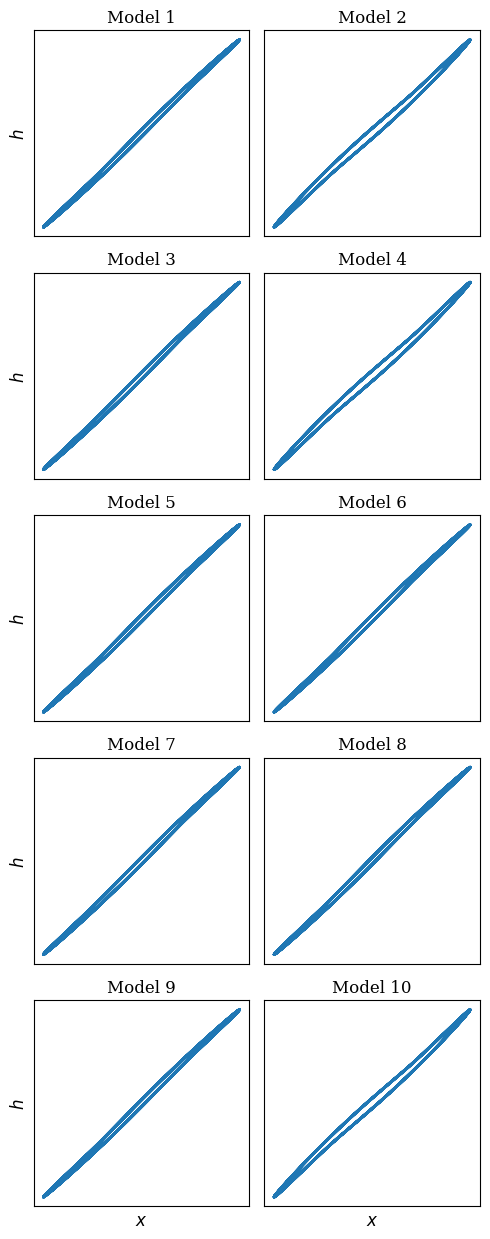}
        \caption{}
    \end{subfigure}
    \caption
   {
   For the 10 models with $d_{\mathrm{lat}}=1$ trained with different random seeds, we inspect their internal embeddings by plotting
    $x(t)$ versus $\mathbf{h}_t$:
    (a) model with PE;
    (b) model without PE.
    }
    \label{fig:10_Models_1D_Transformer_LearnedPE_Latent_Space}
\end{figure}

We also report the internal embeddings across all models  that we have trained for the Van der Pol Oscillator. For the models trained with $d_{\mathrm{lat}}=1$ we plot $x(t)$ against the effective latent coordinate $\mathbf{h}_t$ in Figure \ref{fig:10_Models_1D_Transformer_LearnedPE_Latent_Space}. As it becomes evident, the characteristic limit-cycle structure of the Van der Pol oscillator begins to emerge in the learned representation.

\begin{figure}[h]
    \centering
    \begin{subfigure}{0.48\linewidth}
        \centering
        \includegraphics[width=\linewidth]{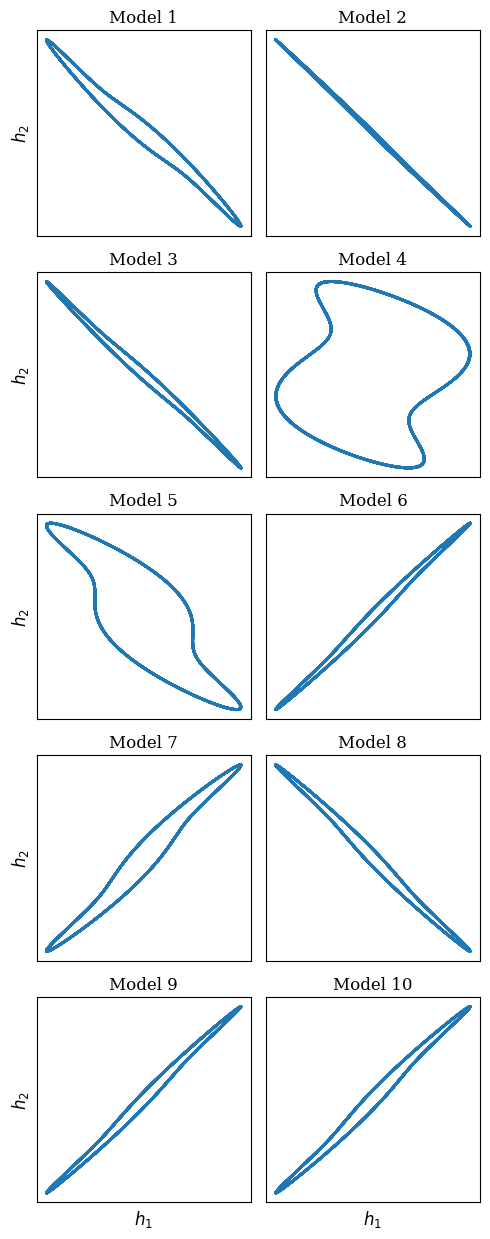}
        \caption{}
    \end{subfigure}
    \hfill
    \begin{subfigure}{0.48\linewidth}
        \centering
        \includegraphics[width=\linewidth]{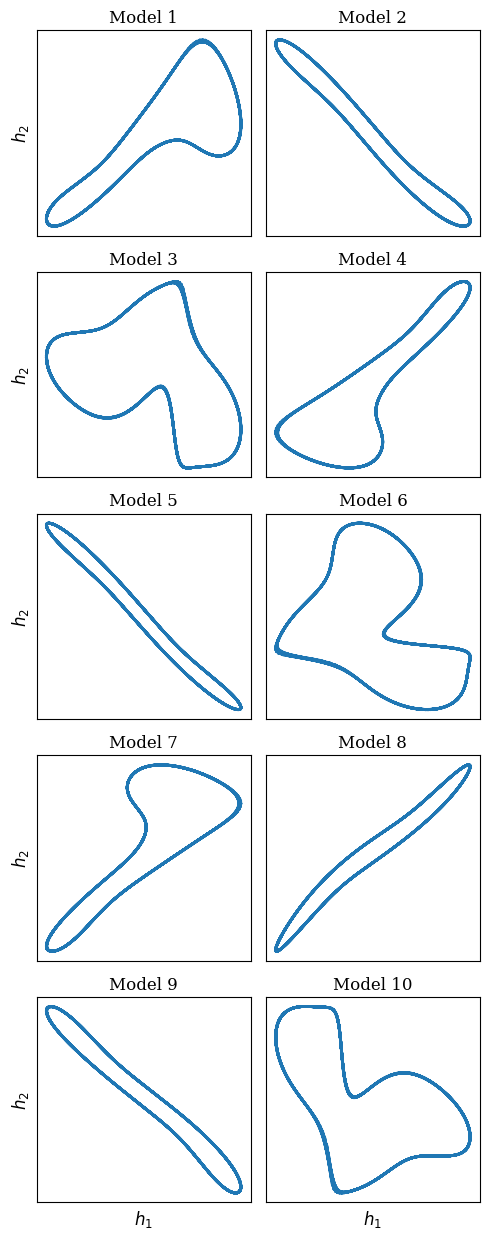}
        \caption{}
    \end{subfigure}
    \caption
   {
   For the 10 models with $d_{\mathrm{lat}}=2$ trained with different random seeds, we inspect their internal embeddings by plotting
    $h_{t,1}$ versus $h_{t,2}$:
    (a) model with PE;
    (b) model without PE.
    }
    \label{fig:10_Models_2D_Transformer_LearnedPE_Latent_Space}
\end{figure}

For the models with $d_{\mathrm{lat}}=2$, Fig.~\ref{fig:10_Models_2D_Transformer_LearnedPE_Latent_Space} further shows that the learned embedding reveals the limit cycles. In this case, the limit-cycle structure is more evident than in the $d_{\mathrm{lat}}=1$ case.

\clearpage
\subsection{Chafee-Infante}
\label{sec:chafee_one_variable_extended}
We provide additional complementary results for the Chafee-Infante equation discussed in Section~\ref{sec:Chafee_Infante} of the main text.

As in the Van der Pol case, we focus on additional visualizations for the models analyzed in the main text. 

\begin{figure}[ht!]
    \centering
\includegraphics[width=0.95\textwidth]{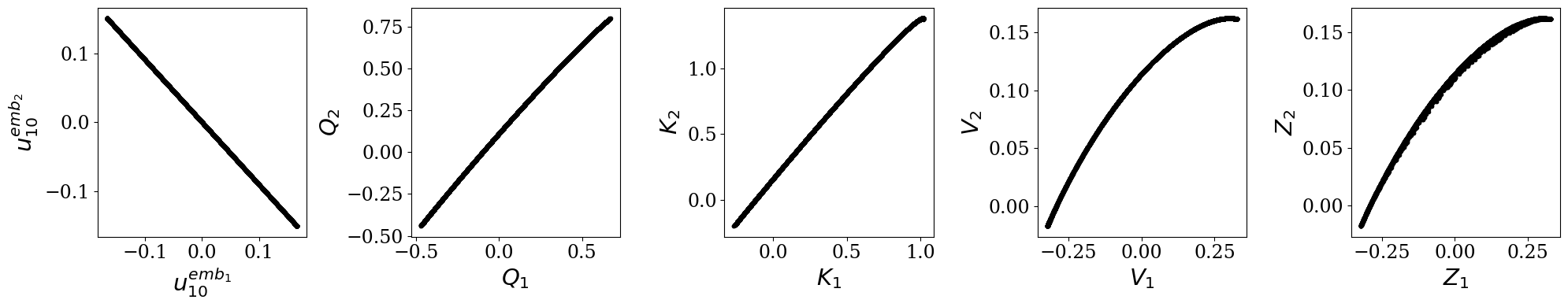}
    \caption{Visualization of attention-related latent variables for one of the Transformers we trained with $d_{\mathrm{lat}}=3$ and without PE.}\label{fig:si_latent_diagnostics}
\end{figure}

\begin{figure}[ht!]
    \centering
\includegraphics[width=0.95\textwidth]{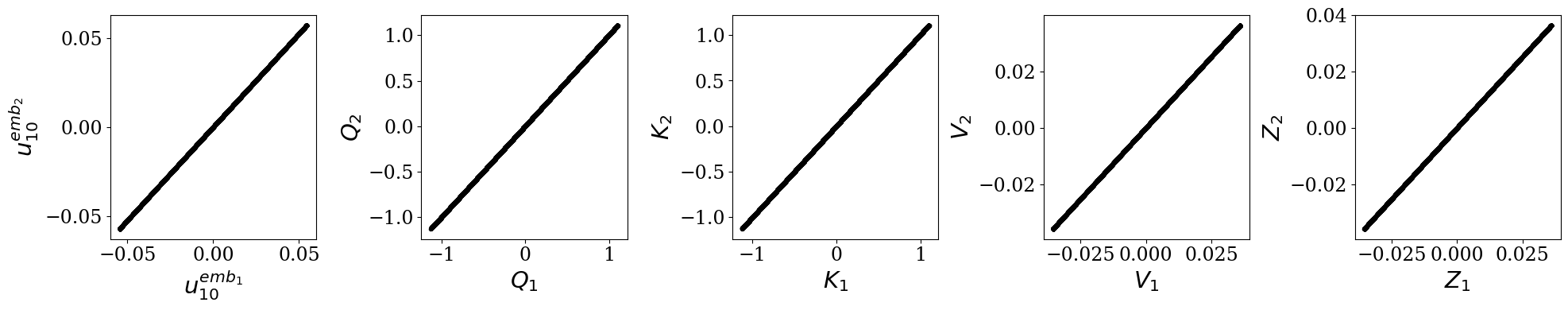}
        \caption{Visualization of attention-related latent variables for one of the Transformers we trained with $d_{\mathrm{lat}}=2$ and without PE.}
    \label{fig:ci2d_latent_ukq}
\end{figure}

We first report visualizations of the query (Q), key (K), value (V), and correction terms for representative models with $d_{\mathrm{lat}}=3$ and $d_{\mathrm{lat}}=2$ latent spaces in Figures \ref{fig:si_latent_diagnostics} and \ref{fig:ci2d_latent_ukq} respectively.

\begin{figure}[h]
    \centering
    \begin{subfigure}{0.43\linewidth}
        \centering
        \includegraphics[width=\linewidth]{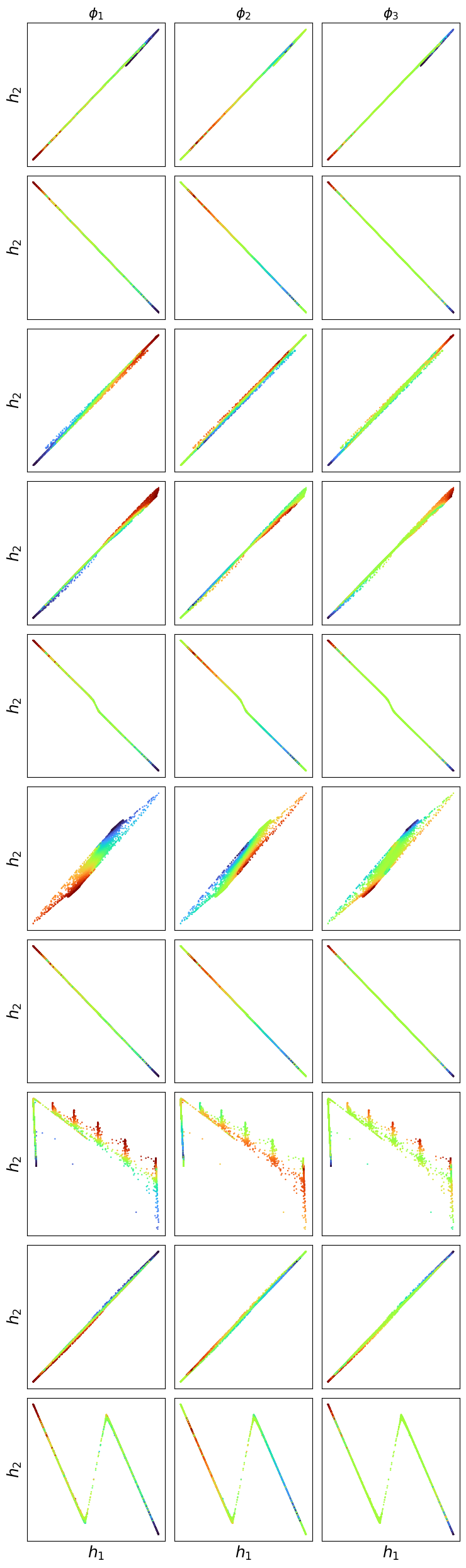}
        \caption{}
    \end{subfigure}
    \hfill
    \begin{subfigure}{0.43\linewidth}
        \centering
        \includegraphics[width=\linewidth]{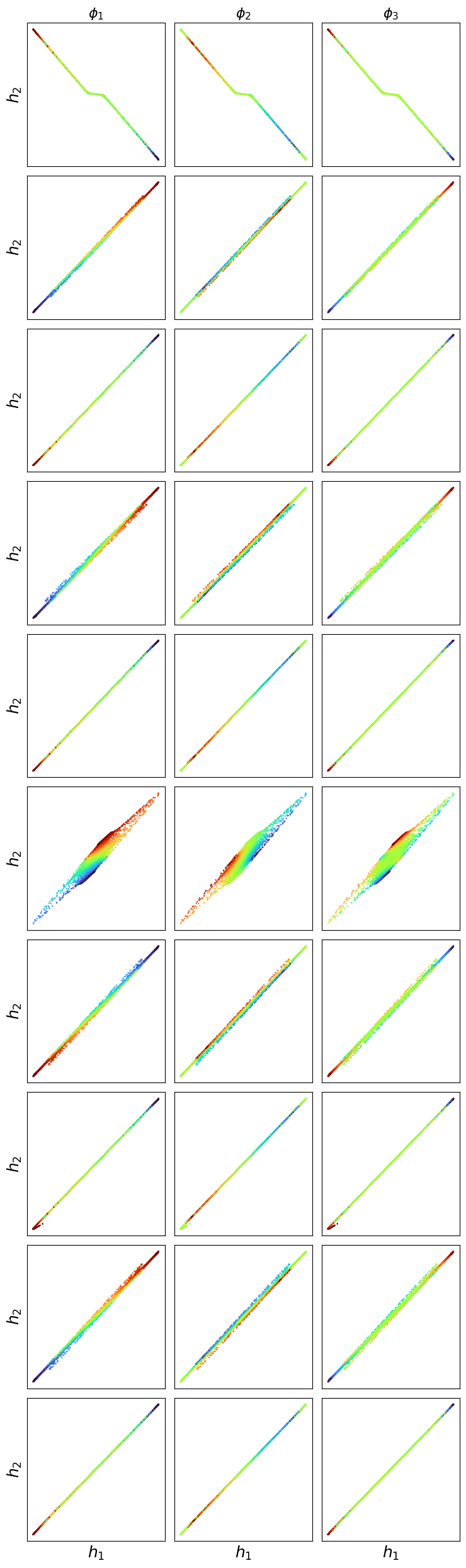}
        \caption{}
    \end{subfigure}
    \caption
   {
   For the 10 models with $d_{\mathrm{lat}}=2$ trained with different random seeds, we inspect their internal embeddings by plotting $\mathbf{h}_t=\mathbf{e}_t+\mathbf{z}_t$ colored by the Fourier modes $\phi_1,\phi_2,\phi_3$
    (a) model with PE;
    (b) model without PE.
    }\label{fig:CI_10_Models_2D_Transformer_LearnedPE_Latent_Space}
\end{figure}

\begin{figure}[h]
    \centering
    \begin{subfigure}{0.43\linewidth}
        \centering
        \includegraphics[width=\linewidth]{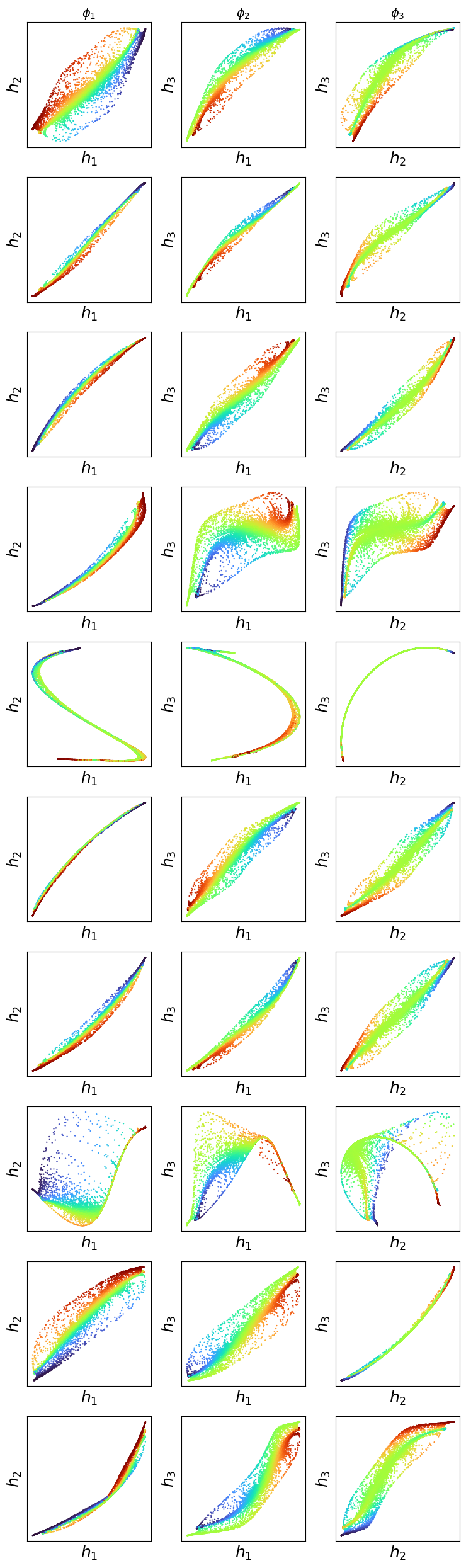}
        \caption{}
    \end{subfigure}
    \hfill
    \begin{subfigure}{0.43\linewidth}
        \centering
        \includegraphics[width=\linewidth]{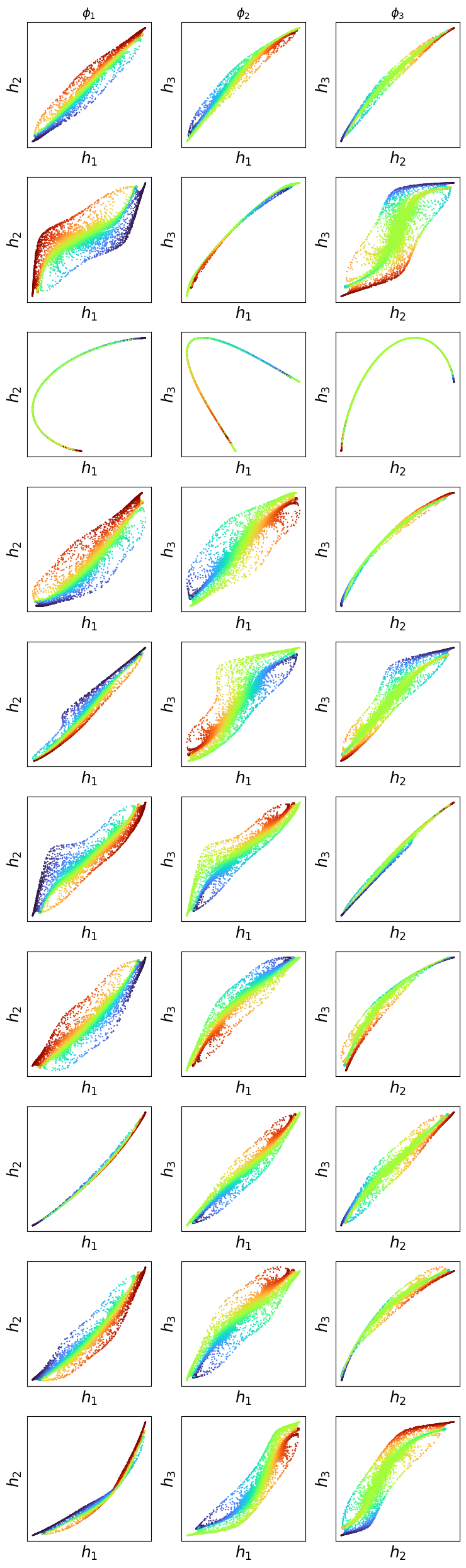}
        \caption{}
    \end{subfigure}
    \caption
   {
   For the 10 models with $d_{\mathrm{lat}}=3$ trained with different random seeds, we inspect their internal embeddings by plotting $\mathbf{h}_t=\mathbf{e}_t+\mathbf{z}_t$ colored by the Fourier modes $\phi_1,\phi_2,\phi_3$
    (a) model with PE;
    (b) model without PE.
    }
\label{fig:CI_10_Models_3D_Transformer_LearnedPE_Latent_Space}
\end{figure}

We also examine the consistency of the learned internal embeddings across all trained models by plotting $\mathbf{h}_t=\mathbf{e}_t+\mathbf{z}_t$, colored by the leading Fourier modes $\vect{\phi}$. The models trained with a latent dimension $d_{\mathrm{lat}}=2$ are shown in Fig.~\ref{fig:CI_10_Models_2D_Transformer_LearnedPE_Latent_Space}. For none of these models does visual inspection of the latent space reveal a clearly unfolded two-dimensional structure. Nevertheless, the Fourier modes still exhibit a discernible relationship with the latent variables.
For the models trained with a latent dimension $d_{\mathrm{lat}}=3$,  in Figure~\ref{fig:CI_10_Models_3D_Transformer_LearnedPE_Latent_Space}, visual inspection of the latent space (based on two-dimensional projections) indicates a clearly unfolded three-dimensional structure across all models (except one). In this case, the leading Fourier modes exhibit a more clear association with the latent variables across all the different random seeds.

\subsection{Chafee-Infante Folding}
\label{sec:chafee_infante_folding}
\begin{figure}[ht!]
    \centering
    \includegraphics[width=0.95\textwidth]{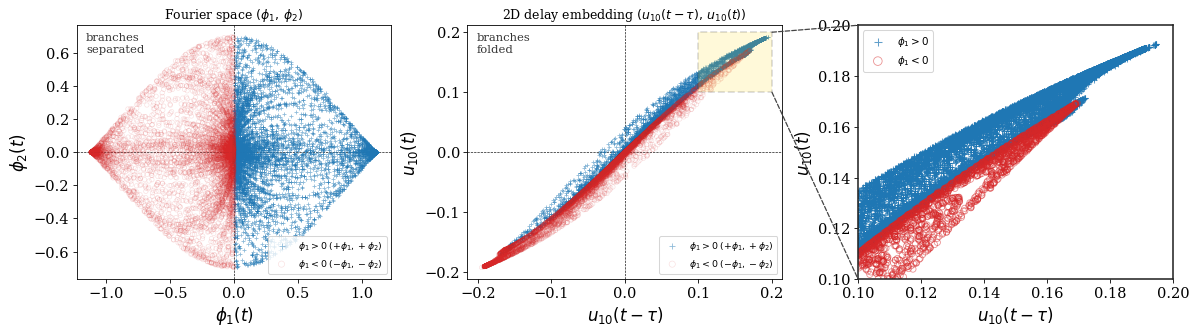}\\[4pt]
    \includegraphics[width=0.95\textwidth]{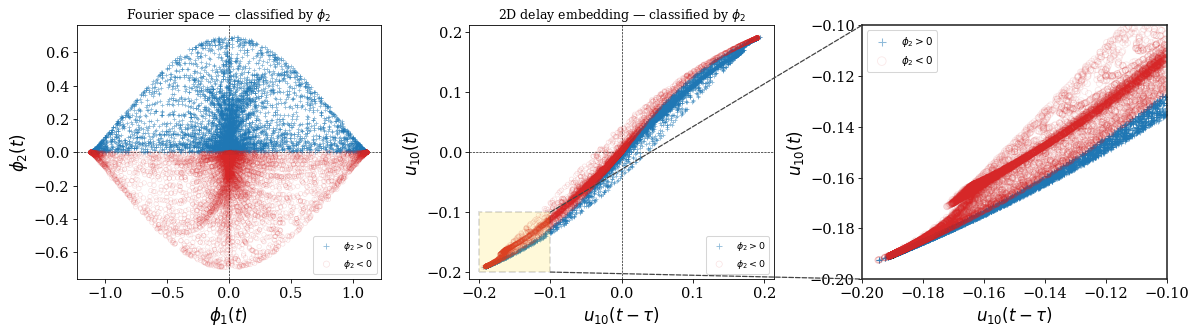}
    \caption{Folding of the Chafee-Infante inertial manifold under the 2D
    delay embedding, \textit{classified} by the sign of $\phi_1$ (top row) and
    $\phi_2$ (bottom row). Left: Fourier coordinate space
    $(\phi_1, \phi_2)$. Middle: 2D delay embedding
    $(u_{10}(t-\tau), u_{10}(t))$. Right: zoomed view of the boxed region.}
    \label{fig:folding}
\end{figure}

In this section, we provide more concrete evidence why a Transformer with $d_{\mathrm{lat}}=2$ is not capable of \textit{unfolding} the inertial manifold. In the left panels of Figure~\ref{fig:folding} we illustrate the inertial manifold in terms of $\phi_1$ and $\phi_2$, colored by the sign of $\phi_1$ (top row) and $\phi_2$ (bottom row). In the Fourier coordinate space $(\phi_1, \phi_2)$ (left panels), the two symmetric branches of the inertial manifold are cleanly separated, as expected from the symmetry of the attractor about the origin. Once the same points are projected onto the two-dimensional delay embedding $(u_{10}(t-\tau), u_{10}(t))$ (middle and right panels), the two branches collapse onto one another: states that are well separated in $(\phi_1, \phi_2)$ nearly fold on top of each other in terms of the scalar observable $u_{10}(t)$ alone. The zoomed insets (right panels) make this overlap more clear, confirming that $u_{10}(t)$ induces a non-injective, folded map from the inertial manifold onto the delay coordinates.

This visualization should be interpreted as evidence of the geometric difficulty faced by the model. The two-delay projection shows that the scalar observable \(u_{10}(t)\) produces a strongly folded view of the inertial manifold: points that are well separated in Fourier coordinates can appear close in low-dimensional delay coordinates. The Transformer is given $n=5$ delayed observations, which should in principle contain sufficient information for reconstruction, but the learned representation must compress this information into the prescribed latent dimension. The poor organization of the $d_{\mathrm{lat}}=2$ latent space therefore suggests that the bottleneck is not simply the absence of temporal information, but the difficulty of unfolding a folded scalar observation while simultaneously compressing it into two coordinates. A third latent coordinate provides additional capacity for this unfolding step before the MLP learns the effective two-dimensional dynamics.

\subsection{Full field cylinder flow reconstruction}
\label{sec:cylinder_recon}

This appendix provides additional details for the Geometric Harmonics reconstruction of the cylinder flow velocity field reported in Section~\ref{sec:cylinder_flow}. We describe the regression setup, the cross-validation protocol, and report per-fold reconstruction errors for both the parameter-aware and parameter-unaware Transformers.

\paragraph{Geometric Harmonics setup}
For each fold, the Geometric Harmonics interpolator is fit on the standardized effective latent coordinates $\mathbf{h}_t$ of the training samples, with the corresponding full velocity-field measurements as targets. The interpolator uses a Gaussian kernel with bandwidth $\varepsilon$ set to a fixed fraction (0.01) of the median pairwise distance among training latent coordinates, and $n_{\mathrm{eig}} = 300$ eigenpairs of the kernel matrix are retained for the interpolation basis. All hyperparameters are held identical across folds and across the parameter-aware and parameter-unaware cases, so that any difference in reconstruction quality is attributable to the latent representation alone.

\paragraph{Cross-validation protocol}
We perform $K = 5$-fold cross-validation, with folds defined by a uniformly random partition of the sample indices using a fixed random seed. For each fold, the interpolator is trained on the remaining four folds and evaluated on the held-out fold. We report two error metrics: the mean absolute error (MAE) over the full reconstructed velocity field, and the MAE restricted to a wake region downstream of the cylinder, defined as the rectangular sub-domain spanning rows $20$--$45$ and columns $35$--$100$ of the discretized $64 \times 128$ field. The wake region was selected to cover the spatial area immediately downstream of the cylinder, where vortex shedding dynamics are most sensitive to the Reynolds number.

\paragraph{Per-fold results}
Table~\ref{tab:cylinder_cv} reports the train and test errors for each fold. Two observations stand out. First, train and test errors are of the same order of magnitude in every fold and for both cases, indicating that the Geometric Harmonics mapping itself generalizes well and is not overfitting. Second, the reconstruction-quality gap between the parameter-aware and parameter-unaware Transformers is consistent across folds: the parameter-unaware reconstruction has both a substantially higher overall error and a disproportionately elevated error in the wake region. The fold-to-fold standard deviation is two orders of magnitude smaller than the difference between the two cases.

\begin{table}[h]
\centering
\scriptsize
\setlength{\tabcolsep}{4pt}
\caption{Per-fold Geometric Harmonics reconstruction MAE for parameter-aware (with-Re) and parameter-unaware (without-Re) Transformers. Wake MAE is restricted to the wake sub-domain.}
\label{tab:cylinder_cv}
\begin{tabular}{llccc}
\toprule
Model & Fold & Train & Test & Wake \\
\midrule
\multirow{6}{*}{With Re}
& 1 & $8{\times}10^{-5}$ & $9{\times}10^{-5}$   & $2.0{\times}10^{-4}$ \\
& 2 & $8{\times}10^{-5}$ & $1.2{\times}10^{-4}$ & $2.7{\times}10^{-4}$ \\
& 3 & $8{\times}10^{-5}$ & $1.0{\times}10^{-4}$ & $2.2{\times}10^{-4}$ \\
& 4 & $7{\times}10^{-5}$ & $1.3{\times}10^{-4}$ & $2.9{\times}10^{-4}$ \\
& 5 & $7{\times}10^{-5}$ & $1.2{\times}10^{-4}$ & $2.6{\times}10^{-4}$ \\
& Mean $\pm$ std & -- & $(1.12{\pm}0.17){\times}10^{-4}$ & $(2.48{\pm}0.40){\times}10^{-4}$ \\
\midrule
\multirow{6}{*}{Without Re}
& 1 & $5.42{\times}10^{-3}$ & $6.95{\times}10^{-3}$ & $1.22{\times}10^{-2}$ \\
& 2 & $5.40{\times}10^{-3}$ & $8.13{\times}10^{-3}$ & $1.47{\times}10^{-2}$ \\
& 3 & $5.58{\times}10^{-3}$ & $6.76{\times}10^{-3}$ & $1.21{\times}10^{-2}$ \\
& 4 & $5.11{\times}10^{-3}$ & $7.73{\times}10^{-3}$ & $1.38{\times}10^{-2}$ \\
& 5 & $5.15{\times}10^{-3}$ & $7.66{\times}10^{-3}$ & $1.37{\times}10^{-2}$ \\
& Mean $\pm$ std & -- & $(7.45{\pm}0.57){\times}10^{-3}$ & $(1.33{\pm}0.11){\times}10^{-2}$ \\
\bottomrule
\end{tabular}
\end{table}

\end{document}